\documentclass[letterpaper,twocolumn,10pt]{article}
\usepackage{usenix2019_v3}

\usepackage{tikz}
\usepackage{amsmath}

\usepackage{filecontents}

\usepackage{appendix}

\usepackage[utf8]{inputenc} 
\usepackage[T1]{fontenc}    

\usepackage{url}

\usepackage{cite}
\usepackage{amsmath,amssymb,amsfonts}
\usepackage{nicefrac}      
\usepackage{algorithm}
\usepackage{algpseudocode}

\usepackage{textcomp}
\usepackage{subcaption}
\usepackage[export]{adjustbox}

\usepackage{booktabs} 
\usepackage{multirow}

\usepackage[skip=3pt,belowskip=-4pt]{subcaption}

\usepackage{color}
\usepackage{xcolor}

\usepackage{balance}

\usepackage{xspace}

\usepackage{verbatim}

\usepackage{afterpage}

\usepackage{authblk}
\usepackage{microtype}     

\usepackage[symbol]{footmisc}

\newcommand{\diff}[2]{\frac{\partial #1}{\partial #2}}
\newcommand{\vct}[1]{\ensuremath{\mathbf{#1}}}
\newcommand{\mat}[1]{\ensuremath{\mathbf{#1}}}
\newcommand{\set}[1]{\ensuremath{\mathcal{#1}}}

\newcommand{\T}{\ensuremath{\top}}

\newcommand{\argmax}{\operatornamewithlimits{\arg\,\max}}

\newcommand{\sign}{\text{sign}}
\newcommand{\argmin}{\operatornamewithlimits{\arg\,\min}}
\newcommand{\bmat}[1]{\begin{bmatrix}#1\end{bmatrix}}

\newcommand{\myparagraph}[1]{\smallskip \noindent \textbf{#1}}
\newcommand{\ie}{{i.e.}\xspace}
\newcommand{\eg}{{e.g.}\xspace}
\newcommand{\etal}{{et al.}\xspace}

\newcommand{\surrogate}{{surrogate}\xspace}
\newcommand{\param}{\ensuremath{\vct w}}

\begin{document}

\date{}

\title{\Large \bf Why Do Adversarial Attacks Transfer? Explaining Transferability of Evasion and Poisoning Attacks}

\author[$\dagger$]{Ambra Demontis}
\author[$\dagger$]{Marco Melis}
\author[$\dagger$]{Maura Pintor}
\author[*]{Matthew Jagielski}
\author[$\dagger$,$\ddagger$]{Battista Biggio}
\author[*]{Alina Oprea}
\author[*]{Cristina Nita-Rotaru}
\author[$\dagger$,$\ddagger$]{Fabio Roli}

\affil[$\dagger$] {Department of Electrical and Electronic Engineering, University of Cagliari, Italy}
\affil[$\ddagger$] {Pluribus One, Italy}
\affil[*] {Northeastern University, Boston, MA, USA}

\maketitle

\begin{abstract}
Transferability captures the ability of an attack against a machine-learning model to be effective against a different, potentially unknown, model. Empirical evidence for transferability  has been shown in previous work, but the underlying reasons why an attack transfers or not are not yet well understood. In this paper, we present a comprehensive analysis aimed to investigate the transferability of both test-time evasion and training-time poisoning attacks. We provide a unifying optimization framework for evasion and poisoning attacks, and a formal definition of transferability of such attacks. We highlight two main factors contributing to attack transferability: the intrinsic adversarial vulnerability of the target model, and the complexity of the surrogate model used to optimize the attack. Based on these insights, we define three metrics that impact an attack's transferability. Interestingly, our results derived from theoretical analysis hold for both evasion and poisoning attacks, and are confirmed experimentally using a wide range of linear and non-linear classifiers and datasets.\footnote[7]{This is the preprint version of our paper accepted for publication at USENIX 2019.}
\end{abstract}

\section{Introduction}

The wide adoption of machine learning (ML) and deep learning algorithms in many critical applications introduces strong incentives for motivated adversaries to manipulate
the results and models generated by these algorithms. Attacks against machine learning systems can happen during multiple stages in the learning pipeline. For instance, in many settings training data is collected online and thus can not be fully trusted. In \emph{poisoning availability attacks}, the attacker controls a certain amount of training data, thus influencing the trained model and ultimately the predictions at testing time on most points in testing set~\cite{Perdisci06,newsome2006paragraph,Nelson08,rubinstein09,biggio12-icml,Newell14,biggio15-icml,Koh17,jagielski18-sp,suciu18-usenix}. \emph{Poisoning integrity attacks} have the goal of modifying predictions on a few targeted points by manipulating the training process~\cite{Koh17,suciu18-usenix}.  On the other hand, \emph{evasion attacks} involve small manipulations of testing data points that results in misprediction at testing time on those points ~\cite{biggio13-ecml,szegedy14-iclr,goodfellow15-iclr,srndic14,Xu16,sharif16-ccs,Dang17,papernot17-asiaccs,carlini17-sp}.

Creating poisoning and evasion attack points is not a trivial task, particularly when many online services avoid disclosing information about their
machine learning algorithms. As a result, attackers are forced to craft their attacks in \emph{black-box} settings,
against a \surrogate model instead of the real model used by the service,
hoping that the attack will be effective on the real model. The {\em transferability} property of an attack is satisfied when an attack developed for a particular machine learning model (\ie, a \surrogate model) is also effective against the target model. Attack transferability was observed in early studies on adversarial examples~\cite{szegedy14-iclr,goodfellow15-iclr} and has gained a lot more  interest in recent years with the advancement of machine learning cloud services. Previous work has reported empirical findings about the transferability of evasion attacks~\cite{biggio13-ecml,szegedy14-iclr,goodfellow15-iclr,papernot16-transf,papernot17-asiaccs,moosavi17-cvpr,dong18-cvpr,liu17-iclr,tramer17-transf,wu18} and, only recently, also on the transferability of  poisoning integrity attacks~\cite{suciu18-usenix}. In spite of these efforts, the question of \emph{when and why do adversarial points transfer} remains largely unanswered.

In this paper we present  the first comprehensive evaluation of transferability of evasion and poisoning availability attacks, understanding the factors contributing to transferability of both attacks. In particular, we consider attacks crafted with gradient-based optimization techniques (e.g., ~\cite{carlini17-sp,madry18-iclr,biggio12-icml}), a popular and successful mechanism used to create attack data points. We unify for the first time evasion and poisoning attacks into an optimization framework that can be instantiated for a range of threat models and adversarial constraints.
We provide a formal definition of transferability and show that, under linearization of the loss function computed under attack, several main factors impact transferability:  the intrinsic \emph{adversarial vulnerability} of the target model, the \emph{complexity} of the surrogate model used to optimize the attacks, and its alignment with the target model.  Furthermore, we derive a new poisoning attack for logistic regression, and perform a comprehensive evaluation of both evasion and poisoning attacks on multiple datasets, confirming our theoretical analysis.

In more detail, the contributions of our work are:

\myparagraph{Optimization framework for evasion and poisoning attacks.} We introduce a unifying framework based on gradient-descent optimization that encompasses both evasion and poisoning attacks. Our framework supports threat models with different adversarial goals (integrity and availability), amount of knowledge available to the adversary (white-box and black-box), as well as different adversarial capabilities (causative or exploratory). Our framework generalizes existing attacks proposed by previous work for evasion~\cite{biggio13-ecml,szegedy14-iclr,goodfellow15-iclr,carlini17-sp,madry18-iclr} and poisoning~\cite{biggio12-icml,mei15-aaai,biggio15-icml,Koh17,biggio17-aisec,jagielski18-sp}.
Under our framework, we derive a novel gradient-based poisoning availability attack against logistic regression.
We remark here that poisoning attacks are more difficult to derive than evasion ones, as they require computing hypergradients from a bilevel optimization problem, to capture the dependency on how the machine-learning model changes while the training poisoning points are modified~\cite{biggio12-icml,mei15-aaai,biggio15-icml,Koh17,biggio17-aisec,jagielski18-sp}.

\myparagraph{Transferability definition and theoretical bound.} We give a formal definition of transferability of evasion and poisoning attacks, and an upper bound on a transfer attack's success. This allows us to derive three metrics connected to \emph{model complexity}. Our formal definition unveils that transferability depends on: (1) the size of input gradients of the target classifier; (2) how well the gradients of the surrogate and target models align; and (3) the variance of the loss landscape optimized to generate the attack points.

\myparagraph{Comprehensive experimental evaluation of transferability.}  We consider a wide range of classifiers, including logistic regression, SVMs with both linear and RBF kernels, ridge regression, random forests, and deep neural networks (both feed-forward and convolutional neural networks), all with different hyperparameter settings to reflect different model complexities. We evaluate the transferability of our attacks on three datasets related to different applications: handwritten digit recognition (MNIST), Android malware detection (DREBIN), and face recognition (LFW). We confirm our theoretical analysis for both evasion and poisoning attacks.

\myparagraph{Insights into transferability.} 
We demonstrate that attack transferability depends strongly on the \emph{complexity} of the target model, \ie, on its inherent vulnerability.  This confirms that reducing the size of input gradients, \eg, via regularization, may allow us to learn more robust classifiers not only against evasion~{\cite{simon18,ross18,varga17-arxiv,lyu15-icdm} but also against poisoning availability attacks. Second, transferability is also impacted by the surrogate model's alignment with the target model. Surrogates with better alignments to their targets (in terms of the angle between their gradients) are more successful at transferring the attack points. Third, surrogate loss functions that are stabler and have lower variance tend to facilitate gradient-based optimization attacks to find better local optima (see Figure~\ref{fig:high-compl-vs-low-compl}). As less complex models exhibit a lower variance of their loss function, they typically result in better surrogates.

\begin{figure}[h]	
	\centering
	\includegraphics[width=.23\textwidth]{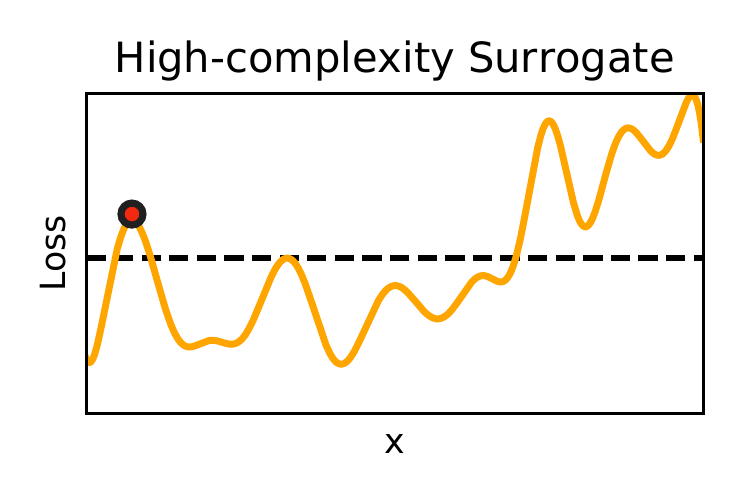}
	\includegraphics[width=.23\textwidth]{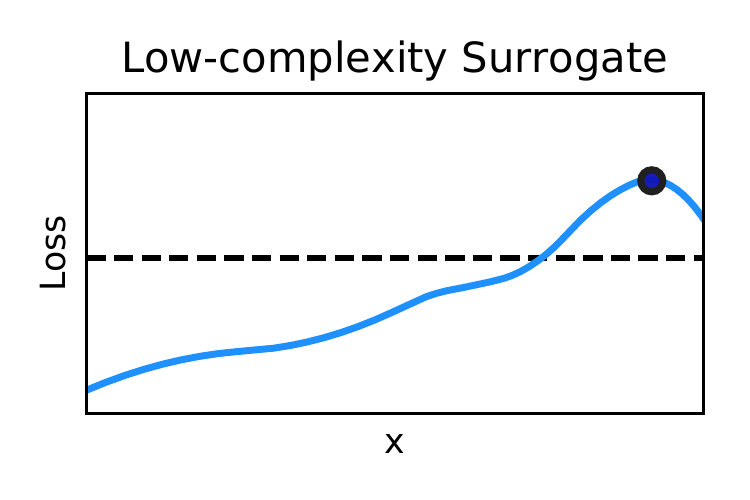}
	\includegraphics[width=.23\textwidth]{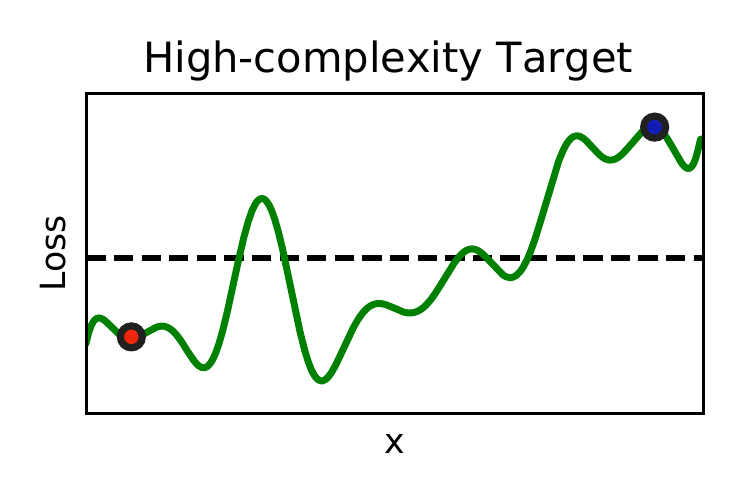}	\includegraphics[width=.23\textwidth]{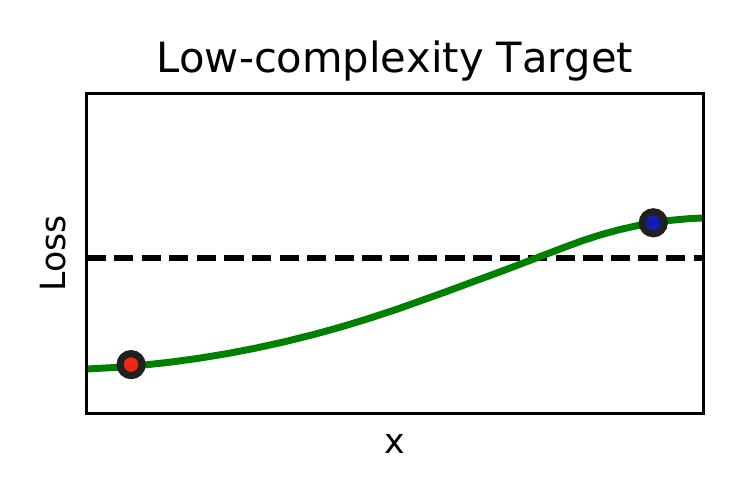}	
	\vspace{-.5em}	
	\caption{Conceptual representation of transferability. We show the loss function of the attack objective as a function of a single feature $x$. The top row includes 2 surrogate models (\emph{high} and \emph{low} complexity), while the bottom row includes both models as targets. The adversarial samples are represented as red dots for the high-complexity surrogate and as blue dots for the low-complexity surrogate. If the adversarial sample loss is below a certain threshold (\ie, the black horizontal line), the point is correctly classified, otherwise it is misclassified. The adversarial point computed against the high-complexity model (top left) lays in a local optimum due to the irregularity of the objective. This point is not effective even against the same classifier trained on a different dataset (bottom left) due to the variance of the high-complexity classifier. The adversarial point computed against the low complexity model (top right), instead, succeeds against both low and high-complexity targets (left and right bottom, respectively).}
	\label{fig:high-compl-vs-low-compl}
\end{figure}

\myparagraph{Organization.}
We discuss background on threat modeling against machine learning in Section~\ref{sect:model}. We introduce our unifying optimization framework for evasion and poisoning attacks, as well as the poisoning attack for logistic regression in Section~\ref{sec:opt}. We then formally define transferability for both evasion and poisoning attacks, and show its approximate connection with the input gradients used to craft the corresponding attack samples (Section~\ref{sect:transfer}). Experiments are reported in Section~\ref{sect:exp}, highlighting connections among regularization hyperparameters, the size of input gradients, and transferability of attacks, on different case studies involving handwritten digit recognition, Android malware detection, and face recognition.
We discuss related work in Section~\ref{sect:related} and conclude in Section~\ref{sec:conclusions}.

\section{Background and Threat Model}
\label{sect:model}

Supervised learning includes: (1) a training phase in which training data is given as input to a learning algorithm, resulting in a trained ML model; (2) a testing phase in which the model is applied to new data and a prediction is generated.
In this paper, we consider a range of adversarial models against machine learning classifiers at both training and testing time. Attackers are defined by: ($i$) their goal or objective in attacking the system; ($ii$) their knowledge of the system; ($iii$) their capabilities in influencing the system through manipulation of the input data. Before we detail each of these, we introduce our notation, and point out that the threat model and attacks considered in this work are suited to binary classification, but can be extended to multi-class settings. 

\myparagraph{Notation.} We denote the sample and label spaces with $\set X$ and $\set Y \in \{-1,+1\}$, respectively, and the training data with $\set D = (\vct x_{i}, y_{i} )_{i=1}^{n}$, where $n$ is the training set size.
We use $L(\set D, \param)$ to denote the \emph{loss} incurred by classifier $f : \set X \mapsto \set Y$ (parameterized by $\param$) on $\set D$. Typically, this is computed by averaging a loss function $\ell(y, \vct x, \param)$ computed on each data point, \ie, $L(\set D, \param) = \frac{1}{n} \sum_{i=1}^n \ell(y_i, \vct x_i, \param)$.
We assume that the classifier $f$ is learned by minimizing an objective function $\set L(\set D, \param)$ on the training data. Typically, this is an estimate of the generalization error, obtained by the sum of the empirical loss $L$ on training data $\set D$ and a regularization term.

\subsection{Threat Model: Attacker's Goal}
\label{sect:goal}
We define the attacker's goal based on the desired security violation. In particular, the attacker may aim to cause either an \emph{integrity} violation, to evade detection without compromising normal system operation; or an \emph{availability} violation, to compromise the normal system functionalities available to legitimate users.

\subsection{Threat Model: Attacker's Knowledge}
\label{sect:knowledge}

We characterize the attacker's knowledge $\vct \kappa$ as a tuple in an abstract knowledge space $\set K$ consisting of four main dimensions, respectively representing knowledge of: ($k.i$) the training data $\set D$; ($k.ii$) the feature set $\set X$; ($k.iii$) the learning algorithm $f$, along with the objective function $\set L$ minimized during training; and ($k.iv$) the parameters $\param$ learned after training the model.
This categorization enables the definition of many different kinds of attacks, ranging from \emph{white-box} attacks with full knowledge of the target classifier to \emph{black-box} attacks in which the attacker has limited information about the target system.

\myparagraph{White-Box Attacks.} We assume here that the attacker has full knowledge of the target classifier, \ie, $\vct \kappa=(\set D, \set X, f, \vct w)$. This setting allows one to perform a worst-case evaluation of the security of machine-learning algorithms, providing empirical upper bounds on the performance degradation that may be incurred by the system under attack.

\myparagraph{Black-Box Attacks.} We assume here that the input feature representation $\set X$ is known. For images, this means that we do consider pixels as the input features, consistently with other recent work on black-box attacks against machine learning~\cite{papernot16-transf,papernot17-asiaccs}. At the same time, the training data $\set D$ and the type of classifier $f$ are not known to the attacker. We consider the most realistic attack model in which the attacker does not have querying access to the classifier.

The attacker can collect a \surrogate~dataset $\hat{\set D}$, ideally sampled from the same underlying data distribution as $\set D$, and train a \emph{\surrogate model} $\hat f$ on such data to approximate the target function $f$. 
Then, the attacker can craft the attacks against $\hat f$, and then check whether they successfully \emph{transfer} to the target classifier $f$.
By denoting limited knowledge of a given component with the \emph{hat} symbol, such black-box attacks can be denoted with $\hat{\vct \kappa}=(\hat{\set D}, \set X, \hat{f}, \hat{\vct w})$.

\subsection{Threat Model: Attacker's Capability}
\label{sect:cap}

This attack characteristic defines how the attacker can influence the system, and how data can be manipulated based on application-specific constraints. If the attacker can manipulate both training and test data, the attack is said to be \emph{causative}. It is instead referred to as \emph{exploratory}, if the attacker can only manipulate test data. These scenarios are more commonly known as \emph{poisoning}~\cite{biggio12-icml,biggio15-icml,mei15-aaai,biggio17-aisec,jagielski18-sp} and \emph{evasion}~\cite{biggio13-ecml,szegedy14-iclr,goodfellow15-iclr,carlini17-sp}.

Another aspect related to the attacker's capability depends on the presence of application-specific constraints on data manipulation; \eg, to evade malware detection, malicious code has to be modified without compromising its intrusive functionality. This may be done against systems leveraging static code analysis, by injecting instructions that will never be executed~\cite{srndic14,demontis17-tdsc,grosse17-esorics}.
These constraints can be generally accounted for in the definition of the optimal attack strategy by assuming that the initial attack sample $\vct x$ can only be modified according to a space of possible modifications $\Phi(\vct x)$. 

\section{Optimization Framework for Gradient-based Attacks}
\label{sec:opt}

We introduce here a general optimization framework that encompasses both evasion and poisoning attacks. Gradient-based attacks have been considered for evasion (e.g., \cite{biggio13-ecml,szegedy14-iclr,goodfellow15-iclr,carlini17-sp,madry18-iclr}) and poisoning (e.g., \cite{biggio12-icml,mei15-aaai,biggio17-aisec,jagielski18-sp}). Our optimization framework not only unifies existing evasion and poisoning attacks, but it also enables the design of new attacks. After defining our general formulation, we instantiate it for evasion and poisoning attacks, and use it to derive a new poisoning availability attack for logistic regression.

\subsection{Gradient-based Optimization Algorithm}
Given the attacker's knowledge $\vct \kappa \in \set K$ and an attack sample $\vct x^{\prime} \in \Phi(\vct x)$ along with its label $y$, the attacker's goal can be defined in terms of an objective function $\set A (\vct x^{\prime}, y, \vct \kappa) \in \mathbb R$ (\eg, a loss function) which measures how effective the attack sample $\vct x^\prime$ is.
The optimal attack strategy can be thus given as:
\begin{eqnarray}
\vct x^{\star} \in \argmax_{\vct x^{\prime} \in \Phi(\vct x)}  \set A (\vct x^{\prime}, y, \vct \kappa) \, .
\label{eq:optim}
\end{eqnarray}
Note that, for the sake of clarity, we consider here the optimization of a single attack sample, but this formulation can be easily extended to account for multiple attack points.
In particular, as in the case of poisoning attacks, the attacker can maximize the objective by iteratively optimizing one attack point at a time~\cite{biggio18,biggio15-icml}.

\myparagraph{Attack Algorithm.} Algorithm~\ref{alg:evasion} provides a general projected gradient-ascent algorithm that can be used to solve the aforementioned problem for both evasion and poisoning attacks. It iteratively updates the attack sample along the gradient of the objective function, ensuring the resulting point to be within the feasible domain through a projection operator $\Pi_\Phi$.
The gradient step size $\eta$ is determined in each update step using a line-search algorithm based on the bisection method, which solves $\max_\eta \set A(\vct x^\prime(\eta), y, \vct \kappa)$, with $\vct x^\prime (\eta)= \Pi_\Phi \left ( \vct x + \eta \nabla_{\vct x} \set A(\vct x, y, \vct \kappa) \right )$. For the line search, in our experiments we consider a maximum of $20$ iterations. This allows us to reduce the overall number of iterations required by Algorithm~\ref{alg:evasion} to reach a local or global optimum.
We also set the maximum number of iterations for Algorithm~\ref{alg:evasion} to $1,000$, but convergence (Algorithm~\ref{alg:evasion}, line~5) is typically reached only after a hundred iterations.

We finally remark that non-differentiable learning algorithms, like decision trees and random forests, can be attacked with more complex strategies~\cite{kantchelian16-icml,ilyas18-icml} or using gradient-based optimization against a differentiable \surrogate learner~\cite{papernot16-transfer,russu16-aisec}.

\begin{algorithm}[t]
	\caption{Gradient-based Evasion and Poisoning Attacks}
	\label{alg:evasion}
	\begin{algorithmic}[1]
		\Require $\vct x, y$: the input sample and its label;
		$\set A(\vct x, y, \vct \kappa)$: the attacker's objective;
		$\vct \kappa = (\set D, \set X, f, \param)$: the attacker's knowledge parameter vector;
		$\Phi(\vct x)$: the feasible set of manipulations that can be made on $\vct x$;
		$t > 0$: a small number.
		\Ensure $\vct x^{\prime}$: the adversarial example.
		\State Initialize the attack sample: $\vct x^{\prime} \gets \vct x$
		\Repeat
		\State Store attack from previous iteration: $\vct x \gets \vct x^\prime$
		\State Update step: $ \vct x^\prime \gets  \Pi_\Phi \left ( \vct x + \eta \nabla_{\vct x} \set A(\vct x, y, \vct \kappa) \right ) $, where the step size $\eta$ is chosen with line search (bisection method), and $\Pi_\Phi$ ensures projection on the feasible domain $\Phi$.
		\Until{$ | \set A (\vct x^\prime, y, \vct \kappa) - \set A (\vct x, y, \vct \kappa) | \le t $}
		\State \Return $\vct x^\prime$
	\end{algorithmic}
\end{algorithm}

\subsection{Evasion Attacks}
\label{sect:evasion}

In evasion attacks, the attacker manipulates test samples to have them misclassified, \ie, to evade detection by a learning algorithm.
For white-box evasion, the optimization problem given in Eq.~\eqref{eq:optim} can be rewritten as:
\begin{eqnarray}
\label{eq:ev-0}
\max_{\vct x^\prime}  & & \ell(y, \vct x^\prime, \param) \, ,\\
\label{eq:ev-1}
{\rm s.t. } && \| \vct x^\prime - \vct x \|_p \leq \varepsilon \, , \\
\label{eq:ev-2}
&& \vct x_{\rm lb} \preceq \vct x^\prime \preceq \vct x_{\rm ub} \, ,
\label{eq:ev-3}
\end{eqnarray}
where $\| \vct v \|_p$ is the $\ell_p$ norm of $\vct v$, and we assume that the classifier parameters $\param$ are known. For the black-box case, it suffices to use the parameters $\hat \param$ of the surrogate classifier $\hat f$.
In this work we consider $\ell(y, \vct x^\prime, \param)=-y f(\vct x^\prime)$, as in~\cite{biggio13-ecml}.

The intuition here is that the attacker maximizes the loss on the adversarial sample with the original class, to cause misclassification to the opposite class.
The manipulation constraints $\Phi(\vct x)$ are given in terms of: ($i$) a distance constraint $\| \vct x^\prime - \vct x \|_p \leq \varepsilon$, which sets a bound on the maximum input perturbation between $\vct x$ (\ie, the input sample) and the corresponding modified adversarial example $\vct x^{\prime}$; and ($ii$) a box constraint $\vct x_{\rm lb} \preceq \vct x^{\prime} \preceq \vct x_{\rm ub}$ (where $\vct u \preceq \vct v$ means that each element of $\vct u$ has to be not greater than the corresponding element in $\vct v$), which bounds the values of the attack sample $\vct x^\prime$.

For images, the former constraint is used to implement either \emph{dense} or \emph{sparse} evasion attacks~\cite{demontis16-spr,russu16-aisec,melis17-vipar}. Normally, the $\ell_{2}$ and the $\ell_{\infty}$ distances between pixel values are used to cause an indistinguishable image blurring effect (by slightly manipulating all pixels). Conversely, the $\ell_{1}$ distance corresponds to a sparse attack in which only few pixels are significantly manipulated, yielding a salt-and-pepper noise effect on the image~\cite{demontis16-spr,russu16-aisec}. The box constraint can be used to bound each pixel value between $0$ and $255$, or to ensure manipulation of only a specific region of the image.
For example, if some pixels should not be manipulated, one can set the corresponding values of $\vct x_{\rm lb}$ and $\vct x_{\rm ub}$ equal to those of $\vct x$.

\begin{figure}[t]
	\centering
	\includegraphics[width=0.41\textwidth]{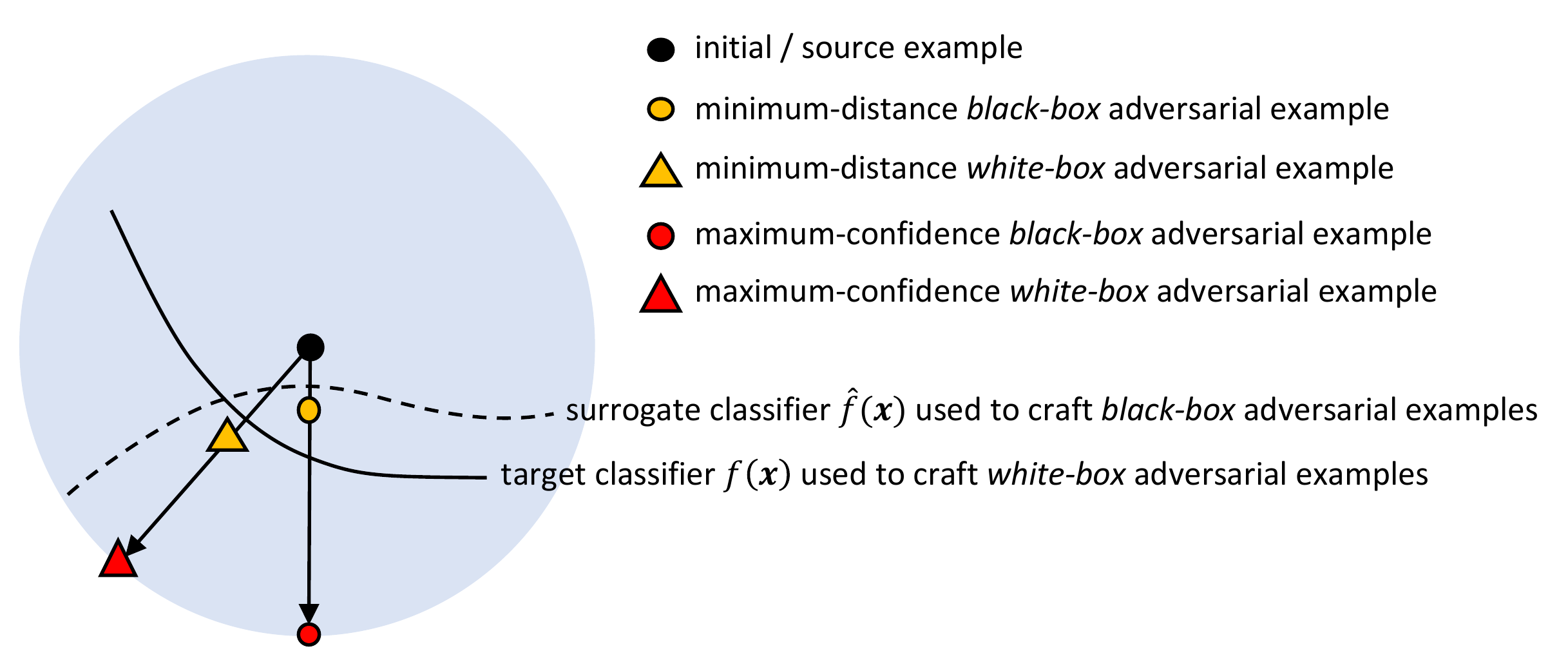}
	\caption{Conceptual representation of maximum-confidence evasion attacks (within an $\ell_2$ ball of radius $\varepsilon$) vs. minimum-distance adversarial examples. Maximum-confidence attacks tend to transfer better as they are misclassified with higher confidence (though requiring more modifications).}
	\vspace{-5pt}
	\label{fig:max-conf}
\end{figure}

\myparagraph{Maximum-confidence vs. minimum-distance evasion.} Our formulation of evasion attacks aims to produce adversarial examples that are misclassified with \emph{maximum confidence} by the classifier, within the given space of feasible modifications.
This is substantially different from crafting minimum-distance adversarial examples, as formulated in~\cite{szegedy14-iclr} and in follow-up work (\eg,~\cite{papernot16-transf}). This difference is conceptually depicted in Fig.~\ref{fig:max-conf}. In particular, in terms of transferability, it is now widely acknowledged that higher-confidence attacks have better chances of successfully transfering to the target classifier (and even of bypassing countermeasures based on gradient masking)~\cite{carlini17-sp,athalye18,dong18-cvpr}. For this reason, in this work we consider evasion attacks that aim to craft adversarial examples misclassified with \emph{maximum} confidence.

\myparagraph{Initialization.} There is another factor known to improve transferability of evasion attacks, as well as their effectiveness in the white-box setting. It consists of running the attack starting from different initialization points to mitigate the problem of getting stuck in poor local optima~\cite{biggio13-ecml,zhang16-tcyb,dong18-cvpr}. In addition to starting the gradient ascent from the initial point $\vct x$, for nonlinear classifiers we also consider starting the gradient ascent from the projection of a randomly-chosen point of the opposite class onto the feasible domain. This double-initialization strategy helps finding better local optima, through the identification of more promising paths towards evasion~\cite{zhang16-tcyb,dong18-cvpr,wu18}.

\subsection{Poisoning Availability Attacks}
\label{sect:poisoning}

Poisoning attacks consist of manipulating training data (mainly by injecting adversarial points into the training set) to either favor intrusions without affecting normal system operation, or to purposely compromise normal system operation to cause a denial of service. The former are referred to as poisoning integrity attacks, while the latter are known as poisoning availability attacks~\cite{biggio18,biggio15-icml}. Recent work has mostly addressed transferability of poisoning integrity attacks~\cite{suciu18-usenix}, including backdoor attacks~\cite{chen17,gu17}. In this work we focus on poisoning availability attacks, as their transferability properties have not yet been widely investigated. Crafting transferable poisoning availability attacks is much more challenging than crafting transferable poisoning integrity attacks, as the latter have a much more modest goal (modifying prediction on a small set of targeted points).

As for the evasion case, we formulate poisoning in a white-box setting, given that the extension to black-box attacks is immediate through the use of \surrogate learners.
Poisoning is formulated as a bilevel optimization problem in which the outer optimization maximizes the attacker's objective $\set A$ (typically, a loss function $L$ computed on untainted data), while the inner optimization amounts to learning the classifier on the poisoned training data~\cite{biggio12-icml,biggio15-icml,mei15-aaai}.
This can be made explicit by rewriting Eq.~\eqref{eq:optim} as:
\begin{eqnarray}	
\label{eq:obj-pois}
\max_{\vct x^\prime} & & L( \set D_{\rm val}, \param^\star) = \sum_{j=1}^m \ell (y_j, \vct x_j, \param^\star) \\
{\rm s.t.} & &  \param^\star \in  \argmin_{\param} \set L (\set D_{\rm tr} \cup (\vct x^\prime, y), \vct \param)
\label{eq:poisoning_problem}
\end{eqnarray}
where ${\set D}_{\rm tr}$ and ${\set D}_{\rm val}$ are the training and validation datasets available to the attacker. The former, along with the poisoning point $\vct x^{\prime}$, is used to train the learner on poisoned data, while the latter is used to evaluate its performance on untainted data, through the loss function $L ({\set D}_{\rm val}, {\param^\star})$.
Notably, the objective function implicitly depends on $\vct x^{\prime}$ through the parameters ${\param^\star}$ of the poisoned classifier.

The attacker's capability is limited by assuming that the attacker can inject only a small fraction $\alpha$ of poisoning points into the training set. Thus, the attacker solves an optimization problem involving a set of poisoned data points ($\alpha n$) added to the training data.

Poisoning points can be optimized via gradient-ascent procedures, as shown in Algorithm~\ref{alg:evasion}. The main challenge is to compute the gradient of the attacker's objective (\ie, the validation loss) with respect to each poisoning point. In fact, this gradient has to capture the implicit dependency of the optimal parameter vector $\vct w^\star$ (learned after training) on the poisoning point being optimized, as the classification function changes while this point is updated. Provided that the attacker function is differentiable w.r.t. $\param$ and $\vct x$, the required gradient can be computed using the chain rule~\cite{biggio12-icml,biggio15-icml,biggio17-aisec,biggio18,mei15-aaai}:
\begin{eqnarray}
\label{eq:poisoning_gradient}	
\nabla_{\vct{x}} \set{A} =  { \nabla_{\vct{x}} L } + {\frac{\partial \vct{\param}}{\partial \vct{x}}}^\T {\nabla_{\vct{w}} L} \, ,
\end{eqnarray}
where the term $\frac{\partial \param}{\partial \vct x}$ captures the implicit dependency of the parameters $\param$ on the poisoning point $\vct x$. Under some regularity conditions, this derivative can be computed by replacing the inner optimization problem with its stationarity (Karush-Kuhn-Tucker, KKT) conditions, \ie, with its implicit equation $\nabla_{\param} \set{L} (\set{D}_{\rm tr} \cup (\vct x^\prime, y ), \param) = \vct 0$~\cite{mei15-aaai,biggio17-aisec}.\footnote{More rigorously, we should write the KKT conditions in this case as $\nabla_{\param} \set{L} (\set{D}_{\rm tr} \cup (\vct x^\prime, y ), \param) \in \vct 0$, as the solution may not be unique.}
By differentiating this expression w.r.t. the poisoning point $\vct{x}$, one yields:
\begin{eqnarray}
\label{eq:implicit_equation}	
\nabla_{\vct{x}} \nabla_{\param} \set {L}  + \frac{\partial \param}{\partial \vct{x}}^\T \nabla_{\vct{w}}^2 \set{L} = \vct 0 \, .
\end{eqnarray}
Solving for $\frac{\partial \param}{\partial \vct{x}}$, we obtain
$\frac{\partial \param}{\partial \vct{x}}^\T = - (\nabla_{\vct{x}} \nabla_{\param} \set{L})(\nabla_{\param}^2 \set{L})^{-1}$, which
can be substituted in Eq.~\eqref{eq:poisoning_gradient} to obtain the required gradient:
\begin{eqnarray}
\label{eq:poisoning_gradient_expanded}	
\nabla_{\vct x} \set{A}  = \nabla_{\vct{x}} L  - (\nabla_{\vct{x}_c} \nabla_{\param} \set{L} )
(\nabla_{\param}^2 \set{L})^{-1}
\nabla_{\param} L \, .
\end{eqnarray}

\myparagraph{Gradients for SVM.} Poisoning attacks against SVMs were first proposed in~\cite{biggio12-icml}. Here, we report a simplified expression for SVM poisoning, with $\set L$ corresponding to the dual SVM learning problem, and $L$ to the hinge loss (in the outer optimization):
\begin{eqnarray}
\label{eq:svm_grad}
\nabla_{\vct{x}_c} \set A = - \alpha_c \diff{\vct{k}_{kc}}{\vct{x}_c} \vct y_k +  \alpha_c \bmat{ \diff{ \vct{k}_{sc} }{\vct{x}_c}  & 0 }
\bmat{
	\mat{K}_{ss} & \vct{1} \\
	\vct{1}^\T & 0  }^{-1}
\bmat{ \mat{K}_{sk} \\ \vct{1}^\T }
\vct y_k \, .
\end{eqnarray}
We use $c$, $s$ and $k$ here to respectively index the attack point, the support vectors, and the validation points for which $\ell(y, \vct x, \param) > 0$ (corresponding to a non-null derivative of the hinge loss). The coefficient $\alpha_c$ is the dual variable assigned to the poisoning point by the learning algorithm, and $\vct k$ and $\mat K$ contain kernel values between the corresponding indexed sets of points.

\myparagraph{Gradients for Logistic Regression.} Logistic regression is a linear classifier that estimates the probability of the positive class using the sigmoid function. A poisoning attack against logistic regression has been derived in~\cite{mei15-aaai}, but maximizing a different outer objective and not directly the validation loss.
One of our contributions is to compute gradients for logistic regression under our optimization framework.
Using logistic loss as the attacker's loss, the poisoning gradient for logistic regression can be computed as:
\begin{eqnarray}
\label{eq:logistic derivatives}	
\nabla_{\vct{x}_c} \set A =  - \bmat{ \nabla_{\vct{x}_c} \nabla_{\vct{\theta}} \set{L} \\ C \;  z_c \; \vct{\theta}}^\T
\bmat{ \nabla_{\vct{\theta}}^2 \set{L}  &  \mat{X} \; \vct{z} \; C \;   \\ C \; \vct{z} \; \mat{X} &  C \sum_{i}^{n} z_i }^{-1} 
\bmat{ \mat{X} (\vct{y}\circ \vct{\sigma} -\vct{y})  \\ \vct{y}^\T (\vct{\sigma}-1) } C, \nonumber
\end{eqnarray}
where $\vct{\theta}$ are the classifier weights (bias excluded), $\circ$ is the element-wise product, $\vct{z}$ is equal to $\vct{\sigma} (1- \vct{\sigma})$, $\sigma$ is the sigmoid of the signed discriminant function (each element of that vector is therefore: $\sigma_i= \frac{1}{1 + \exp(-y_i f_i )} $ with $f_i=\vct{x}_i \vct{\theta} + b$), and:

\begin{eqnarray}
&&\nabla^2_{\vct{\theta}} \set{L} = C \sum_{i}^{n} \vct{x_i} z_i \vct{x}^\T_i + \mathbb{I} ,\\
&& \nabla_{\vct{x}_c} \nabla_{\vct{\theta}} \set{L} =  C( \mathbb{I} \circ (y_c\sigma_c -y_c) + z_c \vct{\theta} \vct{x}_c^\T )
\end{eqnarray}
In the above equations, $\mathbb{I}$ is the identity matrix.

\section{Transferability Definition and Metrics}
\label{sect:transfer}

We discuss here an intriguing connection among transferability of both evasion and poisoning attacks, input gradients and model complexity, and highlight the factors impacting transferability between a surrogate and a target model. Model complexity is a measure of the capacity of a learning algorithm to fit the training data. It is typically penalized to avoid overfitting by reducing either the number of classifier parameters to be learnt or their size (\eg, via regularization)~\cite{bishop:prml:book:2007}. Given that complexity is essentially controlled by the hyperparameters of a given learning algorithm (\eg, the number of neurons in the hidden layers of a neural network, or the regularization hyperparameter $C$ of an SVM), \emph{only models that are trained using the same learning algorithm should be compared in terms of complexity}. As we will see, this is an important point to correctly interpret the results of our analysis. For notational convenience, we denote in the following the attack points as $\vct x^\star = \vct x + \hat{\vct \delta}$, where $\vct x$ is the initial point and $\hat{\vct \delta}$ the adversarial perturbation optimized by the attack algorithm against the \emph{surrogate} classifier, for both evasion and poisoning attacks.
We start by formally defining transferability for evasion attacks, and then discuss how this definition and the corresponding metrics can be generalized to poisoning.

\myparagraph{Transferability of Evasion Attacks.} Given an evasion attack point $\vct x^\star$, crafted against a surrogate learner (parameterized by $\hat \param$), we define its \emph{transferability}
as the loss attained by the target classifier $f$ (parameterized by $\param$) on that point, \ie,
$T= \ell(y, \vct x + \hat{\vct \delta}, \param)$.
This can be simplified through a linear approximation of the loss function as:
\begin{equation}
T= \ell(y, \vct x + \hat{\vct \delta}, \param) \approxeq \ell(y, \vct x, \param ) + \hat{\vct \delta}^\T \nabla_{\vct x} \ell(y, \vct x, \param) \, .
\label{eq:lin-transf}
\end{equation}
This approximation may not only hold for sufficiently-small input perturbations. It may also hold for larger perturbations if the classification function is linear or has a small curvature (\eg, if it is strongly regularized). It is not difficult to see that, for any given point $\vct x, y$, the evasion problem in Eqs.~\eqref{eq:ev-0}-\eqref{eq:ev-1} (without considering the feature bounds in Eq.~\ref{eq:ev-2}) can be rewritten as:
\begin{equation}
\hat{\vct \delta} \in \argmax_{\| \vct \delta \|_p \leq \varepsilon} \ell(y, \vct x + \vct \delta, \hat \param) \, .
\label{eq:delta}
\end{equation}
Under the same linear approximation, this corresponds to the maximization of an inner product over an $\varepsilon$-sized ball:
\begin{equation}
\max_{\| \vct \delta \|_p \leq \varepsilon} \vct \delta^\T \nabla_{\vct x} \ell(y, \vct x, \hat \param)  = \varepsilon \| \nabla_{\vct x} \ell(y, \vct x, \hat \param) \|_q \, ,
\end{equation}
where $\ell_q$ is the dual norm of $\ell_p$.

The above problem is maximized as follows:
\begin{enumerate}
	\item
	For $p=2$, the maximum is $\hat{\vct \delta} = \varepsilon  \frac{\nabla_{\vct x} \ell(y, \vct x, \hat \param)}{\| \nabla_{\vct x} \ell(y, \vct x, \hat \param) \|_2}$;
	
	\item For $p=\infty$, the maximum is $\hat{\vct \delta} \in \varepsilon \cdot  \sign\{\nabla_{\vct x} \ell(y, \vct x, \hat \param)\}$;
	
	\item For $p=1$, the maximum is achieved by setting the values of $\hat{\vct \delta}$ that correspond to the maximum absolute values of $\nabla_{\vct x} \ell(y, \vct x, \hat \param)$ to their sign, \ie, $\pm 1$, and $0$ otherwise.
\end{enumerate}

Substituting the optimal value of $\hat{\vct \delta}$ into Eq.~\eqref{eq:lin-transf}, we can compute the loss increment $\Delta \ell = \hat{\vct \delta}^\T \nabla_{\vct x} \ell(y, \vct x, \param)$ under a transfer attack in closed form; \eg, for $p=2$, it is given as:
\begin{eqnarray}
\Delta \ell  =  \varepsilon \frac{{\nabla_{\vct x} \hat \ell}^\T}{\| \nabla_{\vct x} \hat \ell \|_2} \nabla_{\vct x} \ell
\leq  \varepsilon \| \nabla_{\vct x} \ell  \|_2\, ,
\label{eq:deltaloss}
\end{eqnarray}
where, for compactness, we use $\hat \ell = \ell(y, \vct x, \hat \param)$ and $\ell = \ell(y, \vct x, \param)$.

In this equation, the left-hand side is the increase in the loss function in the black-box case, while the right-hand side corresponds to the white-box case. The upper bound is obtained when the surrogate classifier $\hat{\param}$ is equal to the target $\param$ (white-box attacks). Similar results hold for $p=1$ and $p=\infty$ (using the dual norm in the right-hand side).

\myparagraph{Intriguing Connections and Transferability Metrics.} The above findings reveal some interesting connections among transferability of attacks, model complexity (controlled by the classifier hyperparameters) and input gradients, as detailed below, and allow us to define simple and computationally-efficient transferability metrics. 

\emph{(1) Size of Input Gradients.} The first interesting observation is that transferability depends on the size of the gradient of the loss $\ell$ computed using the \emph{target} classifier, regardless of the surrogate: the larger this gradient is, the larger the attack impact may be. This is inferred from the upper bound in Eq.~\eqref{eq:deltaloss}.
We define the corresponding metric $S(\vct x, y)$ as:
\begin{equation}
\label{eq:S}
S(\vct x, y) = \| \nabla_{\vct x} \ell(y, \vct  x, \param)  \|_q \, ,
\end{equation}
where $q$ is the dual of the perturbation norm.

The size of the input gradient also depends on the complexity of the
given model, controlled, \eg, by its regularization hyperparameter.
Less complex, strongly-regularized classifiers tend to have smaller input gradients, \ie, they learn smoother functions that are more robust to attacks, and vice-versa. Notably, this holds for both evasion and poisoning attacks (\eg, the poisoning gradient in Eq.~\ref{eq:svm_grad} is proportional to $\alpha_c$, which is larger when the model is weakly regularized). 
In Fig.~\ref{fig:input-grads} we report an example showing how increasing regularization (\ie, decreasing complexity) for a neural network trained on MNIST89 (see Sect.~\ref{sect:mnist-evasion}), by controlling its \emph{weight decay}, reduces the average size of its input gradients, improving adversarial robustness to evasion.
	It is however worth remarking that, since complexity is a model-dependent characteristic, the size of input gradients cannot be directly compared across different learning algorithms; \eg, if a linear SVM exhibits larger input gradients than a neural network, we cannot conclude that the former will be more vulnerable. 

\begin{figure}
	\centering
	\includegraphics[width=.75\columnwidth]{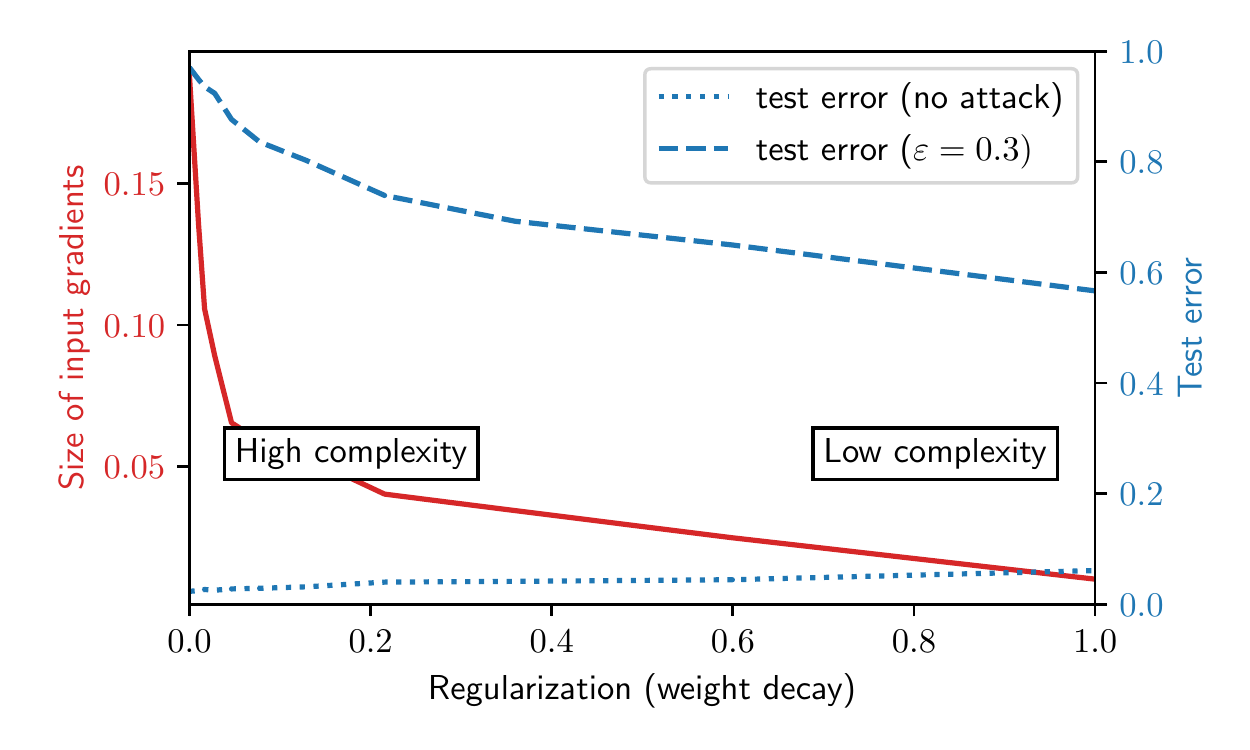}
	\caption{Size of input gradients (averaged on the test set) and test error (in the absence and presence of evasion attacks) against regularization (controlled via weight decay) for a neural network trained on MNIST89 (see Sect.~\ref{sect:mnist-evasion}). Note how the size of input gradients and the test error under attack decrease as regularization (complexity) increases (decreases).}
	\vspace{-4pt}
	\label{fig:input-grads}
\end{figure}

Another interesting observation is that, if a classifier has large input gradients (\eg, due to high-dimensionality of the input space and low level of regularization), for an attack to succeed it may suffice to apply only tiny, \emph{imperceptible} perturbations. As we will see in the experimental section, this explains why adversarial examples against deep neural networks can often only be slightly perturbed to mislead detection, while when attacking less complex classifiers in low dimensions, modifications become more evident.

\emph{(2) Gradient Alignment.} The second relevant impact factor on transferability is based on the alignment of the input gradients of the loss function computed using the target and the surrogate learners. If we compare the increase in the loss function in the black-box case (the left-hand side of Eq.~\ref{eq:deltaloss}) against that corresponding to white-box attacks (the right-hand side), we find that the relative increase in loss, at least for $\ell_2$ perturbations, is given by the following value:
\begin{equation}
\label{eq:R}
R(\vct x, y)=\frac{ {\nabla_{\vct x} \hat{\ell}}^\T \nabla_{\vct x} \ell }{\| \nabla_{\vct x} \hat \ell \|_2 \| \nabla_{\vct x} \ell  \|_2} \, .
\end{equation}
Interestingly, this is exactly the cosine of the angle between the gradient of the loss of the surrogate and that of the target classifier. This is a novel finding which explains why the cosine angle metric between the target and surrogate gradients can well characterize the transferability of attacks, confirming empirical results from previous work~\cite{liu17-iclr}. For other kinds of perturbation, this definition slightly changes, but gradient alignment can be similarly evaluated. Differently from the gradient size $S$, gradient alignment is a pairwise metric, allowing comparisons across different surrogate models; \eg, if a surrogate SVM is better aligned with the target model than another surrogate, we can expect that attacks targeting the surrogate SVM will transfer better.

\begin{figure}
	\centering
	\includegraphics[width=.79\columnwidth]{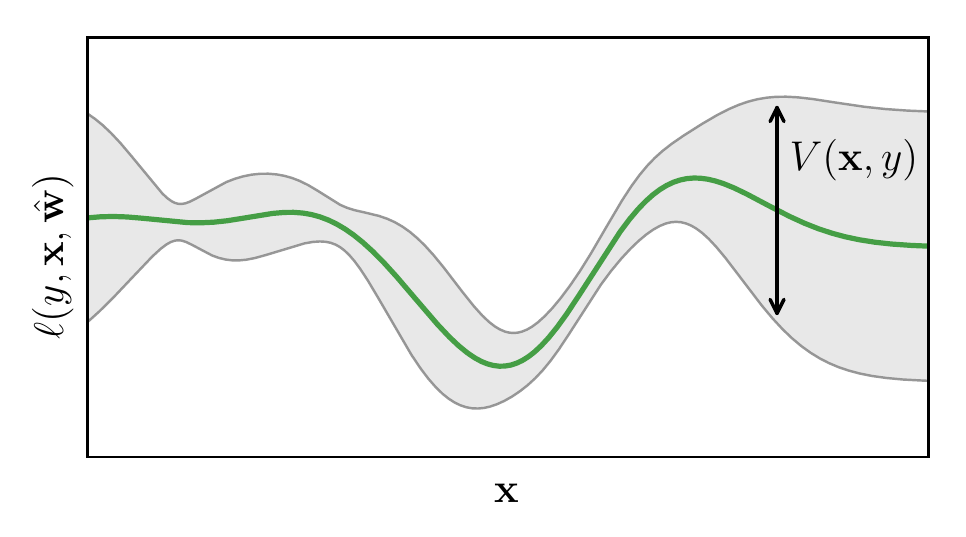}
	\caption{Conceptual representation of the variability of the loss landscape. The green line represents the expected loss with respect to different training sets used to learn the surrogate model, while the gray area represents the variance of the loss landscape. If the variance is too large, local optima may change, and the attack may not successfully transfer.}
	\vspace{-7pt}
	\label{fig:loss-landscape}
\end{figure}

\emph{(3) Variability of the Loss Landscape.} We define here another useful metric to characterize attack transferability. The idea is to measure the variability of the loss function $\hat \ell$ when the training set used to learn the surrogate model changes, even though it is sampled from the same underlying distribution.
The reason is that the loss $\hat \ell$ is exactly the objective function $\set A$ optimized by the attacker to craft evasion attacks (Eq.~\ref{eq:optim}).
Accordingly, if this loss landscape changes dramatically even when simply resampling the surrogate training set (which may happen, \eg, for surrogate models exhibiting a large error variance, like neural networks and decision trees), it is very likely that the local optima of the corresponding optimization problem will change, and this may in turn imply that the attacks will not transfer correctly to the target learner.

We define the variability of the loss landscape simply as the \emph{variance} of the loss, estimated at a given attack point $\vct x, y$:
\begin{equation}
\label{eq:V}
V(\vct x, y) = \mathbb E_{\set D}\{\ell(y, \vct x, \hat \param)^2\} - \mathbb E_{\set D}\{\ell(y, \vct x, \hat \param)\}^2 \, ,
\end{equation}
where $\mathbb E_{\set D}$ is the expectation taken with respect to different (surrogate) training sets.
This is very similar to what is typically done to estimate the variance of classifiers' predictions.
This notion is clarified also in Fig.~\ref{fig:loss-landscape}.
As for the size of input gradients $S$, also the loss variance $V$ should only be compared across models trained with the same learning algorithm.

The transferability metrics $S$, $R$ and $V$ defined so far depend on the initial attack point $\vct x$ and its label $y$. In our experiments, we will compute their mean values by averaging on different initial attack points.

\myparagraph{Transferability of Poisoning Attacks.} For poisoning attacks, we can essentially follow the same derivation discussed before. Instead of defining transferability in terms of the loss attained on the modified test point, we define it in terms of the validation loss attained by the target classifier under the influence of the poisoning points.
This loss function can be linearized as done in the previous case, yielding: $T \approxeq L(\set D, \param) + \hat{\vct \delta}^\T \nabla_{\vct x} L(\set D, \param)$, where $\set D$ are the untainted validation points, and $\hat{\vct \delta}$ is the perturbation applied to the initial poisoning point $\vct x$ against the surrogate classifier. Recall that $L$ depends on the poisoning point through the classifier parameters $\param$, and that the gradient $\nabla_{\vct x} L(\set D, \param)$ here is equivalent to the generic one reported in Eq.~\eqref{eq:poisoning_gradient_expanded}.
It is then clear that the perturbation $\hat{\vct \delta}$ maximizes the (linearized) loss when it is best aligned with its derivative $\nabla_{\vct x} L(\set D, \param)$, according to the constraint used, as in the previous case.
The three transferability metrics defined before can also be used for poisoning attacks provided that we simply replace the evasion loss $\ell(y, \vct x, \param)$ with the validation loss $L(\set D, \param)$.

\section{Experimental Analysis}
\label{sect:exp}

In this section, we evaluate the transferability of both evasion and poisoning attacks across a range of ML models. We highlight some interesting findings about transferability, based on the three metrics developed in Sect.~\ref{sect:transfer}. In particular, we analyze attack transferability in terms of its connection to the size of the input gradients of the loss function,
the gradient alignment between surrogate and target classifiers, and the variability of the loss function optimized to craft the attack points. We provide recommendations on how to choose the most effective surrogate models to craft transferable attacks in the black-box setting.

\subsection{Transferability of Evasion Attacks}

We start by reporting our experiments on evasion attacks.
We consider here two distinct case studies, involving handwritten digit recognition and Android malware detection.

\subsubsection{Handwritten Digit Recognition}
\label{sect:mnist-evasion}

The MNIST89 data includes the MNIST handwritten digits from classes 8 and 9. Each digit image consists of 784 pixels ranging from 0 to 255, normalized in $[0,1]$ by dividing such values by $255$. We run $10$ independent repetitions to average the results on different training-test splits. In each repetition, we run white-box and black-box attacks, using $5,900$ samples to train the target classifier, $5,900$ distinct samples to train the surrogate classifier (without even relabeling the surrogate data with labels predicted by the target classifier; \ie, we do not perform any query on the target), and $1,000$ test samples. We modified test digits in both classes using Algorithm~\ref{alg:evasion} under the $\ell_2$ distance constraint $\| \vct x - \vct x^\prime \|_2 \leq \varepsilon$, with $\varepsilon \in [0,5]$.

\begin{figure}[t]
	\centering
	\includegraphics[width=.85\columnwidth]{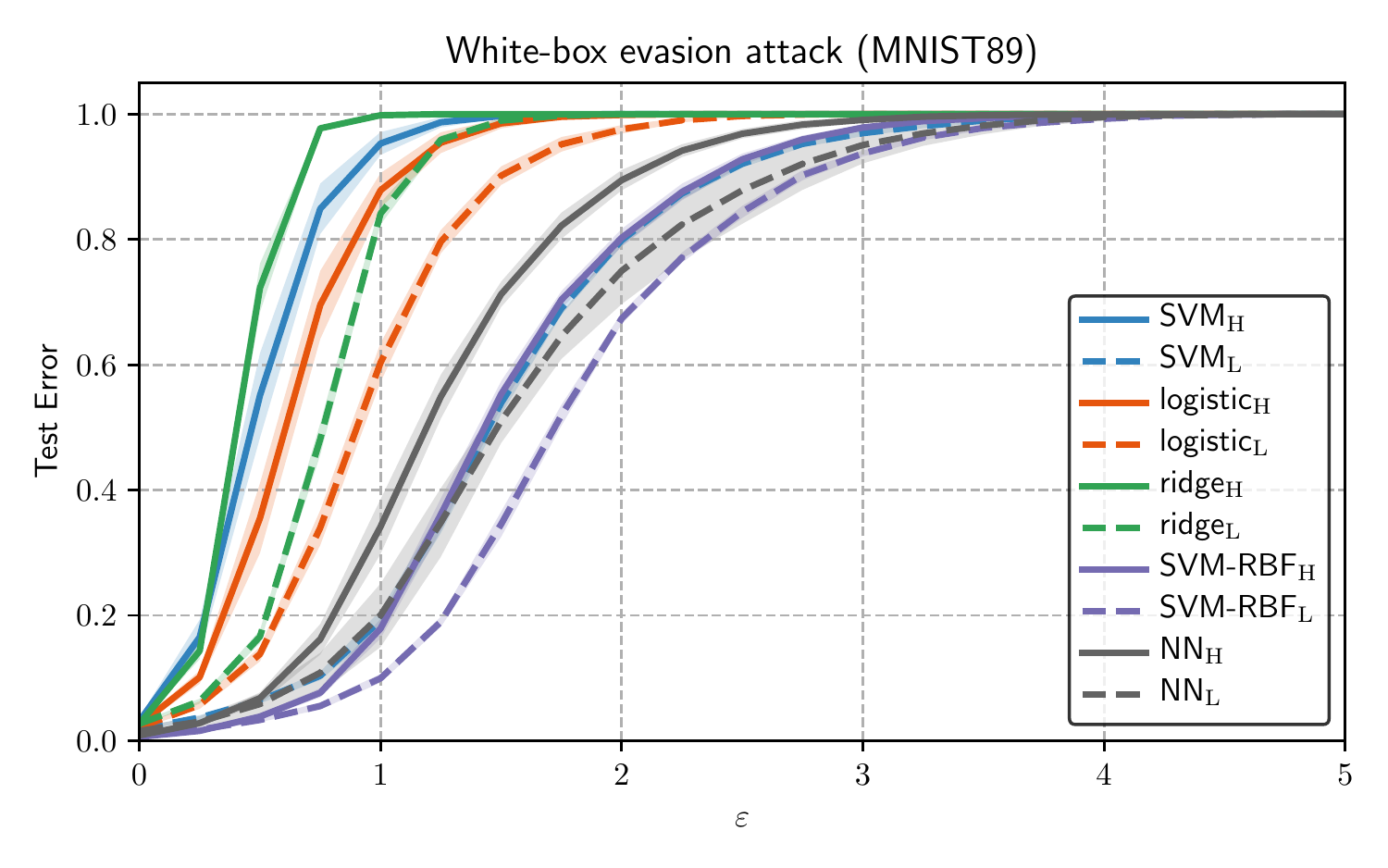}
	\vspace{-1em}
	\caption{White-box evasion attacks on MNIST89. Test error against increasing maximum perturbation $\varepsilon$.}
	\label{fig:ev-pk-mnist89}
\end{figure}

For each of the following learning algorithms, we train a high-complexity (H) and a low-complexity (L) model, by changing its hyperparameters: ($i$) SVMs with linear kernel (SVM$_{\mathrm H}$ with $C=100$ and SVM$_{\mathrm L}$ with $C=0.01$); ($ii$) SVMs with RBF kernel (SVM-RBF$_{\mathrm H}$ with $C=100$ and SVM-RBF$_{\mathrm L}$ with $C=1$, both with $\gamma=0.01$);
($iii$) logistic classifiers (logistic$\rm{_{H}}$ with $C=10$ and logistic$\rm{_{L}}$ with $C=1$);
($iv$) ridge classifiers (ridge$\rm{_{H}}$ with $\alpha = 1$ and ridge$\rm{_{L}}$ with $\alpha = 10$);\footnote{Recall that the level of regularization increases as $\alpha$ increases, and as $C$ decreases.}
($v$) fully-connected neural networks with two hidden layers including 50 neurons each, and ReLU activations (NN$_{\mathrm H}$ with no regularization, \ie, weight decay set to $0$, and NN$_{\mathrm L}$ with weight decay set to $0.01$), trained via cross-entropy loss minimization; and ($vi$) random forests consisting of $30$ trees (RF$_{\mathrm H}$ with no limit on the depth of the trees and RF$_{\mathrm L}$ with a maximum depth of 8).
These configurations are chosen to evaluate the robustness of classifiers that exhibit similar test accuracies but different levels of complexity.

\myparagraph{How does model complexity impact evasion attack success in the white-box setting?} The results for white-box evasion attacks are reported for all classifiers that fall under our framework and can be tested for evasion with gradient-based attacks (SVM, Logistic, Ridge, and NN). This excludes random forests, as they are not differentiable.
We report the complete \emph{security evaluation curves}~\cite{biggio18} in Fig.~\ref{fig:ev-pk-mnist89}, showing the mean test error (over $10$ runs) against an increasing maximum admissible distortion $\varepsilon$.
In Fig.~\ref{fig:ev-scatter-mnist89-a} we report the mean test error at $\varepsilon=1$ for each target model against the size of its input gradients (S, averaged on the test samples and on the $10$ runs).

\begin{figure*}[ht]
	\centering
	\begin{subfigure}[t]{0.22\textwidth}
		\includegraphics[width=\textwidth]{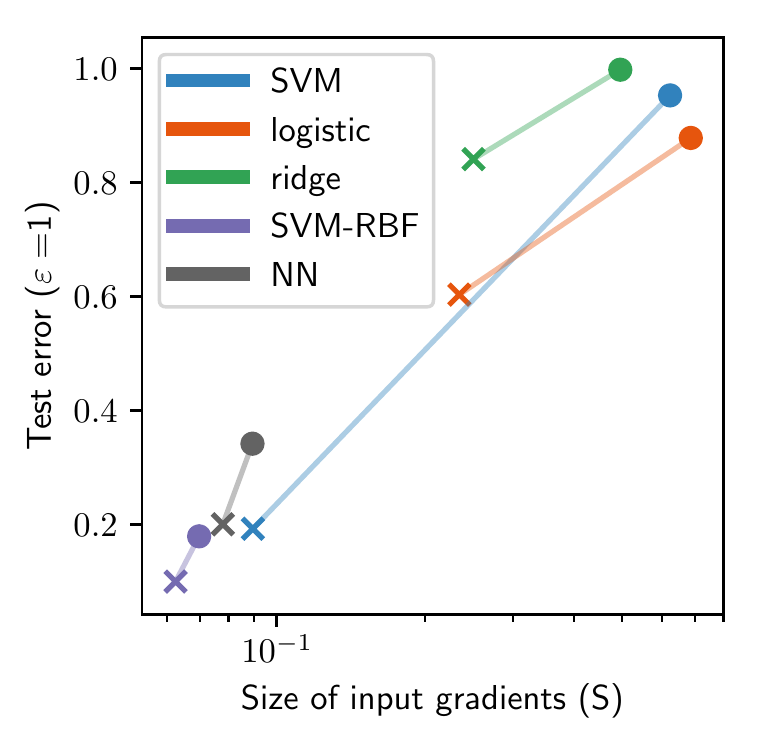}
		\subcaption{} \label{fig:ev-scatter-mnist89-a}
	\end{subfigure}
	\begin{subfigure}[t]{0.225\textwidth}
		\includegraphics[width=\textwidth]{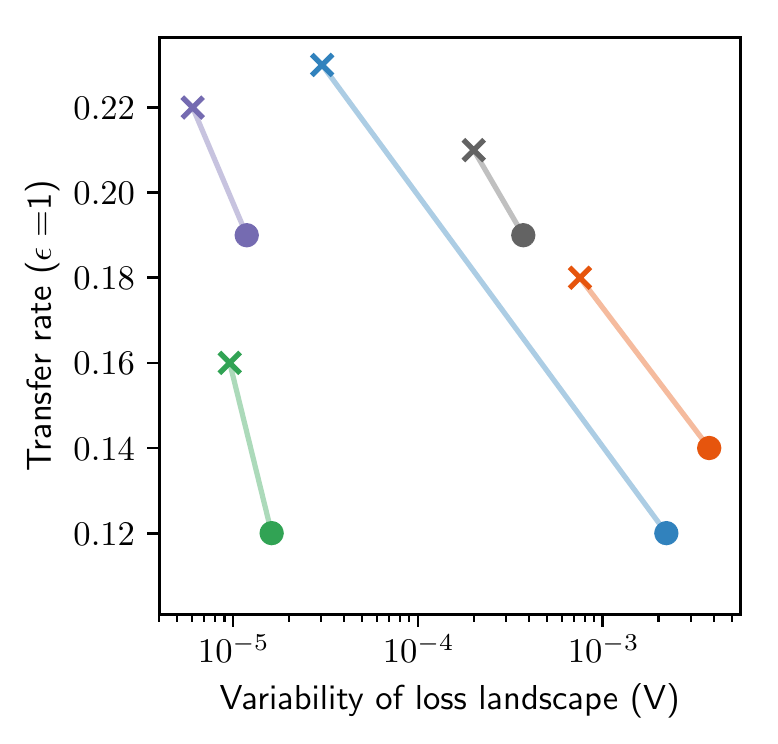}
		\subcaption{} \label{fig:ev-scatter-mnist89-b}
	\end{subfigure}
	\begin{subfigure}[t]{0.222\textwidth}
		\includegraphics[width=\textwidth]{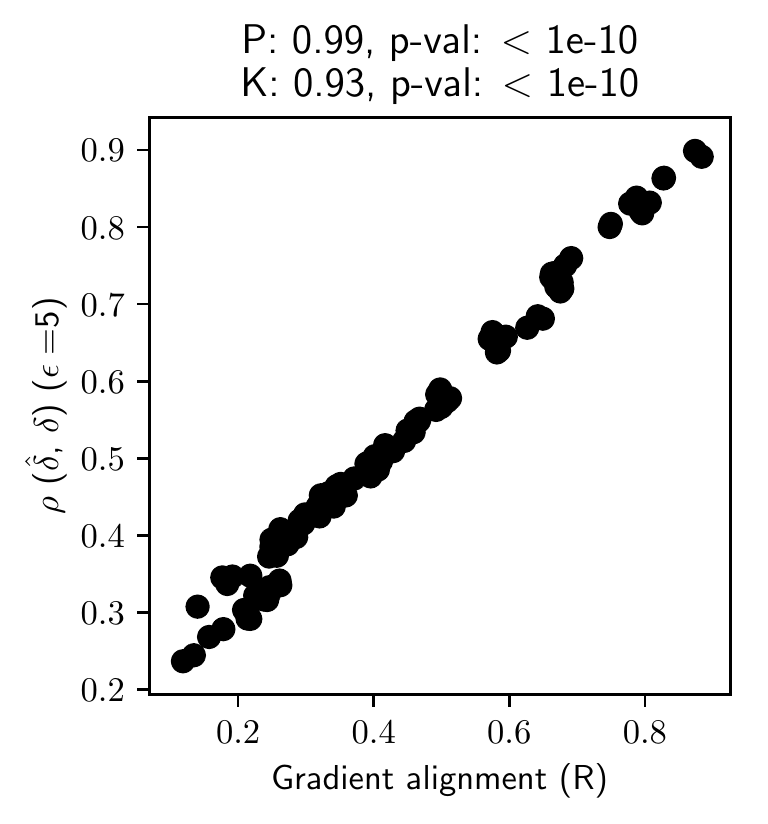}
		\subcaption{} \label{fig:ev-scatter-mnist89-c}
	\end{subfigure}
	\begin{subfigure}[t]{0.219\textwidth}
		\includegraphics[width=\textwidth]{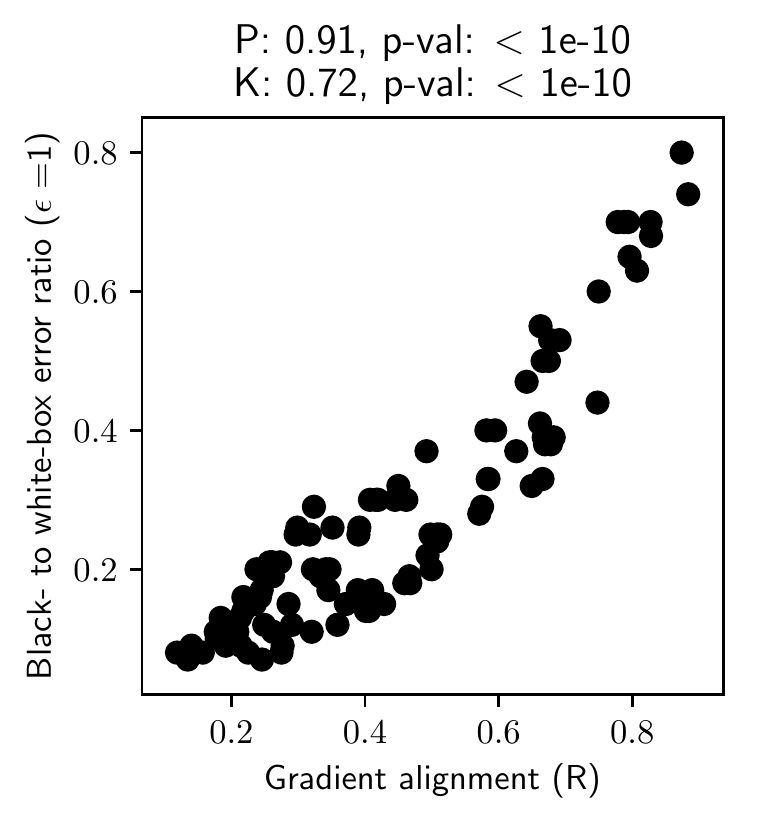}
		\subcaption{} \label{fig:ev-scatter-mnist89-d}
	\end{subfigure}
	\caption{Evaluation of our metrics for evasion attacks on MNIST89. (a) Test error under attack vs average size of input gradients ($S$) for low- (denoted with `$\times$') and high-complexity (denoted with `$\circ$') classifiers. (b) Average transfer rate vs variability of loss landscape (V). (c) Pearson correlation coefficient $\rho(\hat{\vct \delta},\vct \delta)$ between black-box ($\hat{\vct \delta}$) and white-box ($\vct \delta$) perturbations (values in Fig.~\ref{fig:ev-corr-angles-mnist89}, right) vs gradient alignment (R, values in Fig.~\ref{fig:ev-corr-angles-mnist89}, left) for each target-surrogate pair. Pearson (P) and Kendall (K) correlations between $\rho$ and R are also reported along with the $p$-values obtained from a permutation test to assess statistical significance.}
	\label{fig:ev-scatter-mnist89}
\end{figure*}

The results show that, for each learning algorithm, the low-complexity model has smaller input gradients, and it is less vulnerable to evasion than its high-complexity counterpart, confirming our theoretical analysis.
This is also confirmed by the $p$-values reported in Table~\ref{tab:stats} (first column), obtained by running a binomial test for each learning algorithm to compare the white-box test error of the corresponding high- and low-complexity models. All the $p$-values are smaller than $0.05$, which confirms 95\% statistical significance.
Recall that these results hold only when comparing models trained using the same learning algorithm. This means that we can compare, \eg, the $S$ value of SVM$_{\mathrm H}$ against SVM$_{\mathrm L}$, but not that of SVM$_{\mathrm H}$ against logistic$_{\mathrm H}$. In fact, even though logistic$_{\mathrm H}$ exhibits the largest $S$ value, it is not the most vulnerable classifier.
Another interesting finding is that nonlinear classifiers tend to be less vulnerable than linear ones.

\begin{table}[t]
	\centering
	\resizebox{0.48\textwidth}{!}{%
		\begin{tabular}{@{}r c rr  rr c rr  rr  @{}}
			
			\multicolumn{1}{l}{} && \multicolumn{4}{c}{\textbf{Evasion}}                              && \multicolumn{4}{c}{\textbf{Poisoning}}                         \\ \cmidrule{3-6} \cmidrule{8-11}
			\multicolumn{1}{l}{} && \multicolumn{2}{c}{\textbf{MNIST89}} & \multicolumn{2}{c}{\textbf{DREBIN}} && \multicolumn{2}{c}{\textbf{MNIST89}} & \multicolumn{2}{c}{\textbf{LFW}} \\ \cmidrule{3-6} \cmidrule{8-11} 
			&&  $\epsilon=1$ & $\epsilon=1$  &  $\epsilon=5$ & $\epsilon=30$  && $5\%$ & $20\%$ &  $5\%$ & $20\%$ \\
			\toprule
			\textbf{SVM}  &&  <1e-2 & <1e-2  &  <1e-2 & <1e-2  && <1e-2  & <1e-2 &  <1e-2  & 0.75 \\
			\textbf{logistic} && <1e-2 &  <1e-2 & <1e-2 & 0.02 && <1e-2 & <1e-2 & 0.10 & 0.21 \\
			\textbf{ridge} && <1e-2 & <1e-2 & <1e-2 & <1e-2 && 0.02  & <1e-2  & 0.02  &  0.75   \\
			\textbf{SVM-RBF}  && <1e-2 & <1e-2 & <1e-2 & <1e-2 && <1e-2  & <1e-2 & <1e-2 &  0.11 \\
			\textbf{NN} && <1e-2 & <1e-2 & <1e-2 & 0.02 &&              &              &            &            \\ \bottomrule
		\end{tabular}%
	}
	\caption{Statistical significance of our results. For each attack, dataset and learning algorithm, we report the $p$-values of two two-sided binomial tests, to respectively reject the null hypothesis that: ($i$) for white-box attacks, the test errors of the high- and low-complexity target follow the same distribution; and ($ii$) for black-box attacks, the transfer rates of the high- and low-complexity surrogate follow the same distribution.
			Each test is based on $10$ samples, obtained by comparing the error of the high- and low-complexity models for each learning algorithm in each repetition. In the first (second) case, success corresponds to a larger test (transfer) error for the high-complexity target (low-complexity surrogate).}
	\vspace{-5pt}
	\label{tab:stats}
\end{table}

\myparagraph{\bf How do evasion attacks transfer between models in black-box settings?} In Fig.~\ref{fig:ev-lk-mnist89} we report the results for black-box evasion attacks, in which the attacks against surrogate models (in rows) are transferred to the target models (in columns). The top row shows results for surrogates trained using only 20\% of the surrogate training data, while in the bottom row surrogates are trained using all surrogate data, \ie, a training set of the same size as that of the target. The three columns report results for $\varepsilon \in \{1, 2, 5\}$.

\begin{figure*}[ht]
	\centering
	\includegraphics[width=.331\textwidth,trim={0 2cm 0 0},clip]{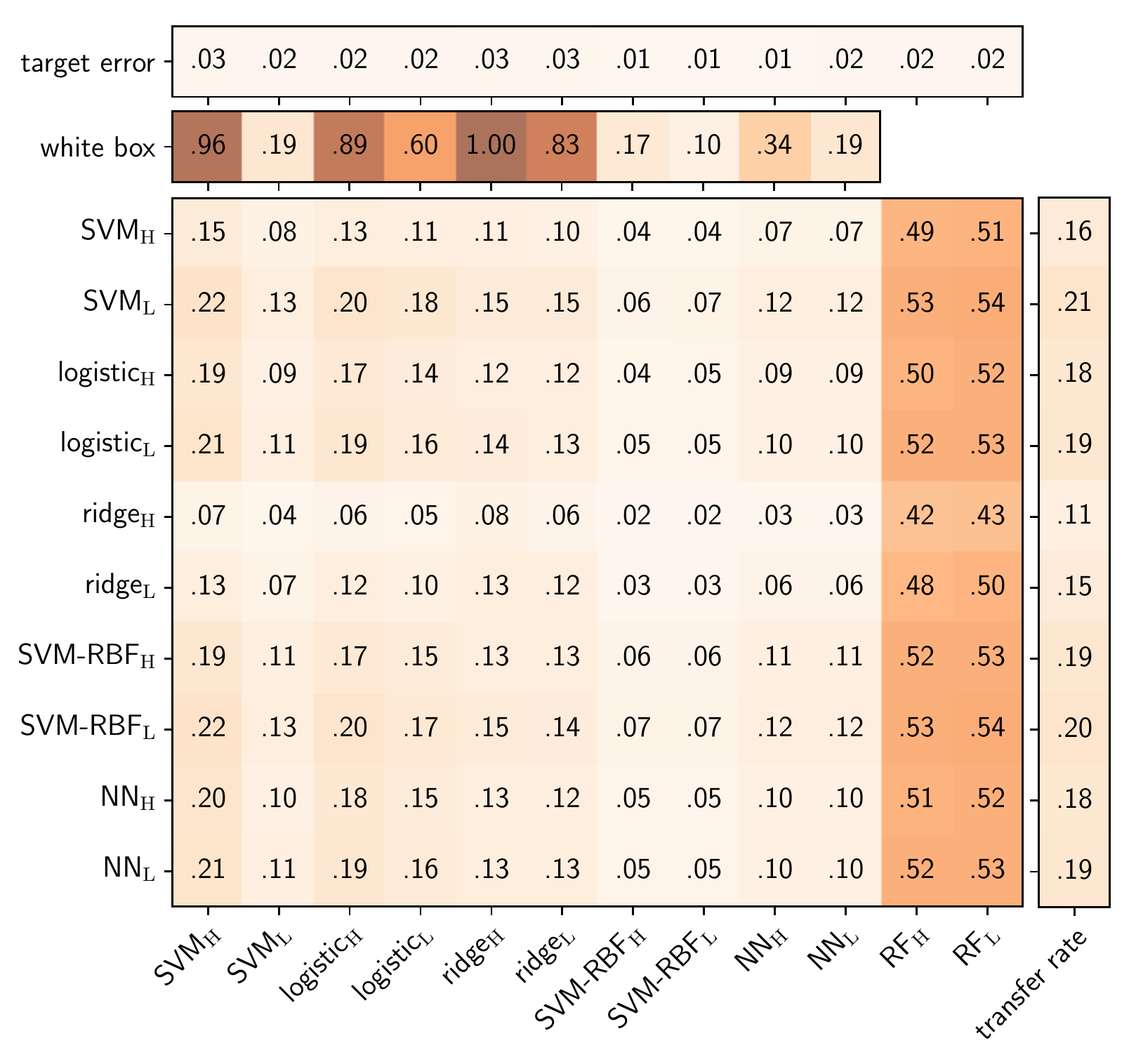}
	\includegraphics[width=.285\textwidth,trim={2.26cm 2cm 0 0},clip]{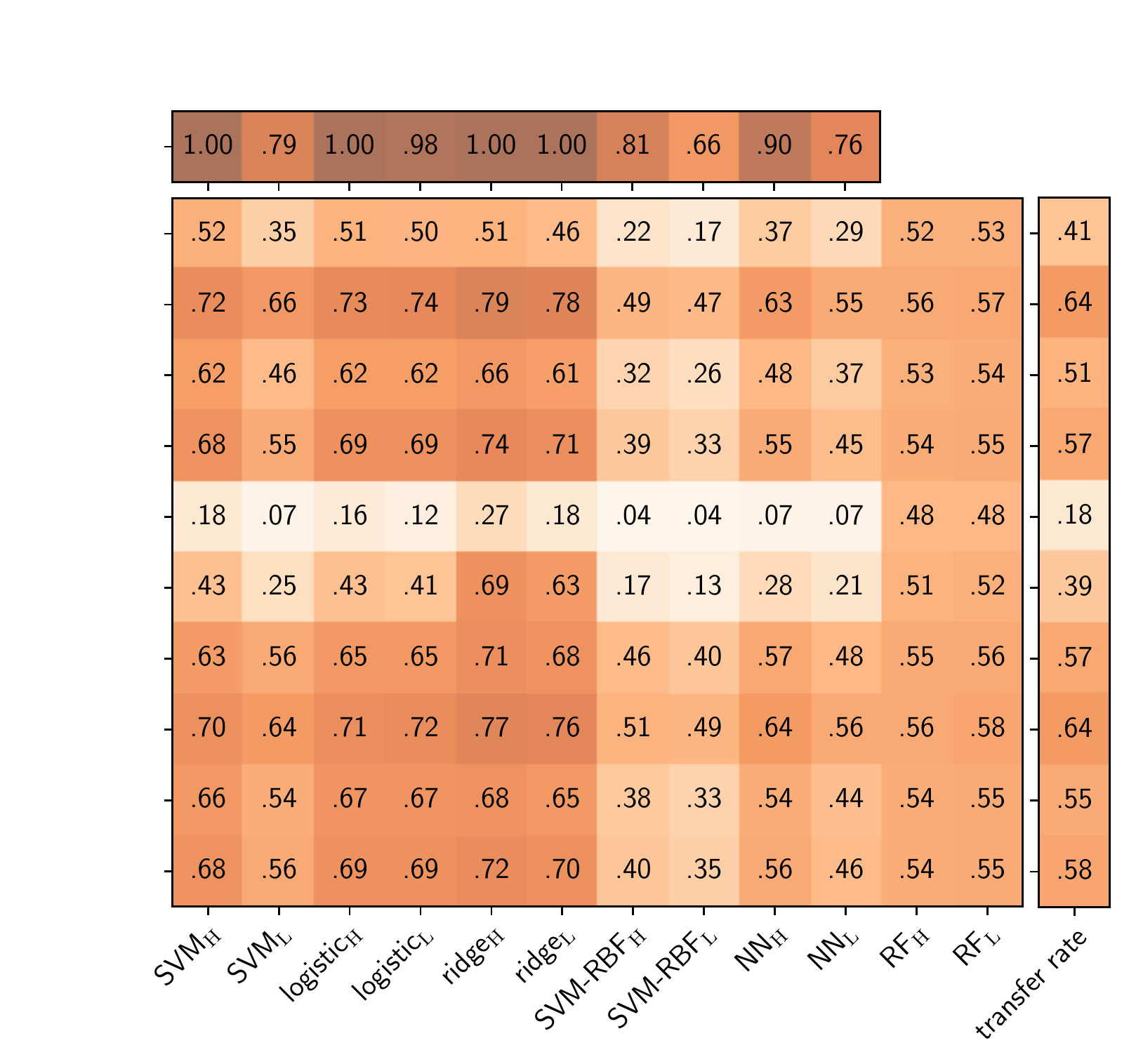}
	\includegraphics[width=.285\textwidth,trim={2.26cm 2cm 0 0},clip]{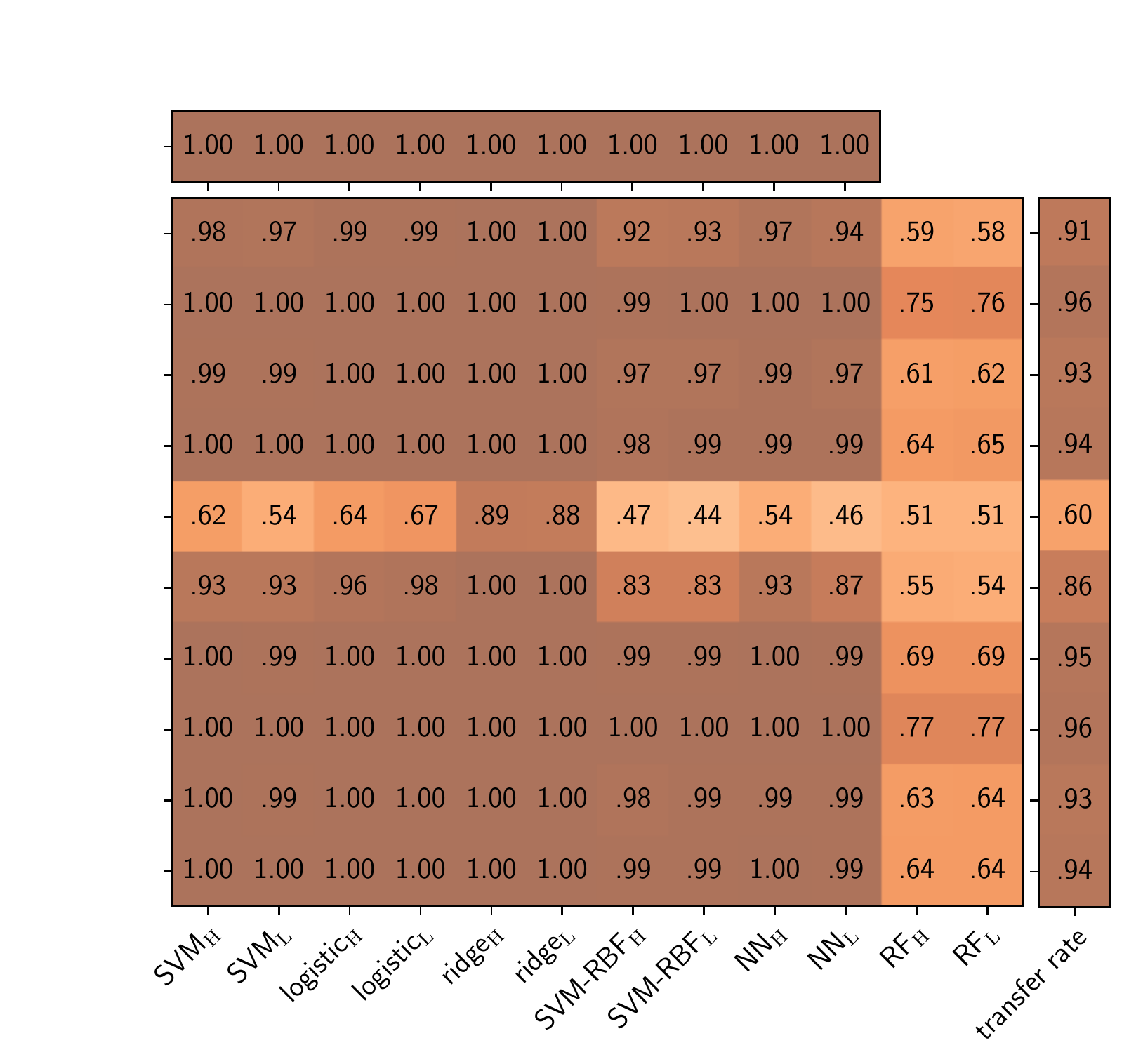}\vspace{.4em}\\
	\begin{subfigure}[t]{0.332\textwidth}
		\centering
		\includegraphics[width=\textwidth,trim={0 0 0 1.6cm},clip]{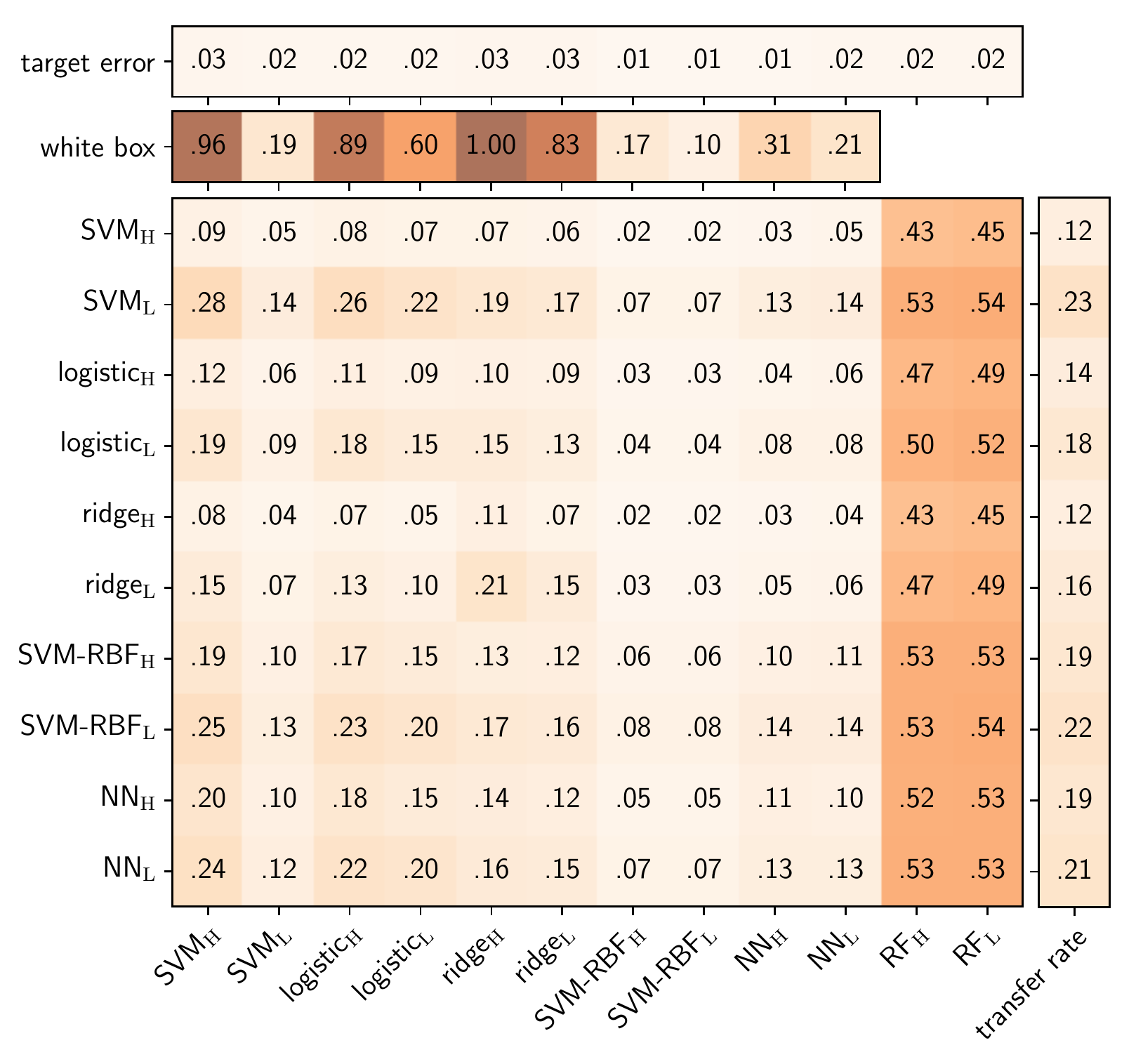}
		\subcaption{$\varepsilon = 1$}
	\end{subfigure}
	\begin{subfigure}[t]{0.285\textwidth}
		\centering
		\includegraphics[width=\textwidth,trim={2.26cm 0 0 1.6cm},clip]{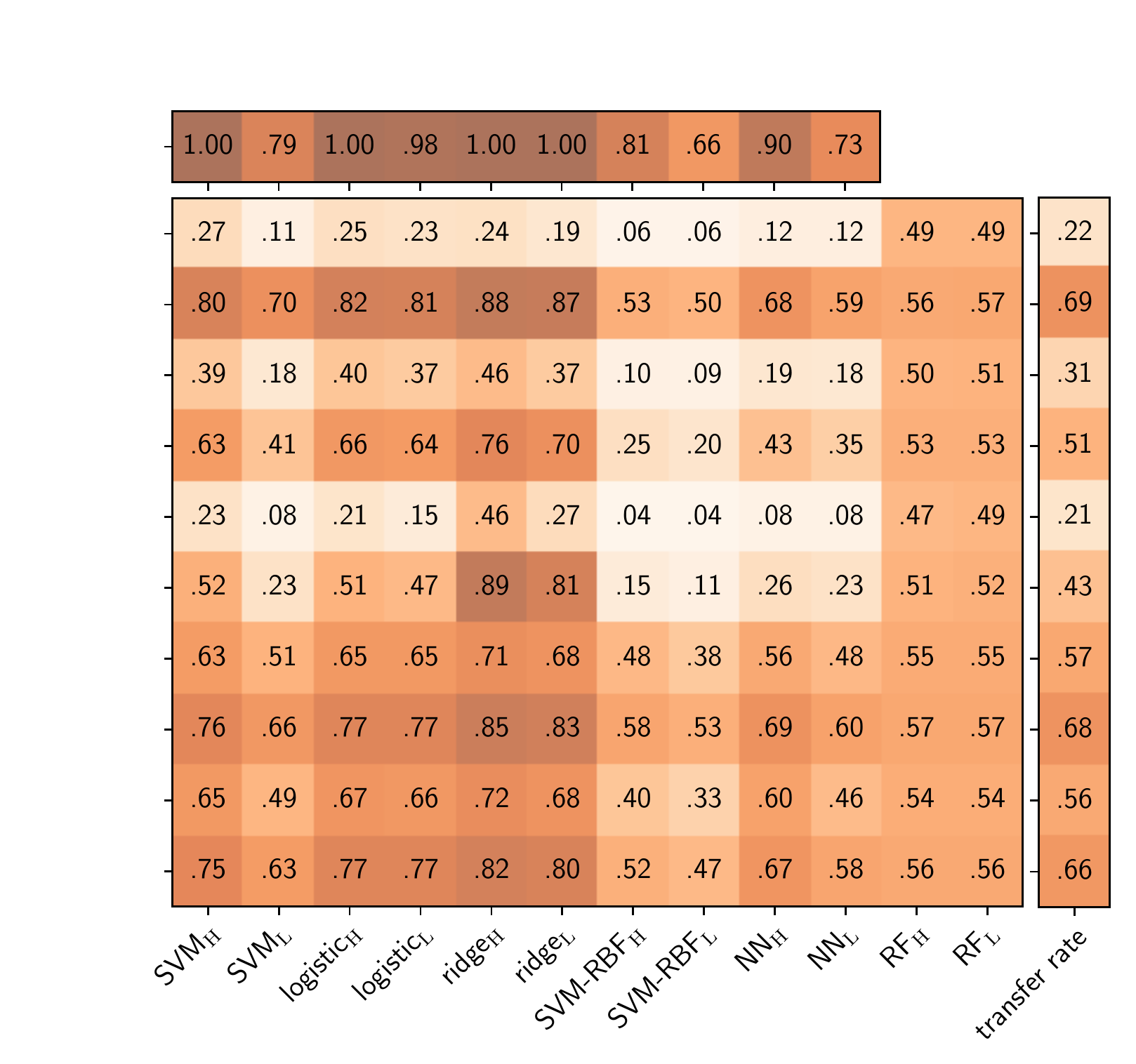}
		\subcaption{$\varepsilon = 2$}
	\end{subfigure}
	\begin{subfigure}[t]{0.285\textwidth}
		\centering
		\includegraphics[width=\textwidth,trim={2.26cm 0 0 1.6cm},clip]{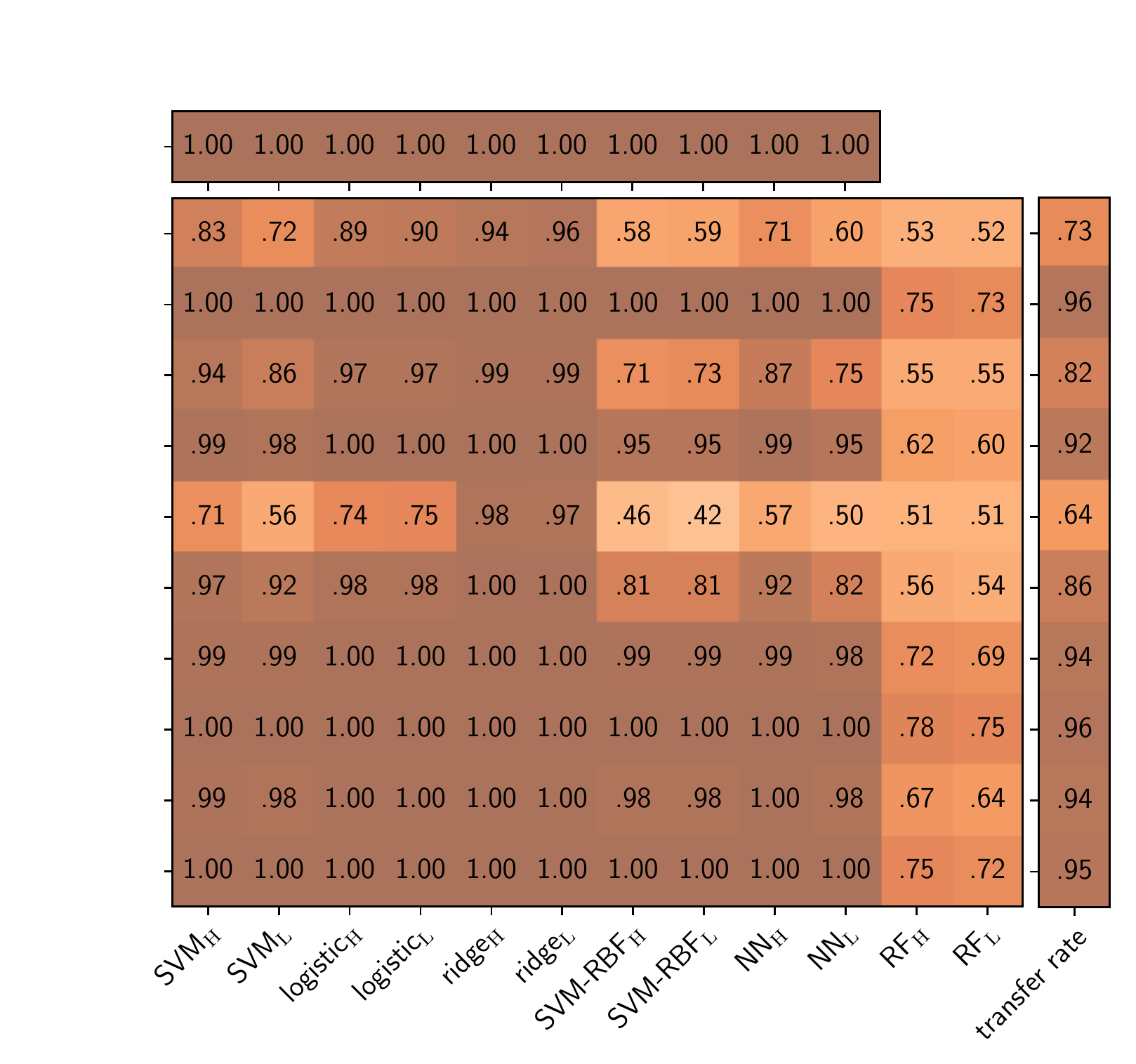}
		\subcaption{$\varepsilon = 5$}
	\end{subfigure}
	\caption{Black-box (transfer) evasion attacks on MNIST89. Each cell contains the test error of the target classifier (in columns) computed on the attack samples crafted against the surrogate (in rows).
		Matrices in the top (bottom) row correspond to attacks crafted against surrogate models trained with 20\% (100\%) of the surrogate training data, for $\varepsilon \in \{1, 2, 5\}$.
		The test error of each target classifier in the absence of attack (target error) and under (white-box) attack are also reported for comparison, along with the mean transfer rate of each surrogate across targets. Darker colors mean higher test error, \ie, better transferability.}
	\vspace{-3pt}
	\label{fig:ev-lk-mnist89}
\end{figure*}

It can be noted that lower-complexity models (with stronger regularization) provide better surrogate models, on average. In particular, this can be seen best in the middle column for medium level of perturbation, in which the lower-complexity models (SVM$_{\mathrm L}$, logistic$_{\mathrm L}$, ridge$_{\mathrm L}$, and SVM-RBF$_{\mathrm L}$) provide on average higher error when transferred to other models. The reason is that they learn smoother and stabler functions, that are capable of better approximating the target function.
Surprisingly, this holds also when using only 20\% of training data, as the black-box attacks relying on such low-complexity models still transfer with similar test errors. This means that most classifiers can be attacked in this black-box setting with almost no knowledge of the model, no query access, but provided that one can get a small amount of data similar to that used to train the target model.

\begin{figure}[ht]
	\centering
	\includegraphics[width=.25\textwidth]{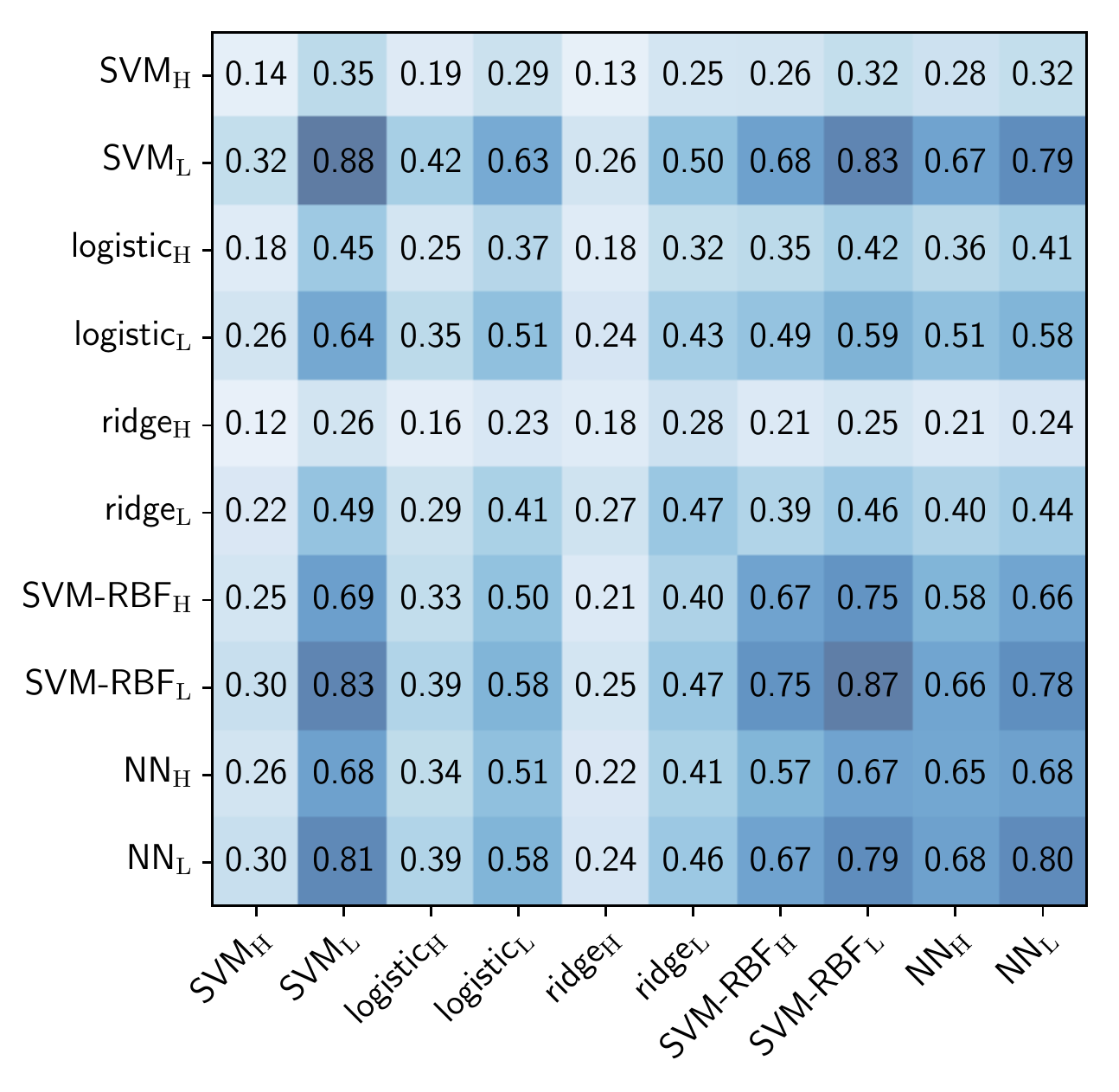}
	\includegraphics[width=.207\textwidth,trim={2.26cm 0 0 0},clip]{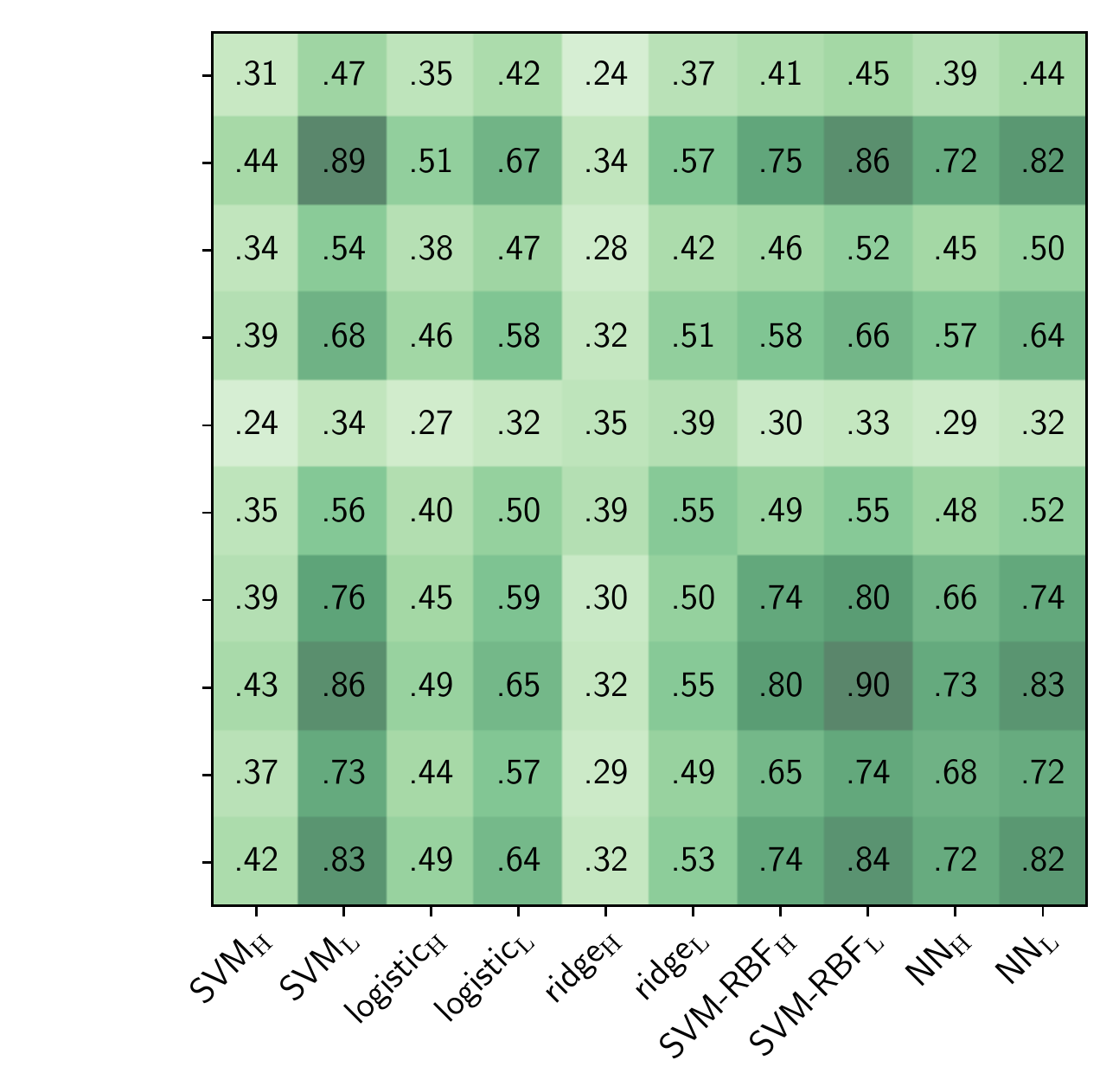}
	\vspace{-.5em}
	\caption{Gradient alignment and perturbation correlation for evasion attacks on MNIST89. \emph{Left:}
		Gradient alignment $R$ (Eq.~\ref{eq:R}) between surrogate (rows) and target (columns) classifiers, averaged on the unmodified test samples. \emph{Right:} Pearson correlation coefficient $\rho(\vct \delta, \hat{\vct \delta})$ between white-box and black-box perturbations for $\varepsilon=5$.}
	\label{fig:ev-corr-angles-mnist89}
\end{figure}

These findings are also confirmed by looking at the variability of the loss landscape, computed as discussed in Sect.~\ref{sect:transfer} (by considering $10$ different training sets), and reported against the average transfer rate of each surrogate model in Fig.~\ref{fig:ev-scatter-mnist89-b}.
It is clear from that plot that higher-variance classifiers are less effective as surrogates than their less-complex counterparts, as the former tend to provide worse, unstable approximations of the target classifier.
To confirm the statistical significance of this result, for each learning algorithm we also compare the mean transfer errors of high- and low-complexity surrogates with a binomial test whose $p$-values (always lower than $0.05$) are reported in Table~\ref{tab:stats} (second column).

Another interesting, related observation is that the adversarial examples computed against lower-complexity surrogates have to be perturbed more to evade (see Fig.~\ref{fig:evasion-digits}), whereas the perturbation of the ones computed against complex models can be smaller. This is again due to the instability induced by high-complexity models into the loss function optimized to craft evasion attacks, whose sudden changes cause the presence of closer local optima to the initial attack point.

\emph{On the vulnerability of random forests.} A noteworthy finding is that random forests can be effectively attacked at small perturbation levels using most other models (see last two columns in Fig.~\ref{fig:ev-lk-mnist89}). We looked at the learned trees and discovered that trees often are susceptible to small changes. In one example, a node of the tree checked if a particular feature value was above 0.002, and classified samples as digit 8 if that condition holds (and digit 9 otherwise). The attack modified that feature from 0 to 0.028, causing it to be immediately misclassified. This vulnerability is intrinsic in the selection process of the threshold values used by these decision trees to split each node. The threshold values are selected among the existing values in the dataset (to correctly handle categorical attributes). Therefore, for pixels which are highly discriminant (\eg, mostly black for one class and white for the other), the threshold will be either very close to one extreme or the other, making it easy to subvert the prediction by a small change.
Since $\ell_2$-norm attacks change almost all feature values, with high probability the attack modifies at least one feature on every path of the tree, causing misclassification.

\myparagraph{Is gradient alignment an effective transferability metric?} In Fig.~\ref{fig:ev-corr-angles-mnist89}, we report on the left the gradient alignment computed between surrogate and target models, and on the right the Pearson correlation coefficient $\rho(\hat{\vct \delta}, \vct \delta)$ between the perturbation optimized against the surrogate (\ie, the black-box perturbation $\hat{\vct \delta}$) and that optimized against the target (\ie, the white-box perturbation $\vct \delta$).
We observe immediately that gradient alignment provides an accurate measure of transferability: the higher the cosine similarity, the higher the correlation (meaning that the adversarial examples crafted against the two models are similar). We correlate these two measures in Fig.~\ref{fig:ev-scatter-mnist89-c}, and show that such correlation is statistically significant for both Pearson and Kendall coefficients. In Fig.~\ref{fig:ev-scatter-mnist89-d} we also correlate gradient alignment with the ratio between the test error of the target model in the black- and white-box setting (extrapolated from the matrix corresponding to $\varepsilon=1$ in the bottom row of Fig.~\ref{fig:ev-lk-mnist89}), as suggested by our theoretical derivation. The corresponding permutation tests confirm statistical significance.
We finally remark that gradient alignment is extremely fast to evaluate, as it does not require simulating any attack, but it is only a relative measure of the attack transferability, as the latter also depends on the complexity of the target model; \ie, on the size of its input gradients.

\begin{figure}[ht]
	\centering
	\resizebox{0.85\columnwidth}{!}{\begin{tabular}{cccc}
			\centering
			SVM$_{\mathrm L}$
			& SVM$_{\mathrm H}$
			& SVM-RBF$_{\mathrm L}$
			& SVM-RBF$_{\mathrm H}$
			\\
			$\varepsilon=1.7$
			&  $\varepsilon=0.45$
			& $\varepsilon=1.1$
			&  $\varepsilon=0.85$
			\\			
			\includegraphics[width=0.11\textwidth]{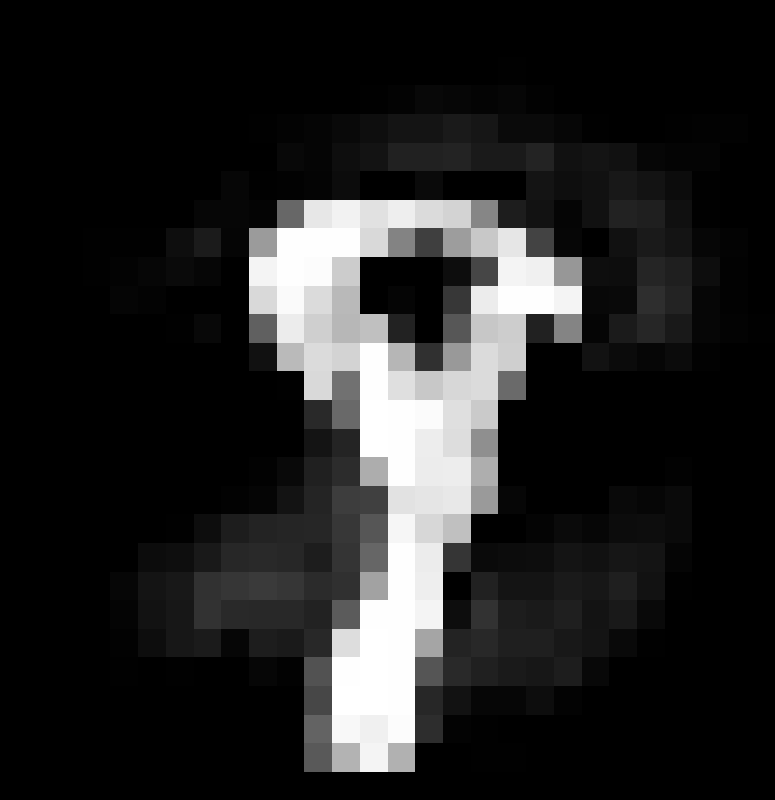} &
			\includegraphics[width=0.11\textwidth]{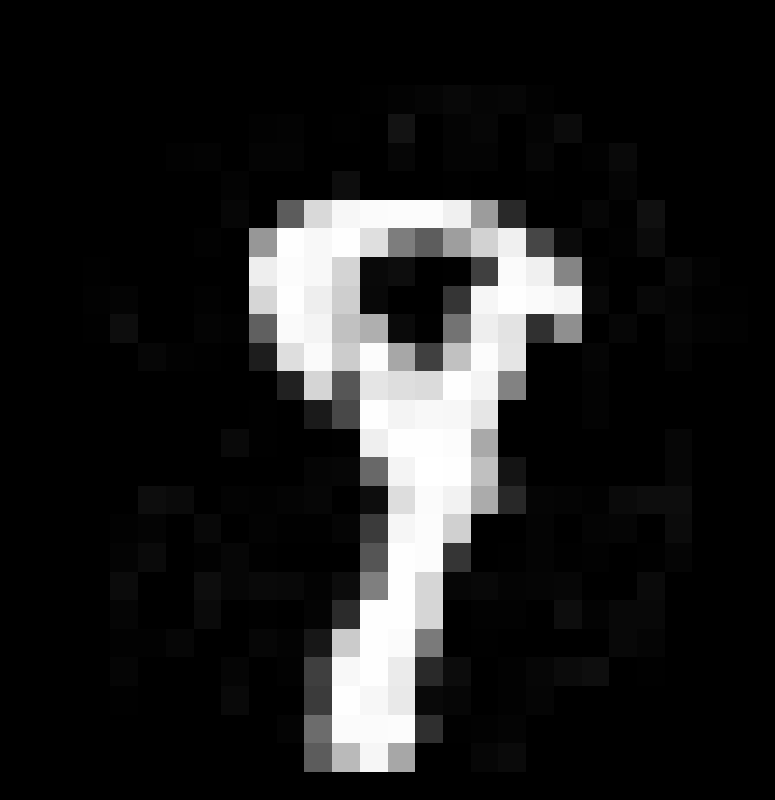} &
			\includegraphics[width=0.11\textwidth]{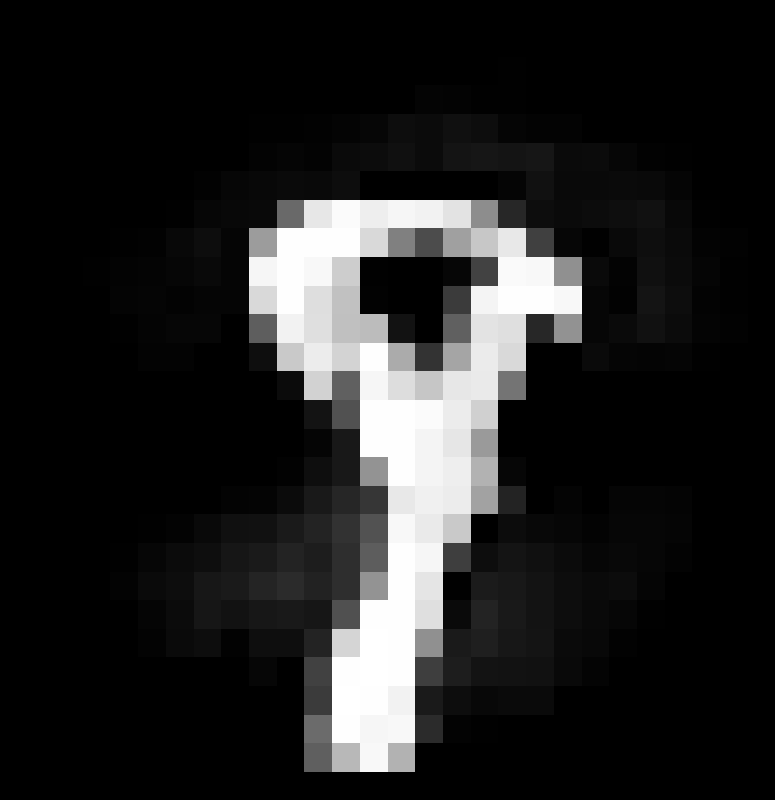} &
			\includegraphics[width=0.11\textwidth]{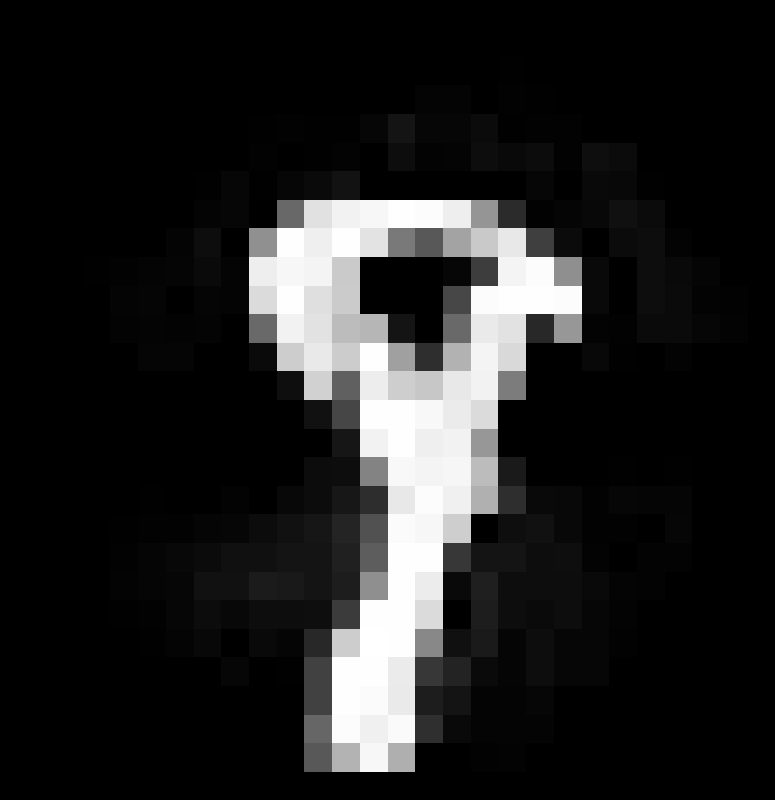}
			\vspace{0.25em}\\
			$\varepsilon=2.35$
			&$\varepsilon=0.95$
			& $\varepsilon=2.9$
			& $\varepsilon=2.65$
			\\ 					
			\includegraphics[width=0.11\textwidth]{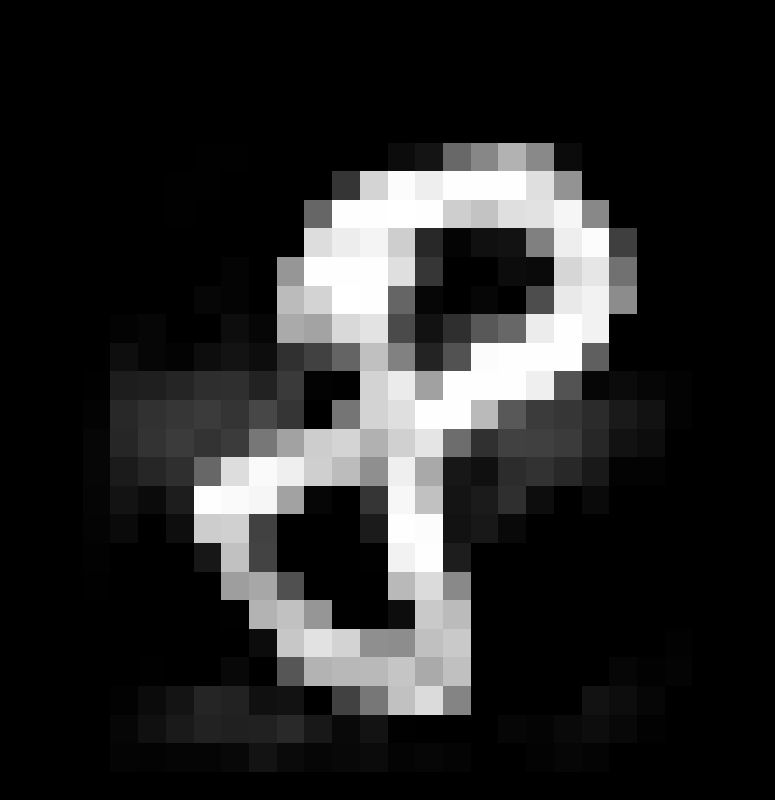} &
			\includegraphics[width=0.11\textwidth]{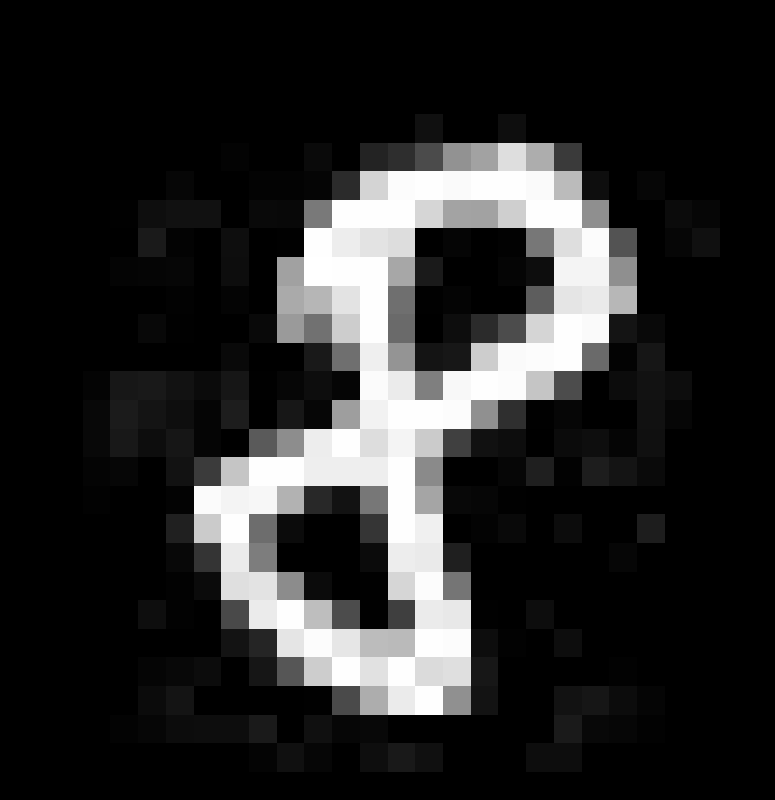} &
			\includegraphics[width=0.11\textwidth]{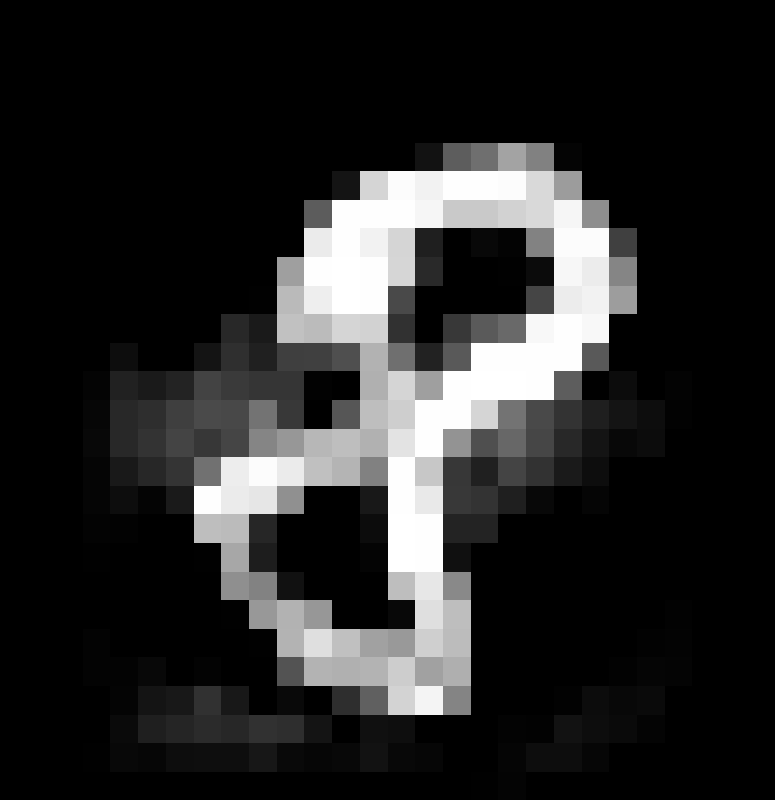} &
			\includegraphics[width=0.11\textwidth]{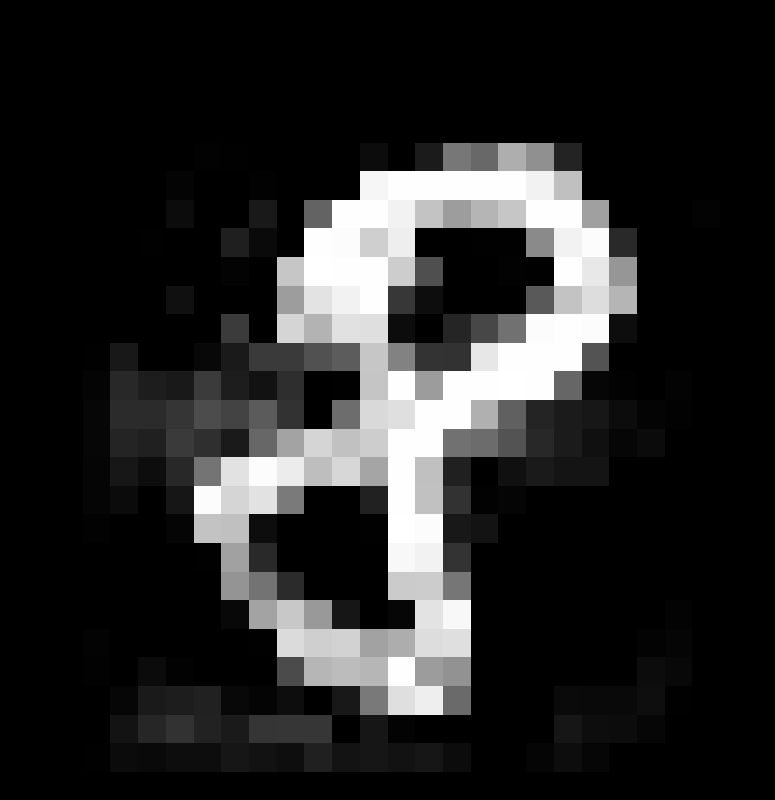}
		\end{tabular}}
		\caption{Digit images crafted to evade linear and RBF SVMs. The values of $\varepsilon$ reported here correspond to the minimum perturbation required to evade detection. Larger perturbations are required to mislead low-complexity classifiers (L), while smaller ones suffice to evade high-complexity classifiers (H).}
		\label{fig:evasion-digits}
	\end{figure}

	\subsubsection{Android Malware Detection}
	
	\begin{figure}
		\centering
		\includegraphics[width=.85\columnwidth]{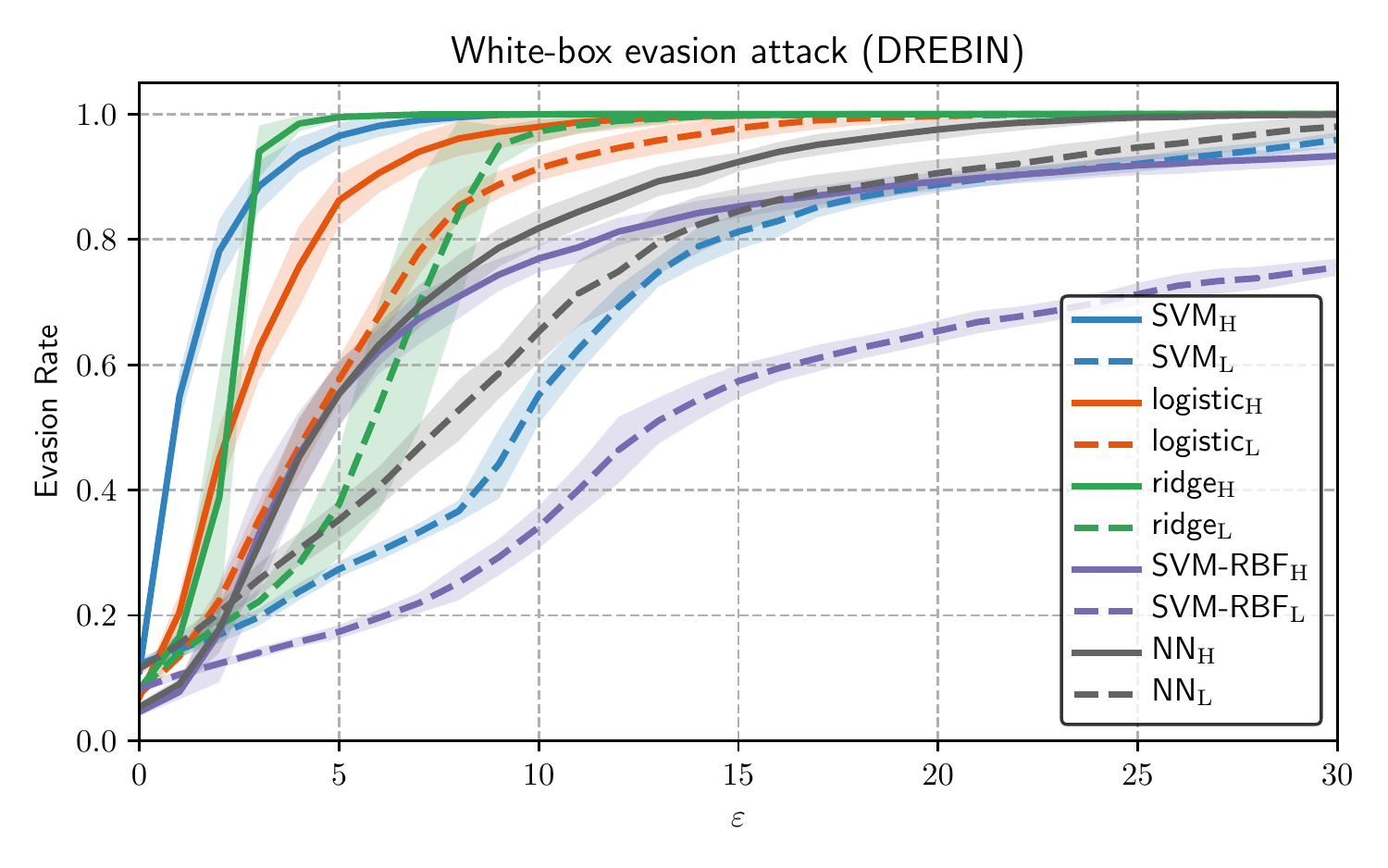}
		\vspace{-1em}
		\label{fig:pk-drebin}
		\caption{White-box evasion attacks on DREBIN. Evasion rate against increasing maximum perturbation $\varepsilon$.}
		\label{fig:ev-pk-drebin}
	\end{figure}
	
	\begin{figure*}[ht]
		\centering
		\begin{subfigure}[t]{0.22\textwidth}
			\includegraphics[width=\textwidth]{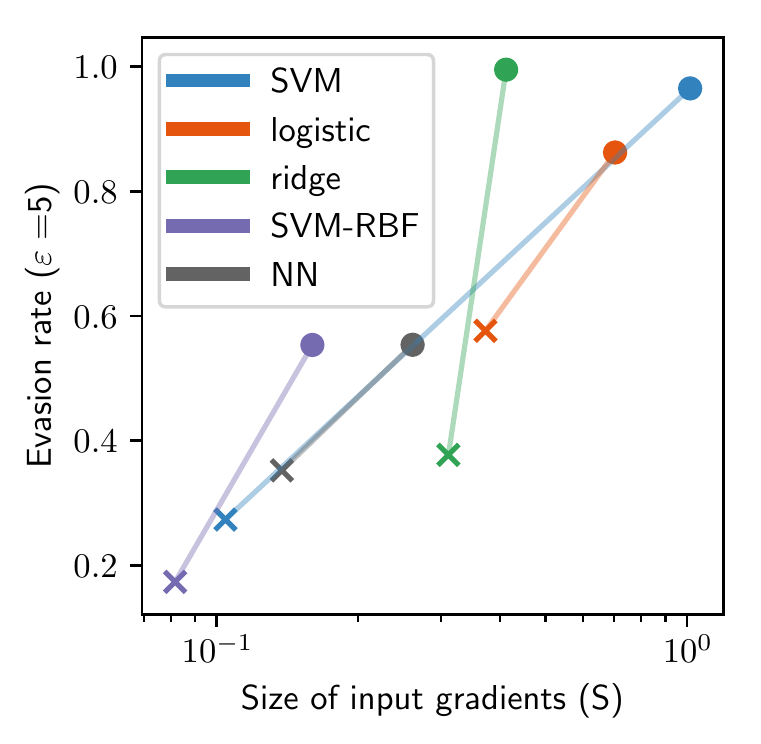}
			\subcaption{}
		\end{subfigure}
		\begin{subfigure}[t]{0.22\textwidth}
			\includegraphics[width=\textwidth]{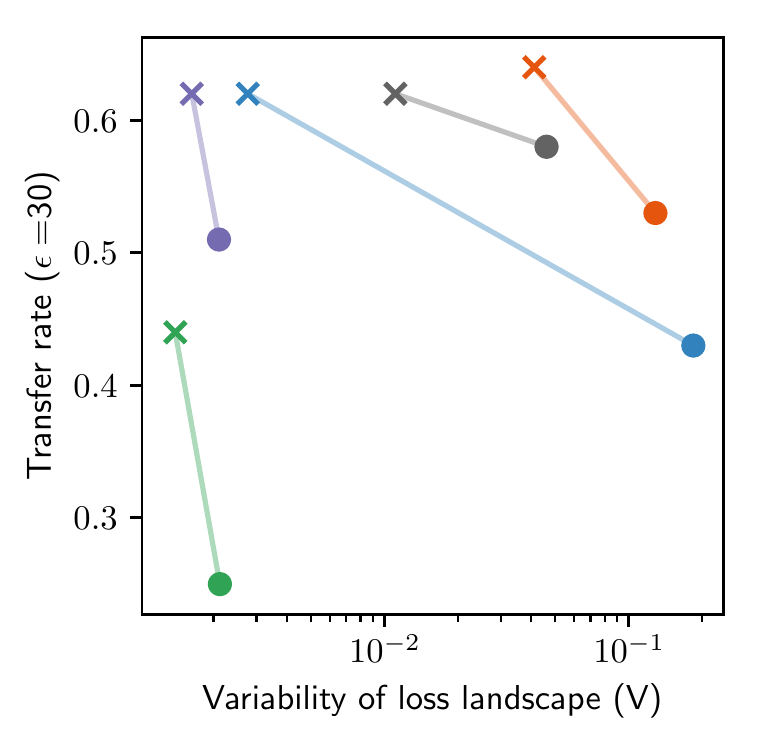}
			\subcaption{}
		\end{subfigure}
		\begin{subfigure}[t]{0.222\textwidth}
			\includegraphics[width=\textwidth]{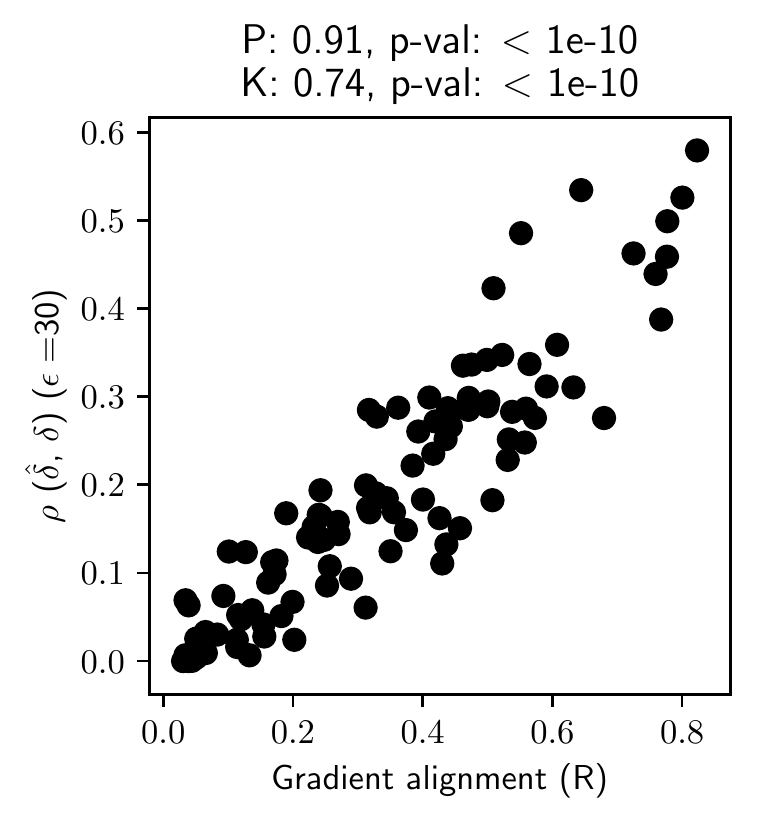}
			\subcaption{}
		\end{subfigure}
		\begin{subfigure}[t]{0.219\textwidth}
			\includegraphics[width=\textwidth]{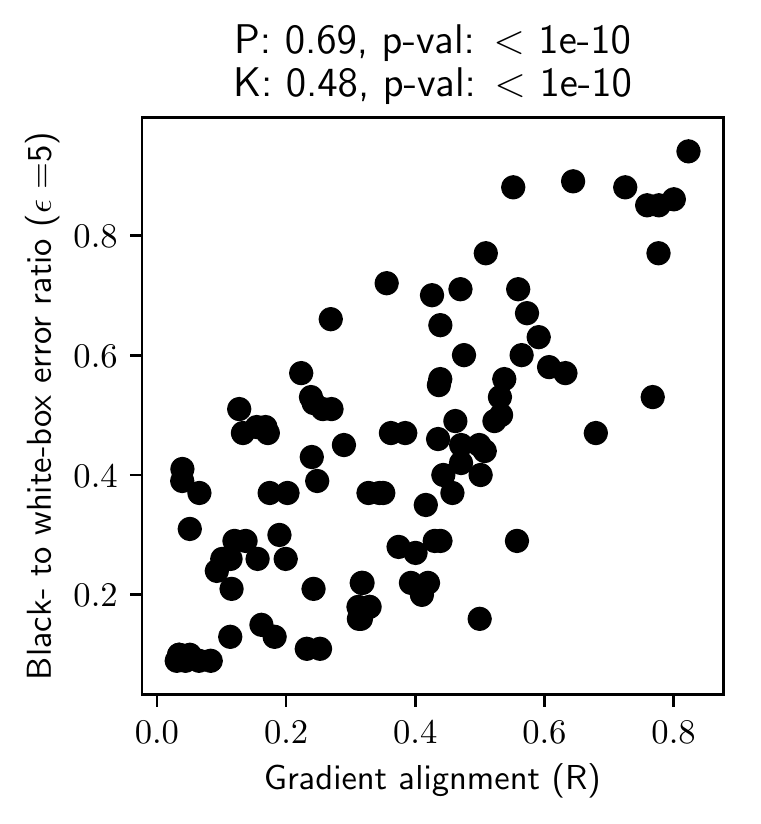}
			\subcaption{}
		\end{subfigure}
		\caption{Evaluation of our metrics for evasion attacks on DREBIN. See the caption of Fig.~\ref{fig:ev-scatter-mnist89} for further details.}
		\label{fig:ev-scatter-drebin}
	\end{figure*}
	
	\begin{figure*}[ht]
		\centering
		\includegraphics[width=.331\textwidth,trim={0 2cm 0 0},clip]{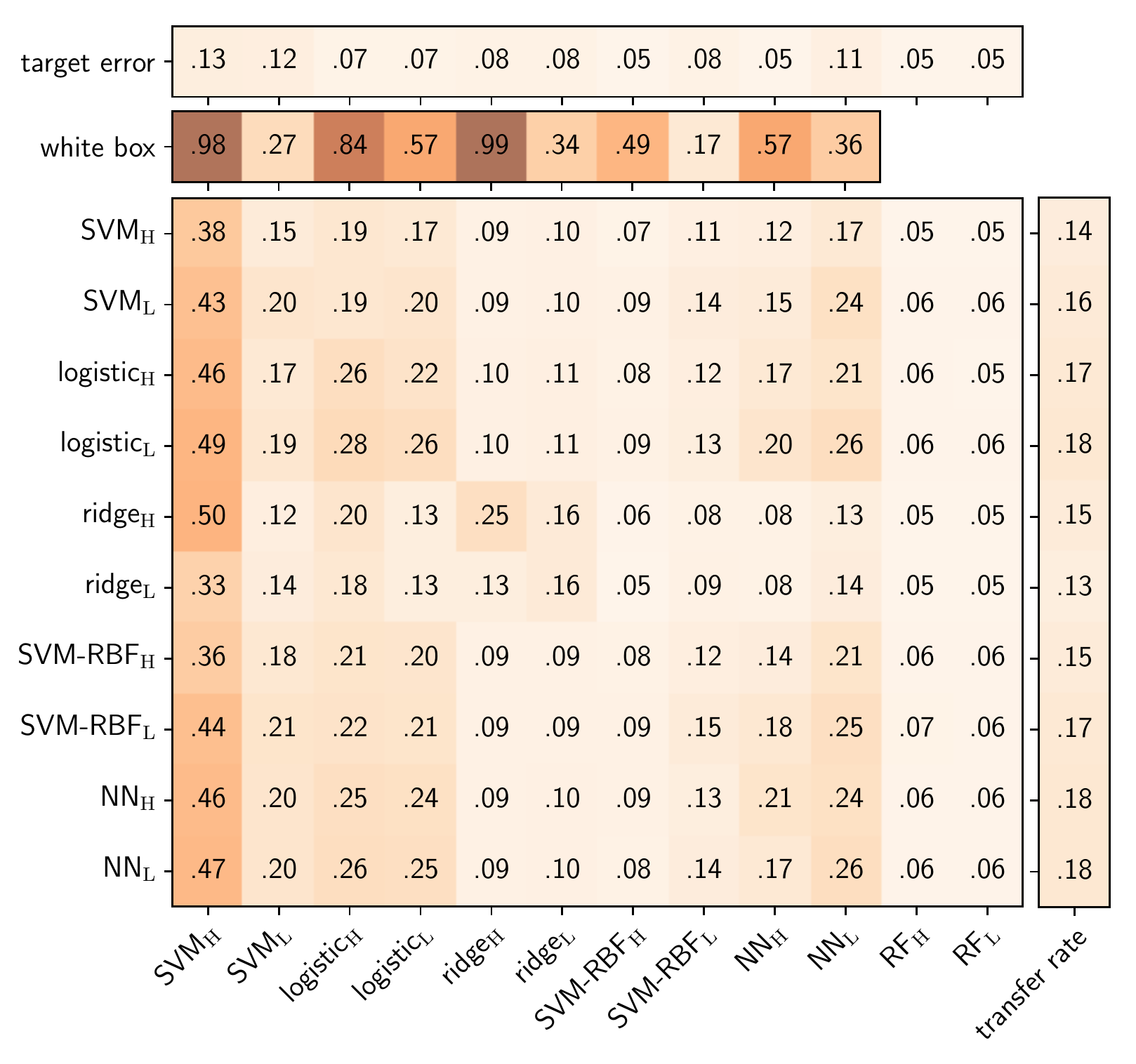}
		\includegraphics[width=.285\textwidth,trim={2.25cm 2cm 0 0},clip]{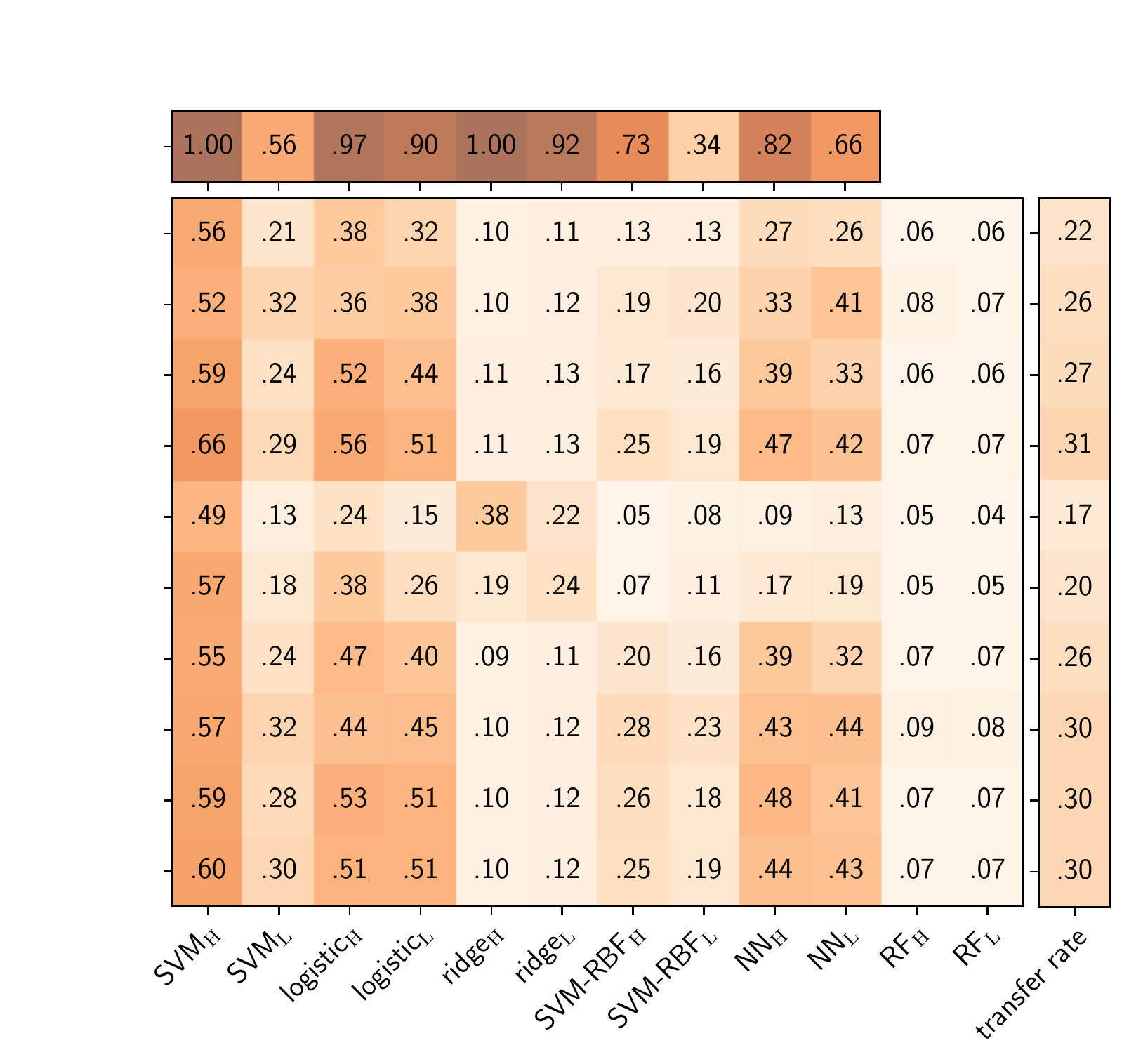}
		\includegraphics[width=.285\textwidth,trim={2.25cm 2cm 0 0},clip]{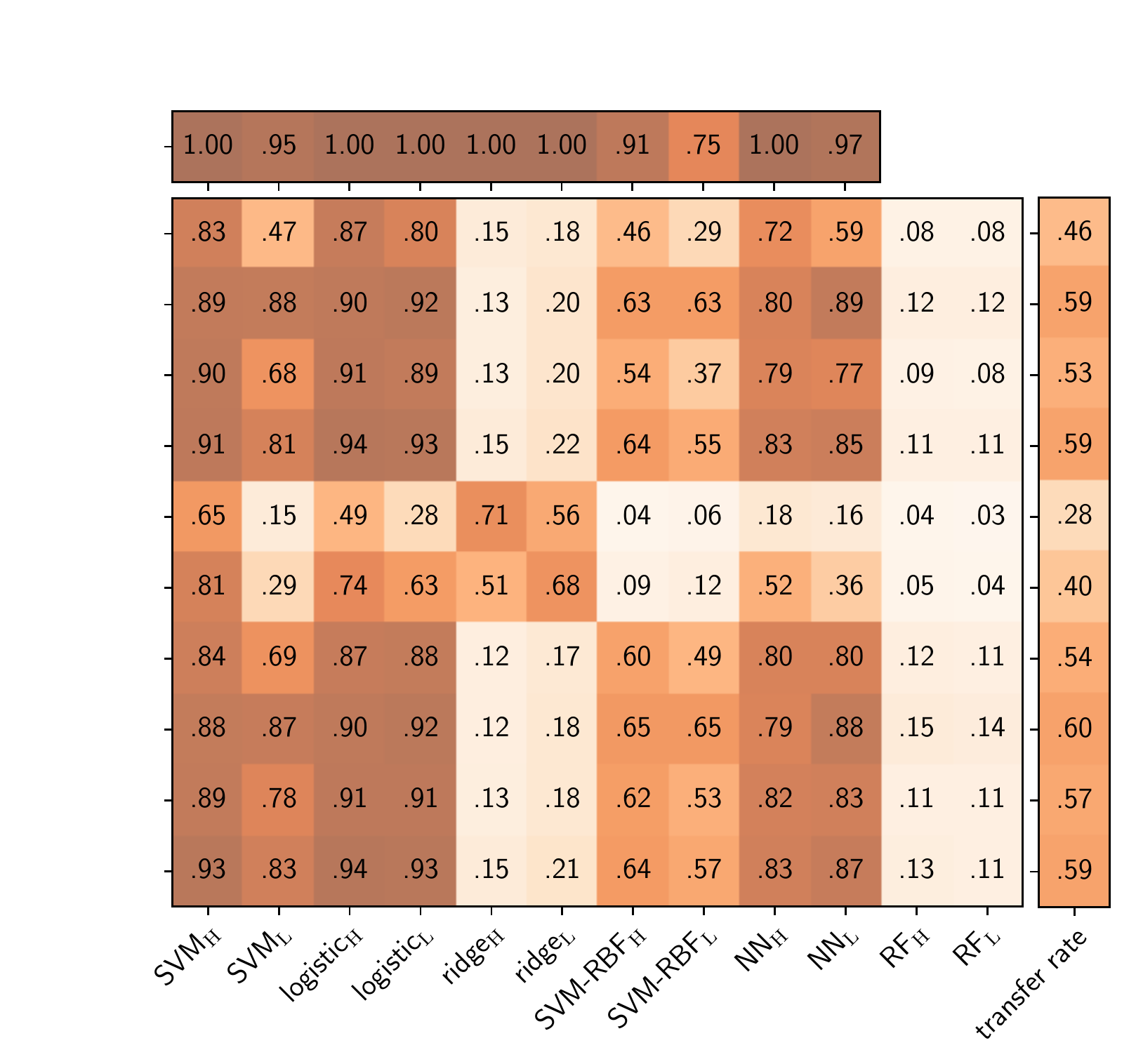}\vspace{.4em}\\
		\begin{subfigure}[t]{0.332\textwidth}
			\centering
			\includegraphics[width=\textwidth,trim={0 0 0 1.6cm},clip]{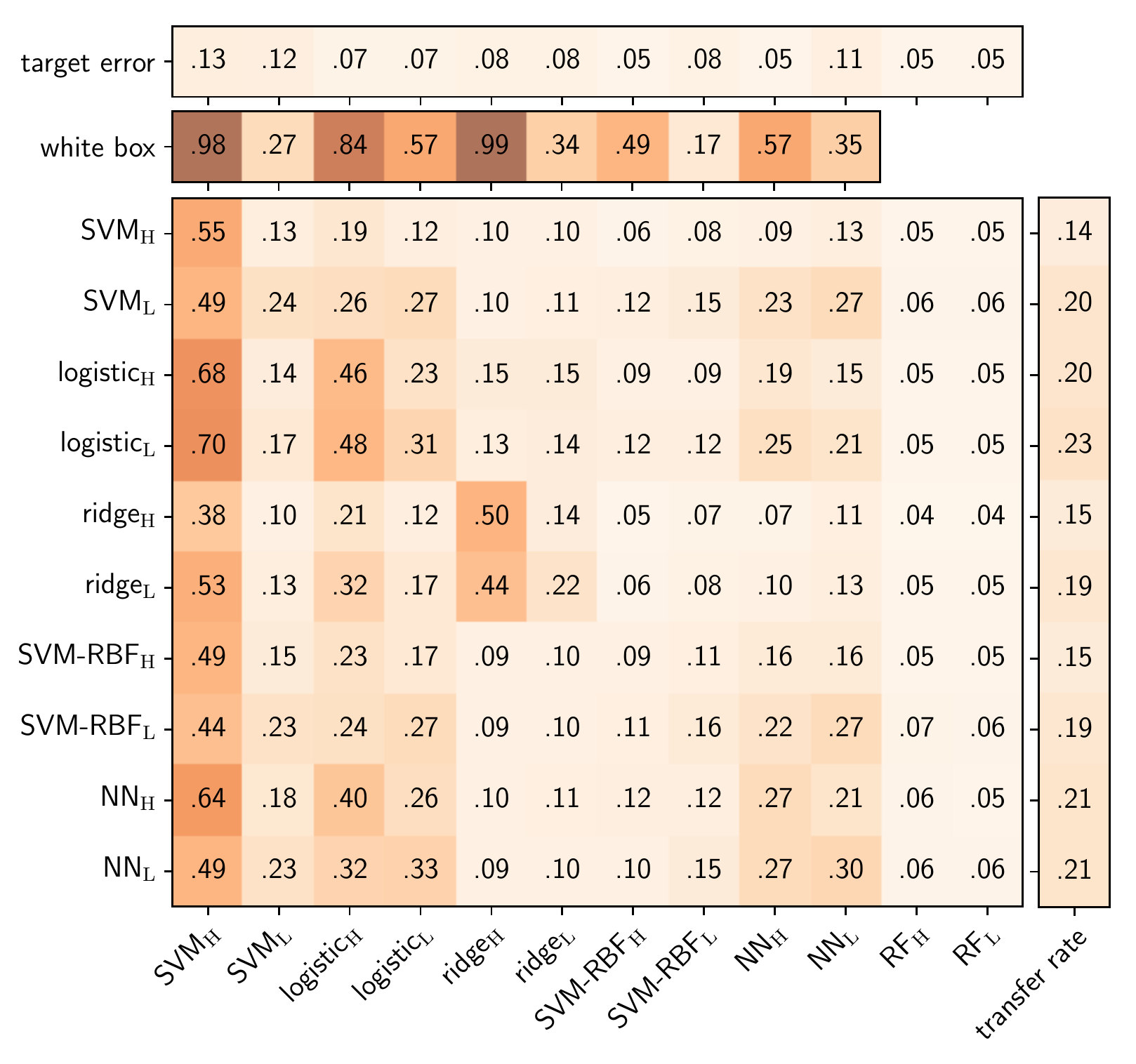}
			\subcaption{$\varepsilon = 5$}
		\end{subfigure}
		\begin{subfigure}[t]{0.285\textwidth}
			\centering
			\includegraphics[width=\textwidth,trim={2.25cm 0 0 1.6cm},clip]{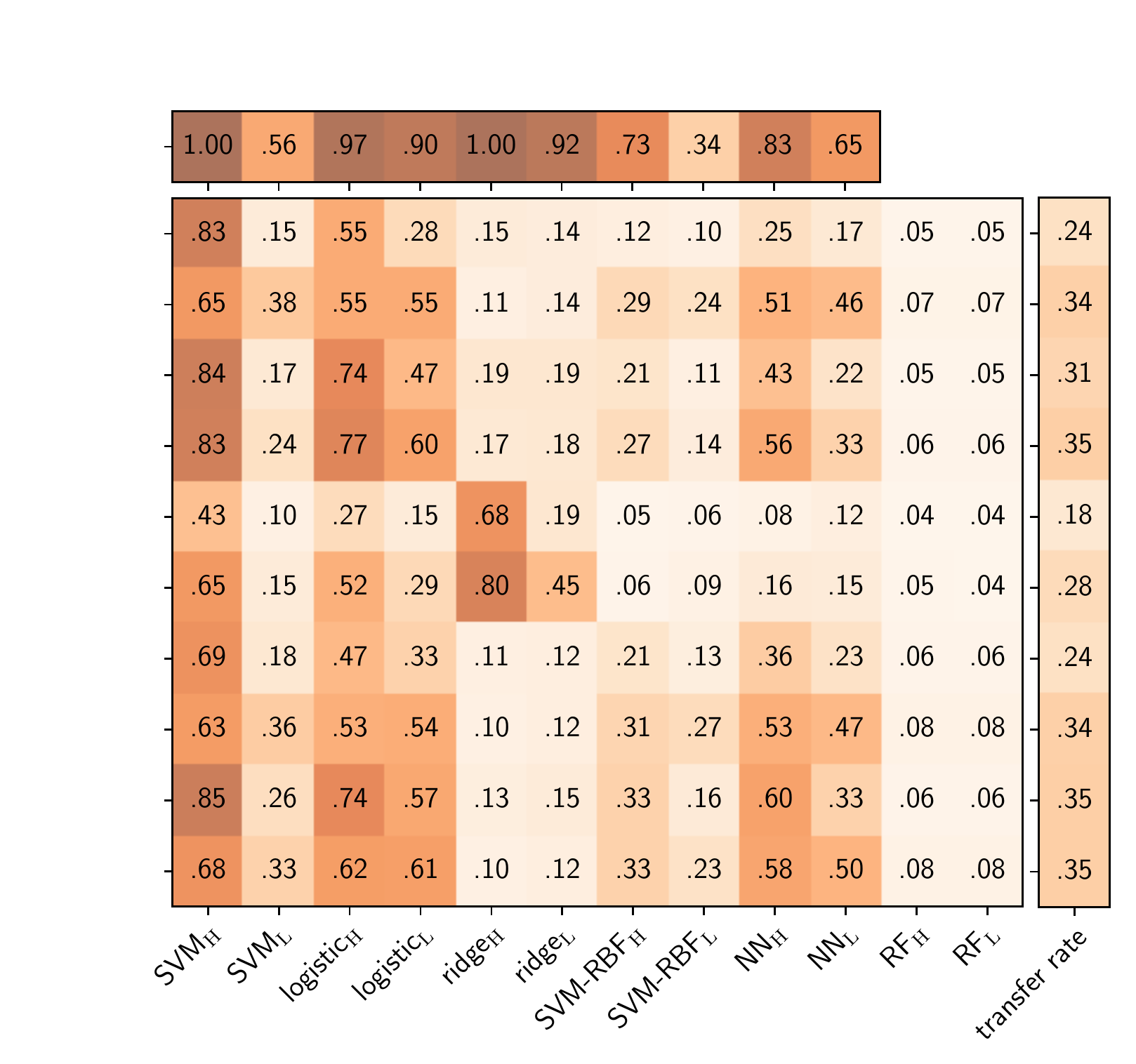}
			\subcaption{$\varepsilon = 10$}
		\end{subfigure}
		\begin{subfigure}[t]{0.285\textwidth}
			\centering
			\includegraphics[width=\textwidth,trim={2.25cm 0 0 1.6cm},clip]{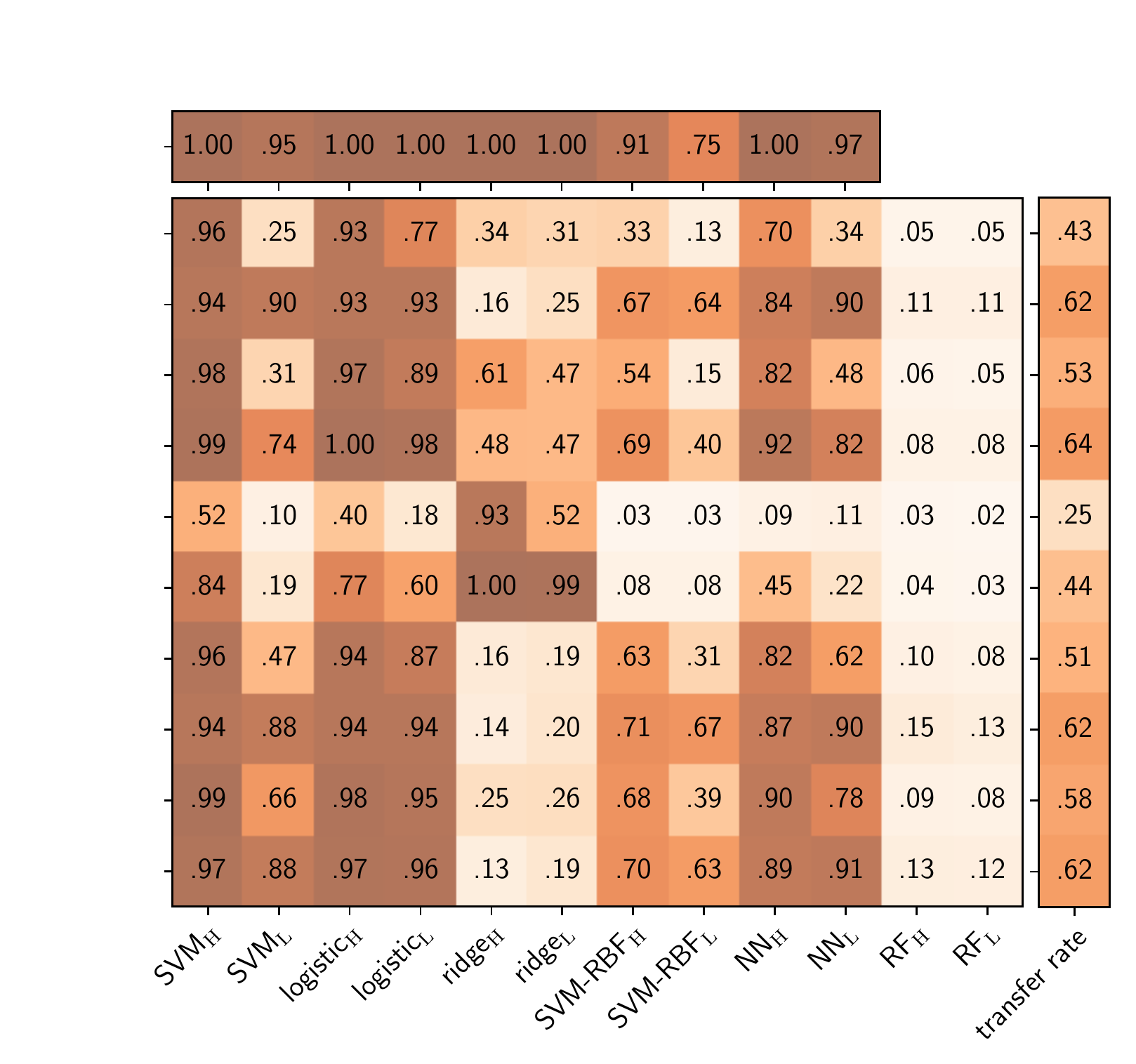}
			\subcaption{$\varepsilon = 30$}
		\end{subfigure}	
		\caption{Black-box (transfer) evasion attacks on DREBIN. See the caption of Fig.~\ref{fig:ev-lk-mnist89} for further details.}
		\label{fig:ev-lk-drebin}
	\end{figure*}
	
	\begin{figure}[ht]
		\centering
		\includegraphics[width=.54\columnwidth]{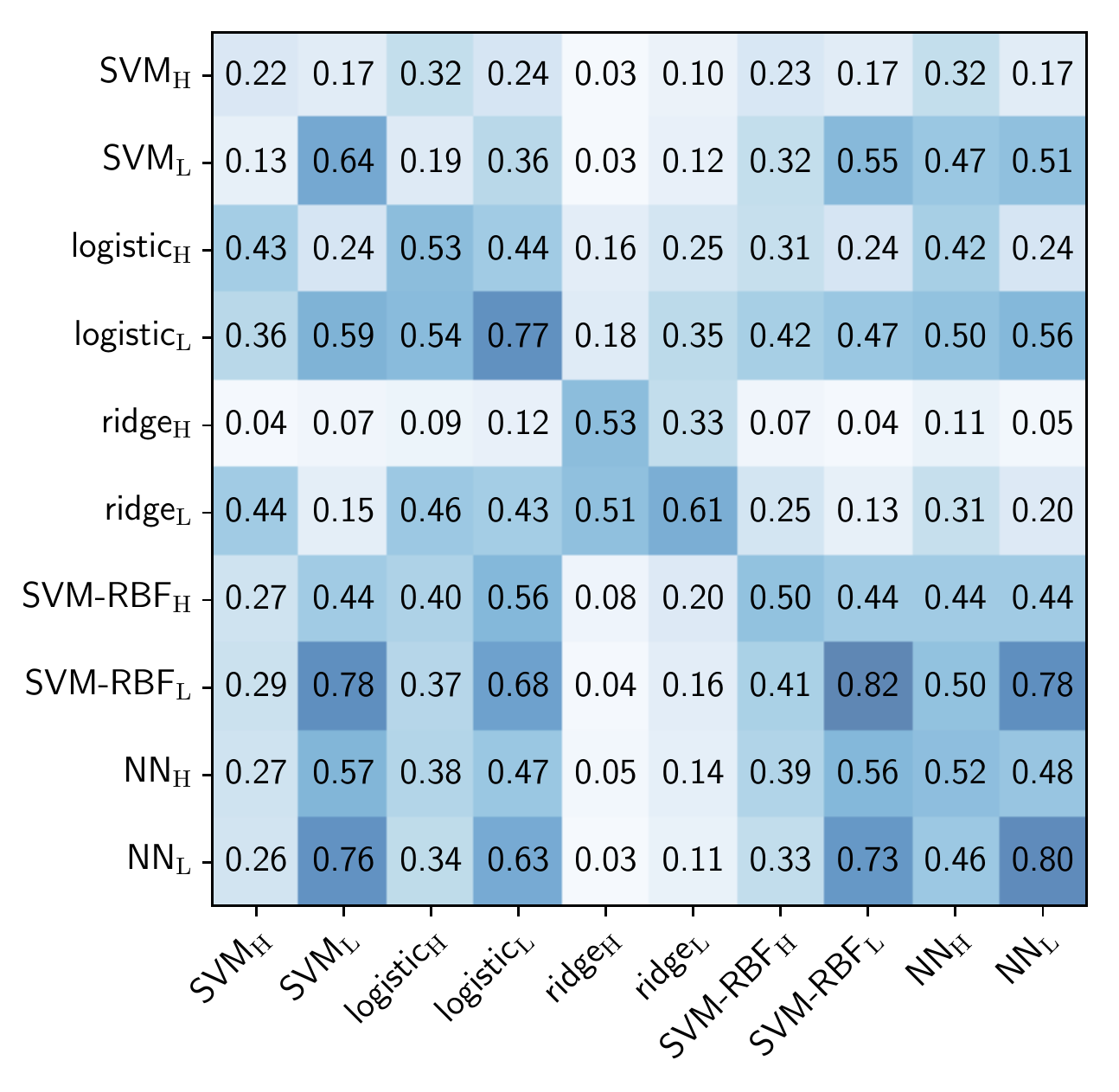}
		\includegraphics[width=.448\columnwidth,trim={2.24cm 0 0 0},clip]{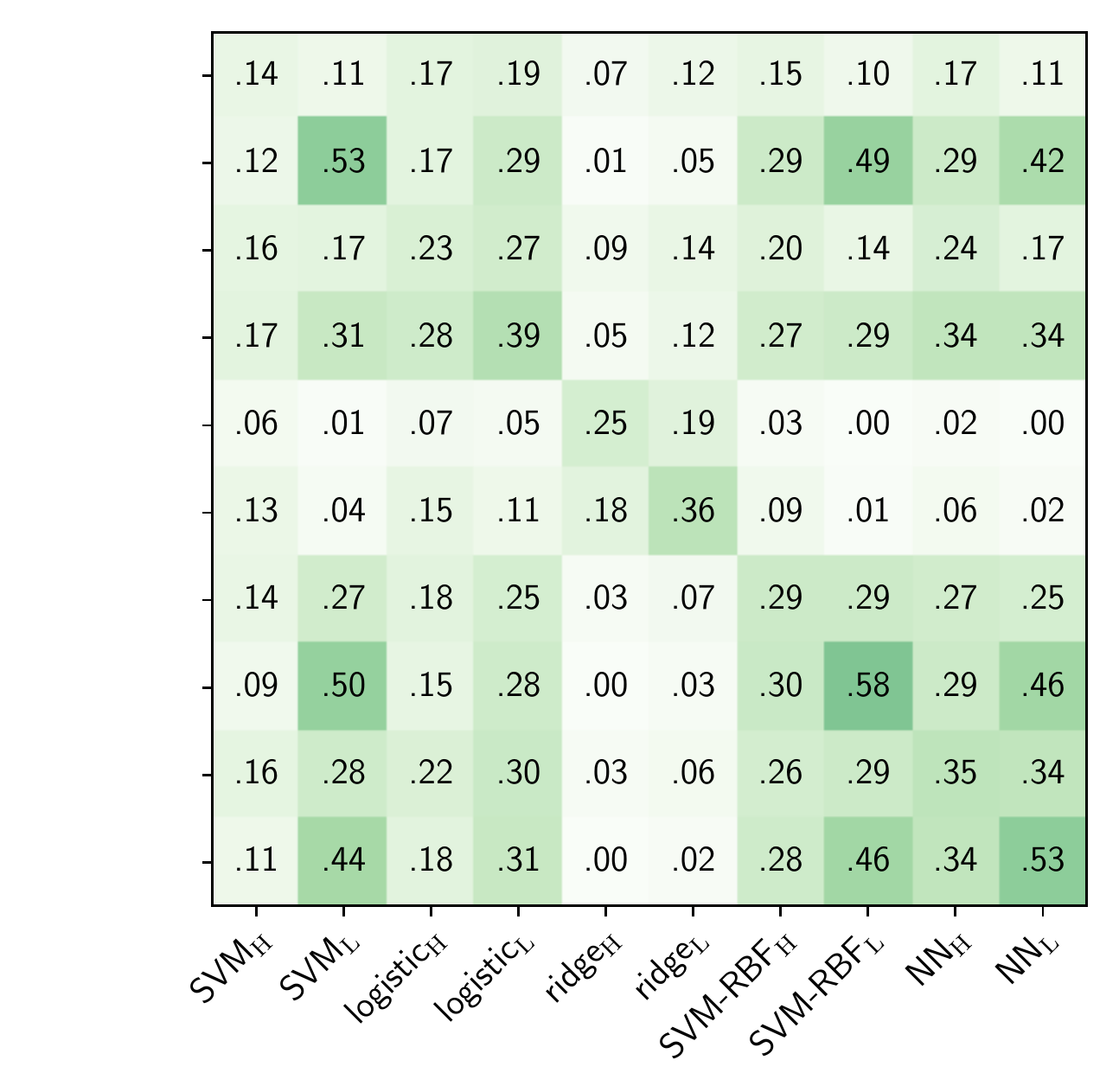}
		\vspace{-1.5em}
		\caption{Gradient alignment and perturbation correlation (at $\varepsilon=30$) for evasion attacks on DREBIN. See the caption of Fig.~\ref{fig:ev-corr-angles-mnist89} for further details.}
		\label{fig:ev-corr-angles-drebin}
	\end{figure}

	The Drebin data~\cite{rieck14-drebin} consists of around 120,000 legitimate and around 5000 malicious Android applications, labeled using the VirusTotal service. A sample is labeled as malicious (or positive, $y=+1$) if it is classified as such from at least five out of ten anti-virus scanners, while it is flagged as legitimate (or negative, $y=-1$) otherwise. The structure and the source code of each application is encoded as a \emph{sparse} feature vector consisting of around a million binary features denoting the presence or absence of permissions, suspicious URLs and other relevant information that can be extracted by statically analyzing Android applications. Since we are working with sparse binary features, we use the $\ell_1$ norm for the attack.
	
	We use $30,000$ samples to learn surrogate and target classifiers, and the remaining $66,944$ samples for testing. The classifiers and their hyperparameters are the same used for MNIST89, apart from ($i$) the number of hidden neurons for NN$_{\mathrm H}$ and NN$_{\mathrm L}$, set to $200$, ($ii$) the weight decay of NN$_{\mathrm L}$, set to $0.005$; and ($iii$) the maximum depth of RF$_{\mathrm L}$, set to $59$.
	
	We perform feature selection to retain those $5,000$ features which maximize information gain, \ie, $|p(x_k=1|y= +1)- p(x_k=1|y= -1)|$, where $x_k$ is the $k^{\rm th}$ feature. While this feature selection process does not significantly affect the detection rate (which is only reduced by $2\%$, on average, at $0.5\%$ false alarm rate), it drastically reduces the computational complexity of classification.
	
	In each experiment, we run white-box and black-box evasion attacks on $1,000$ distinct malware samples (randomly selected from the test data) against an increasing number of modified features in each malware $\varepsilon \in \{0, 1, 2, \ldots, 30\}$. This is achieved by imposing the $\ell_1$ constraint $\| \vct x^\prime - \vct x\|_1 \leq \varepsilon$. As in previous work, we further restrict the attacker to only \emph{inject} features into each malware sample, to avoid compromising its intrusive functionality~\cite{demontis17-tdsc,biggio13-ecml}.
	
	To evaluate the impact of the aforementioned evasion attack, we measure the evasion rate (\ie, the fraction of malware samples misclassified as legitimate) at $0.5\%$ false alarm rate (\ie, when only $0.5\%$ of the legitimate samples are misclassified as malware). As in the previous experiment, we report the complete \emph{security evaluation curve} for the white-box attack case, whereas we report only the value of test error for the black-box case. The results, reported in Figs.~\ref{fig:ev-pk-drebin},~\ref{fig:ev-scatter-drebin},~\ref{fig:ev-lk-drebin}, and~\ref{fig:ev-corr-angles-drebin}, along with the statistical tests in Table~\ref{tab:stats} (third and fourth columns) confirm the main findings of the previous experiments. One significant difference is that random forests are much more robust in this case. The reason is that the $\ell_1$-norm attack (differently from the $\ell_2$) only changes a small number of features, and thus the probability that it will change features in all the ensemble trees is very low.

\subsection{Transferability of Poisoning Attacks}

\begin{figure}[t]
	\centering
	\raisebox{0.02\height}{\includegraphics[width=.75\columnwidth,trim={0 0.2cm 0 0},clip]{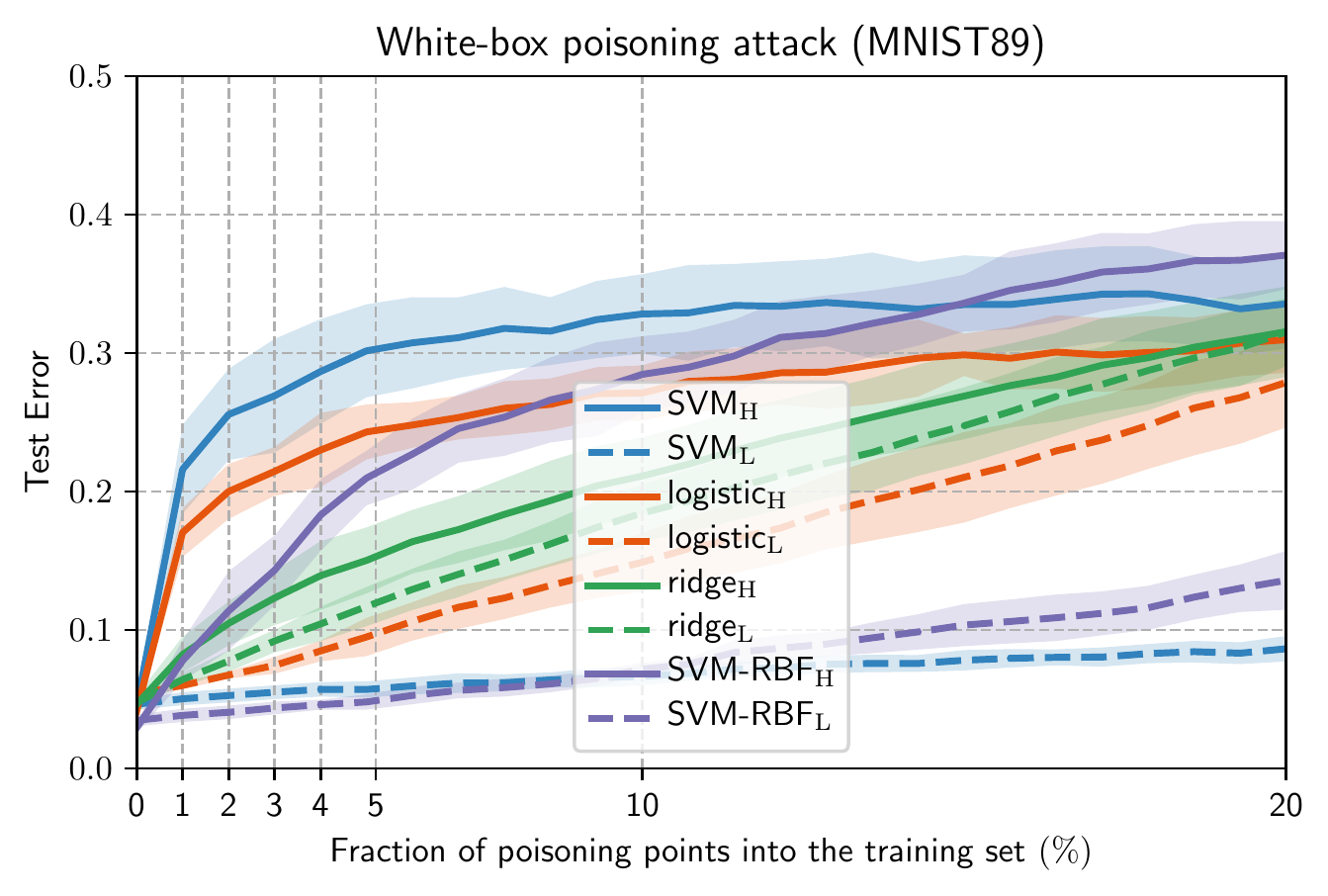}}
	\vspace{-0.5em}
	\caption{White-box poisoning attacks on MNIST89. Test error against an increasing fraction of poisoning points.}
	\label{fig:pois-pk-mnist89}
\end{figure}

\begin{figure*}[t]
	\centering
	\begin{subfigure}[t]{0.22\textwidth}
		\includegraphics[width=\textwidth]{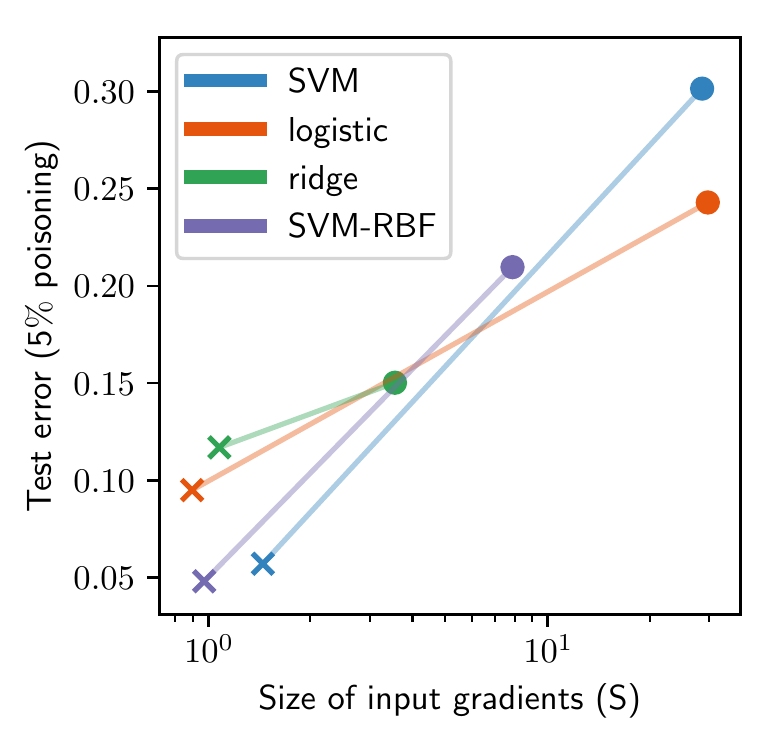}
		\caption{} \label{fig:pois-scatter-mnist89-a}
	\end{subfigure}
	\begin{subfigure}[t]{0.22\textwidth}
		\includegraphics[width=\textwidth]{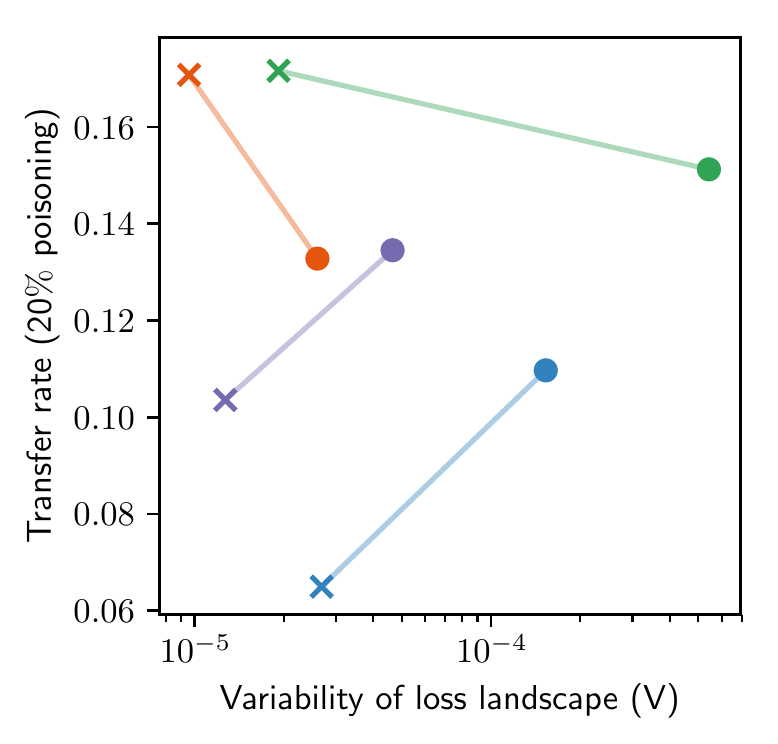}
		\caption{} \label{fig:pois-scatter-mnist89-b}
	\end{subfigure}
	\begin{subfigure}[t]{0.217\textwidth}
		\includegraphics[width=\textwidth]{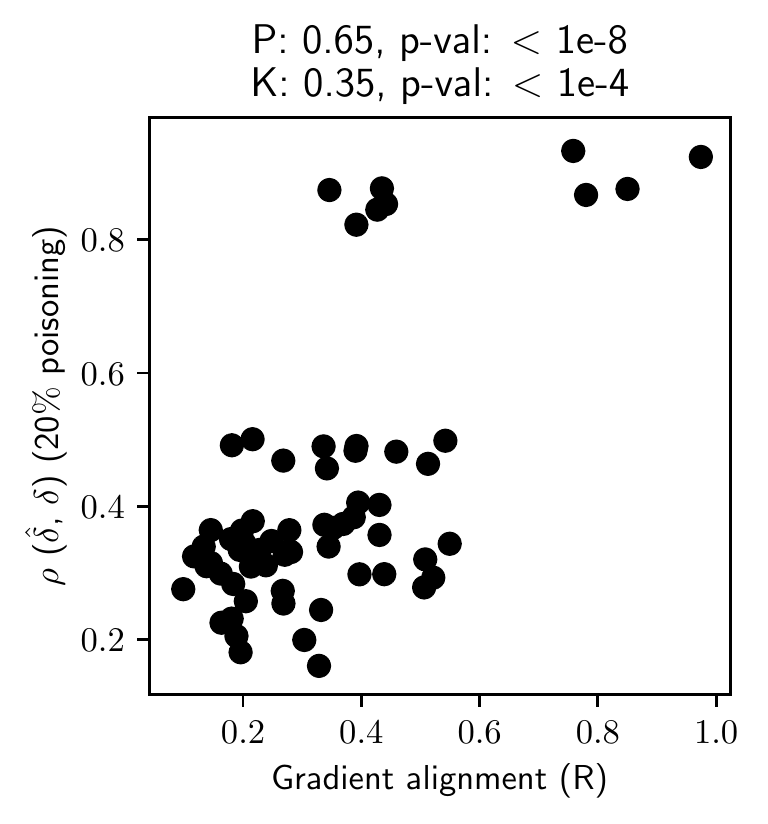}
		\caption{} \label{fig:pois-scatter-mnist89-c}
	\end{subfigure}
	\begin{subfigure}[t]{0.215\textwidth}
		\includegraphics[width=\textwidth]{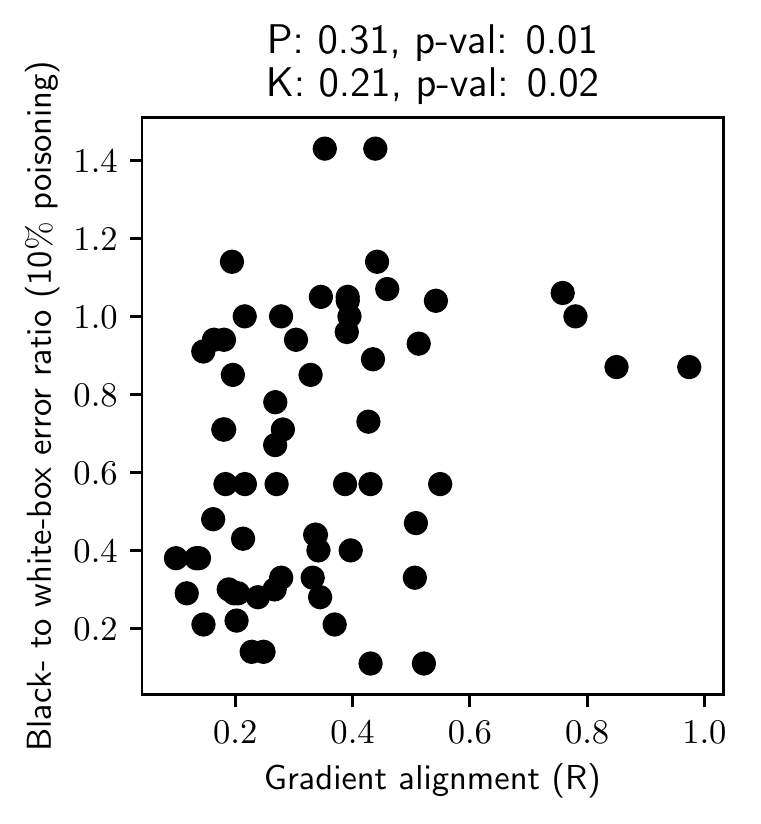}
		\caption{} \label{fig:pois-scatter-mnist89-d}
	\end{subfigure}
	\caption{Evaluation of our metrics for poisoning attacks on MNIST89. See the caption of Fig.~\ref{fig:ev-scatter-mnist89} for further details.}
	\label{fig:pois-scatter-mnist89}
\end{figure*}

\begin{figure*}[t]
	\centering
	\begin{subfigure}[t]{0.331\textwidth}
		\centering
		\includegraphics[width=\textwidth,trim={0 0 0 0},clip]{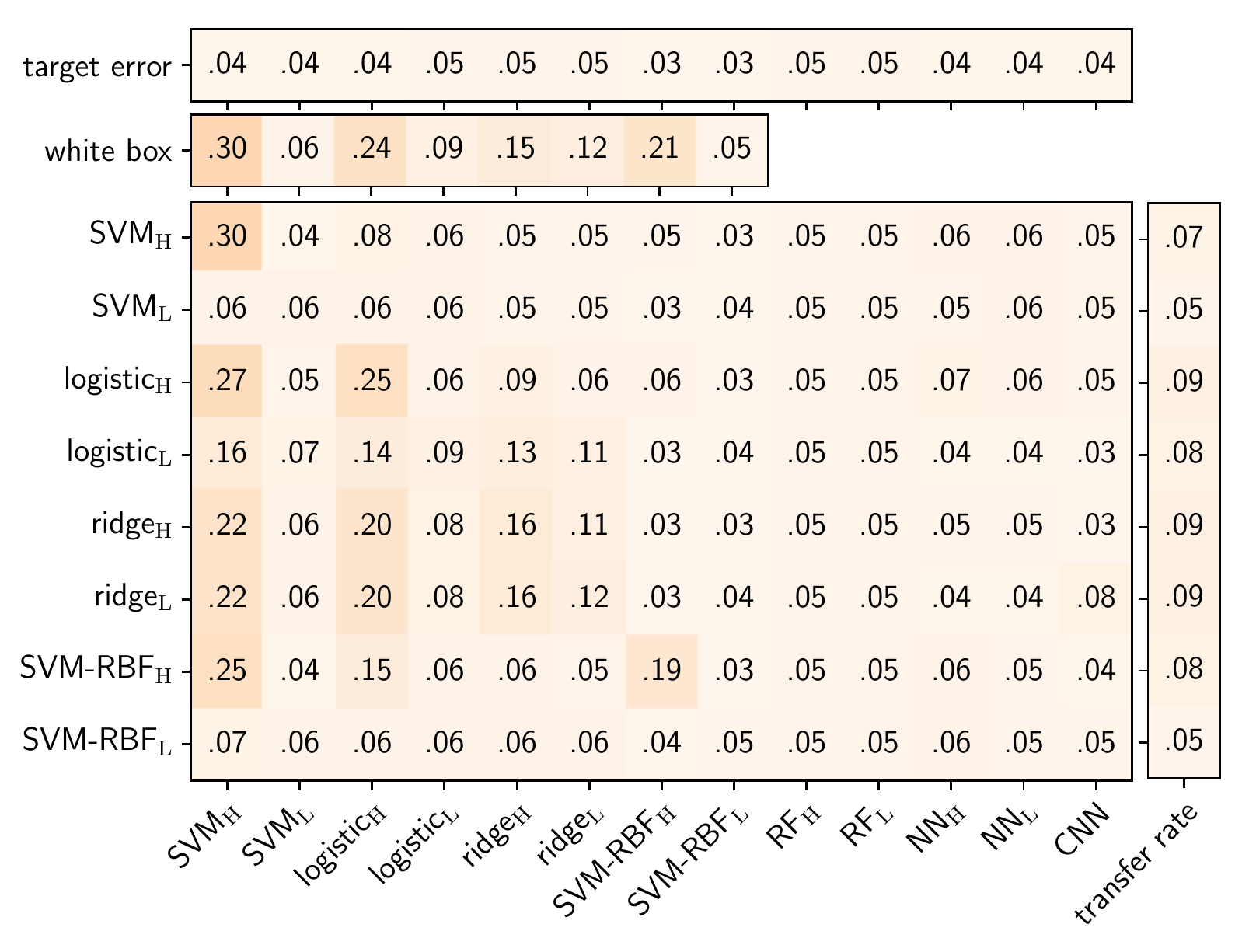}
		\subcaption{5\% poisoning}
	\end{subfigure}
	\begin{subfigure}[t]{0.285\textwidth}
		\centering
		\includegraphics[width=\textwidth,trim={2.21cm 0 0 0},clip]{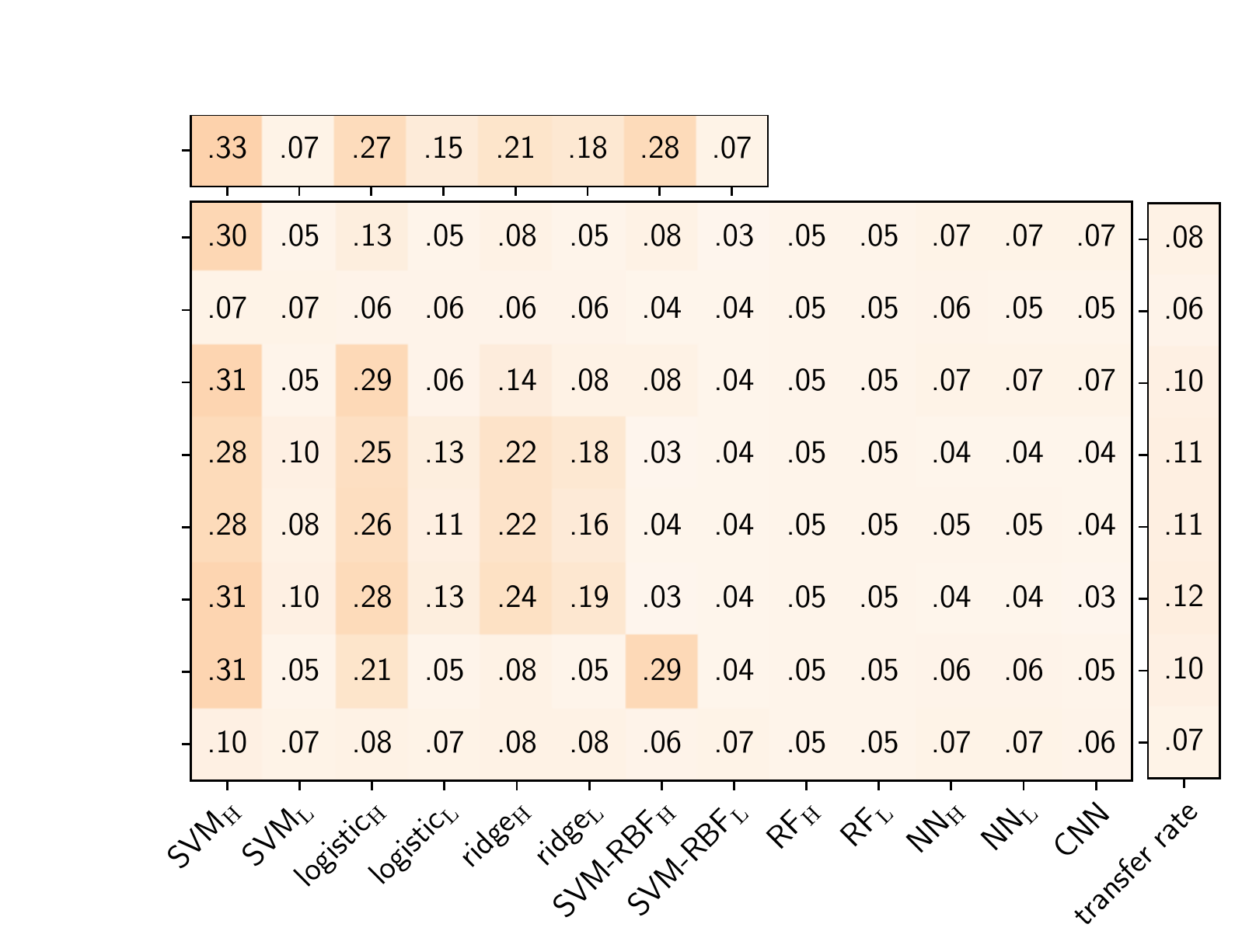}
		\subcaption{10\% poisoning}
	\end{subfigure}
	\begin{subfigure}[t]{0.285\textwidth}
		\centering
		\includegraphics[width=\textwidth,trim={2.20cm 0 0 0},clip]{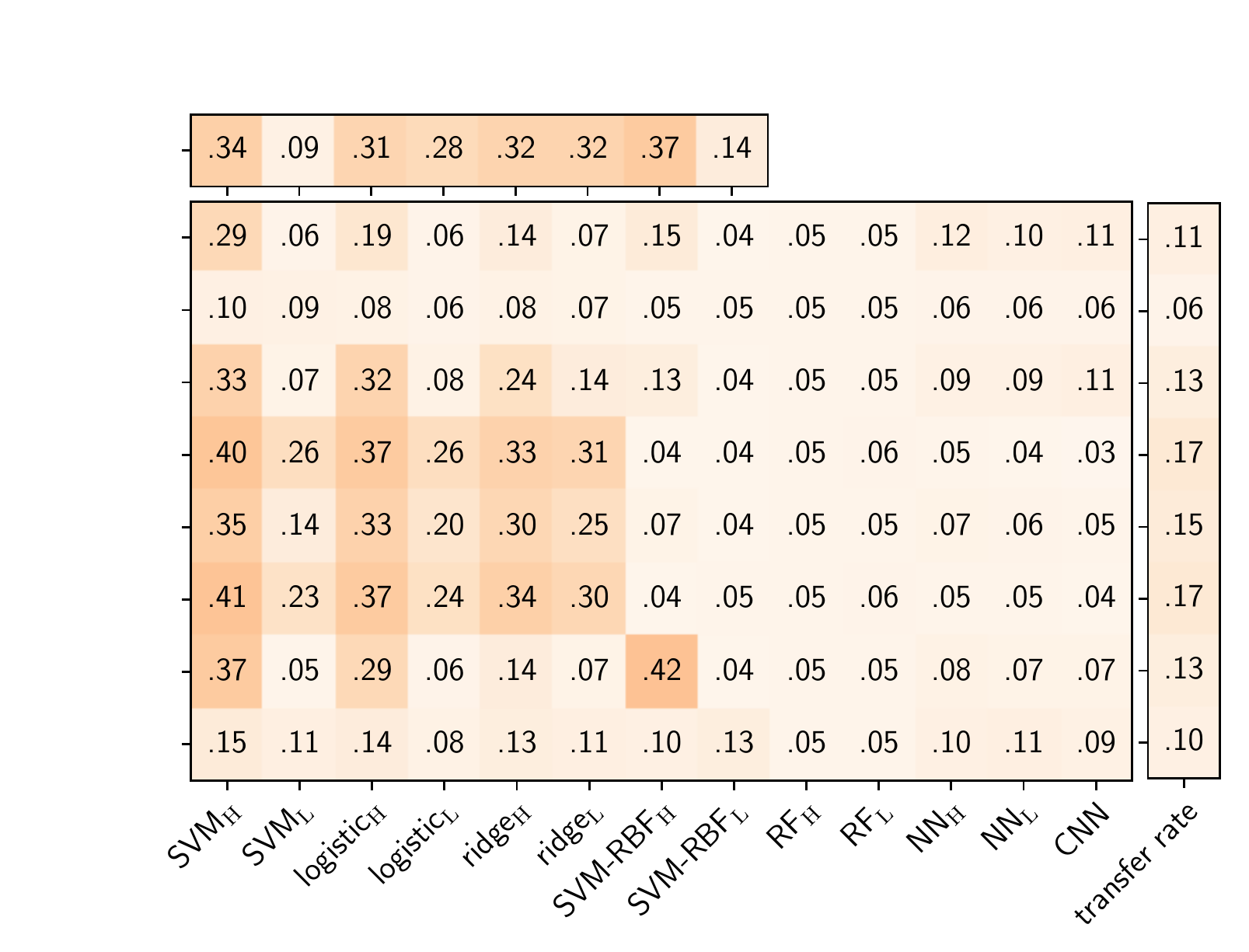}
		\subcaption{20\% poisoning}
	\end{subfigure}
	\caption{Black-box (transfer) poisoning attacks on MNIST89. See the caption of Fig.~\ref{fig:ev-lk-mnist89} for further details.}
	\label{fig:pois-lk-mnist}
\end{figure*}

For poisoning attacks, we report experiments on handwritten digits and face recognition.

\subsubsection{Handwritten Digit Recognition} We apply our optimization framework to poison SVM, logistic, and ridge classifiers in the white-box setting. Designing efficient poisoning availability attacks against neural networks is still an open problem due to the complexity of the bilevel optimization and the non-convexity of the inner learning problem.  Previous work has mainly considered integrity poisoning attacks against neural networks~\cite{Koh17,suciu18-usenix,biggio18}, and it is believed that neural networks are much more resilient to poisoning availability attacks due to their memorization capability. Poisoning random forests is not feasible with gradient-based attacks, and we are not aware of any existing attacks for this ensemble method.
We thus consider as surrogate learners: ($i$) linear SVMs with
$C=0.01$ (SVM$\rm{_{L}}$) and $C=100$ (SVM$\rm{_{H}}$); ($ii$) logistic classifiers with $C=0.01$ (logistic$\rm{_{L}}$) and $C=10$ (logistic$\rm{_{H}}$);  ($iii$) ridge classifiers with $\alpha = 100$ (ridge$\rm{_{L}}$) and $\alpha = 10 $ (ridge$\rm{_{H}}$); and ($iv$) SVMs with RBF kernel with $\gamma = 0.01$ and $C=1$ (SVM-RBF$\rm{_{L}}$) and $C=100$ (SVM-RBF$\rm{_{H}}$).
We additionally consider as target classifiers: ($i$) random forests with $100$ base trees, each with a maximum depth of 6 for RF$\rm{_{L}}$, and with no limit on the maximum depth for RF$\rm{_{H}}$; ($ii$) feed-forward neural networks with two hidden layers of 200 neurons each and ReLU activations, trained via cross-entropy loss minimization with different regularization (NN$\rm{_{L}}$ with weight decay $0.01$ and NN$\rm{_{H}}$ with no decay); and ($iii$) the Convolutional Neural Network (CNN) used in~\cite{carlini17-aisec}.

We consider $500$ training samples, $1,000$  validation samples to compute the attack, and a separate set of $1,000$ test samples to evaluate the error.
The test error is computed against an increasing number of poisoning points into the training set, from $0\%$ to $20\%$ (corresponding to 125 poisoning points).
The reported results are averaged on $10$ independent, randomly-drawn data splits.

\noindent {\bf How does model complexity impact poisoning attack success in the white-box setting?} The results for white-box poisoning are reported in Fig.~\ref{fig:pois-pk-mnist89}.
Similarly to the evasion case, high-complexity models (with larger input gradients, as shown in Fig.~\ref{fig:pois-scatter-mnist89-a}) are more vulnerable to poisoning attacks than their low-complexity counterparts (\ie, given that the same learning algorithm is used). This is also confirmed by the statistical tests in the fifth column of Table~\ref{tab:stats}. Therefore, model complexity plays a large role in a model's robustness also against poisoning attacks, confirming our analysis.

\begin{figure}[t]	
	\centering
	\includegraphics[height=.24\textwidth]{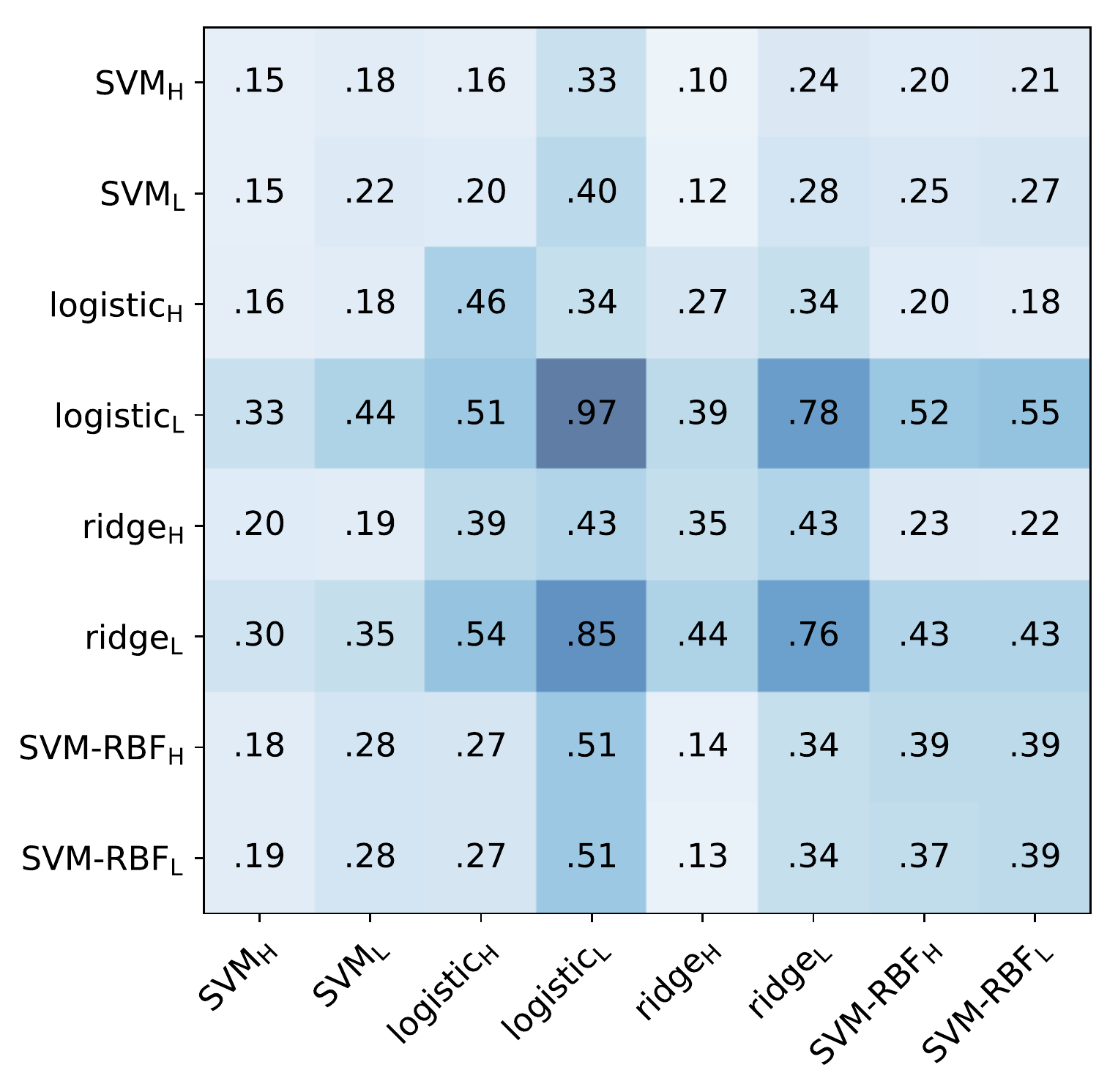}
	\raisebox{-0.01\height}{\includegraphics[height=.242\textwidth]{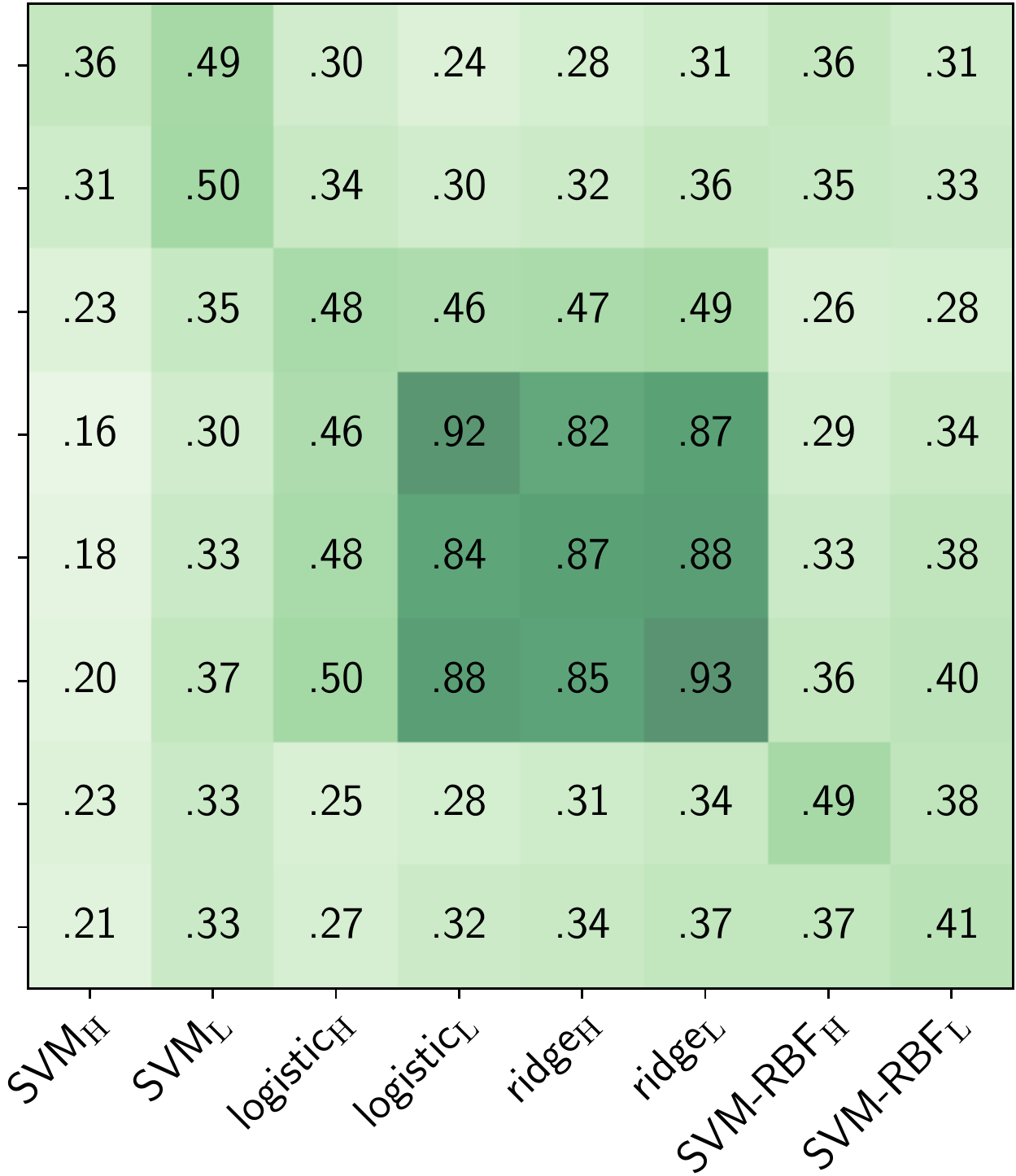}}
	\vspace{-0.5em}
	\caption{Gradient alignment and perturbation correlation (at 20\% poisoning) for poisoning attacks on MNIST89. See the caption of Fig.~\ref{fig:ev-corr-angles-mnist89} for further details.}
	\label{fig:pois-corr-angles-mnist89}
\end{figure}

\begin{figure}[ht]
	\centering
	\resizebox{0.85\columnwidth}{!}{
		\begin{tabular}{cccc}
			SVM$_{\mathrm L}$ & SVM$_{\mathrm H}$  & SVM-RBF$_{\mathrm L} $ & SVM-RBF$_{\mathrm H}$\\
			\includegraphics[width=0.11\textwidth]{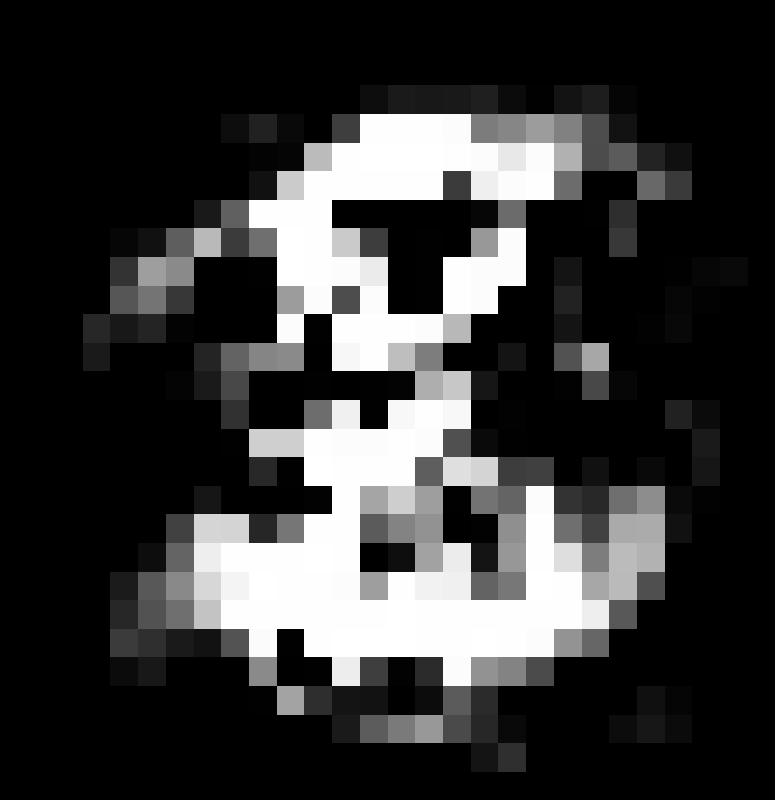}&
			\includegraphics[width=0.11\textwidth]{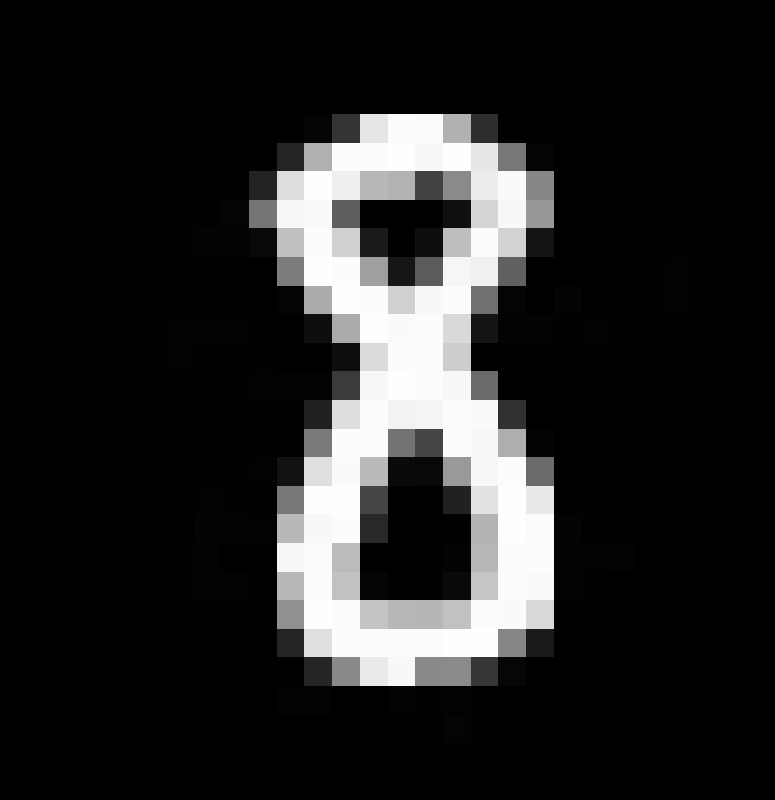}&
			\includegraphics[width=0.11\textwidth]{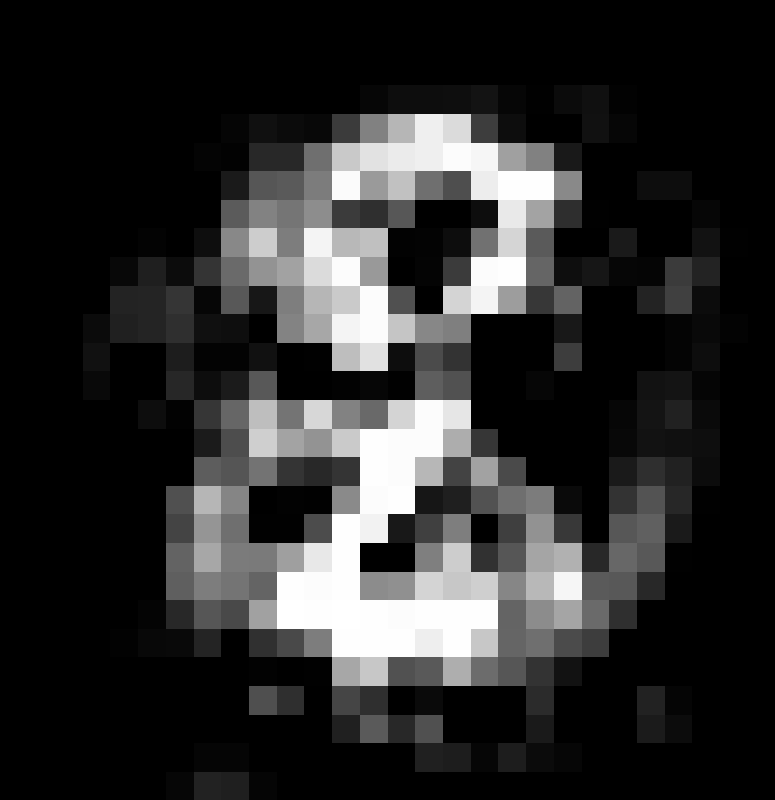}&
			\includegraphics[width=0.11\textwidth]{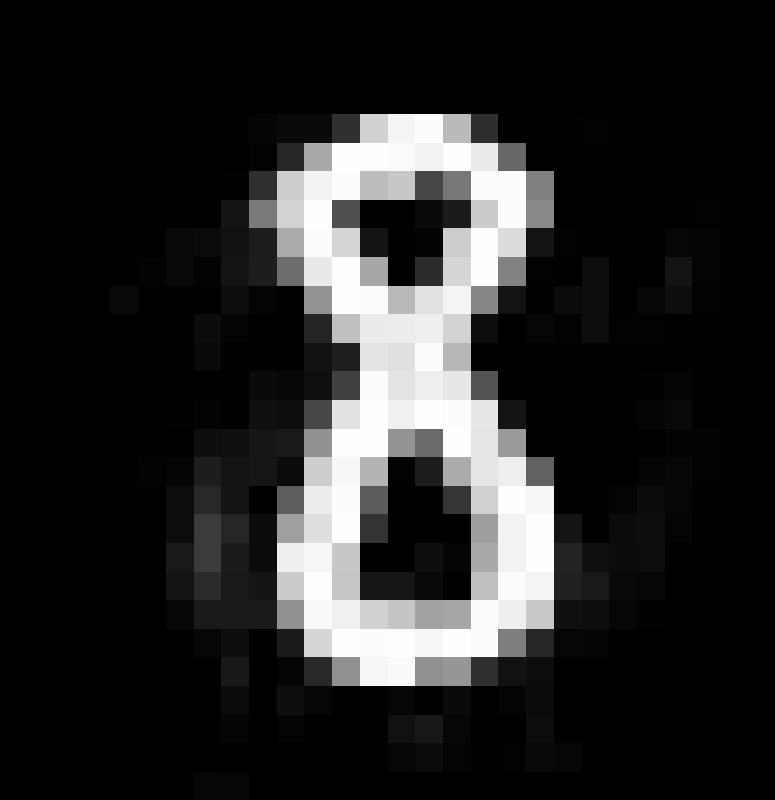} \vspace{0.5em}\\
			\includegraphics[width=0.11\textwidth]{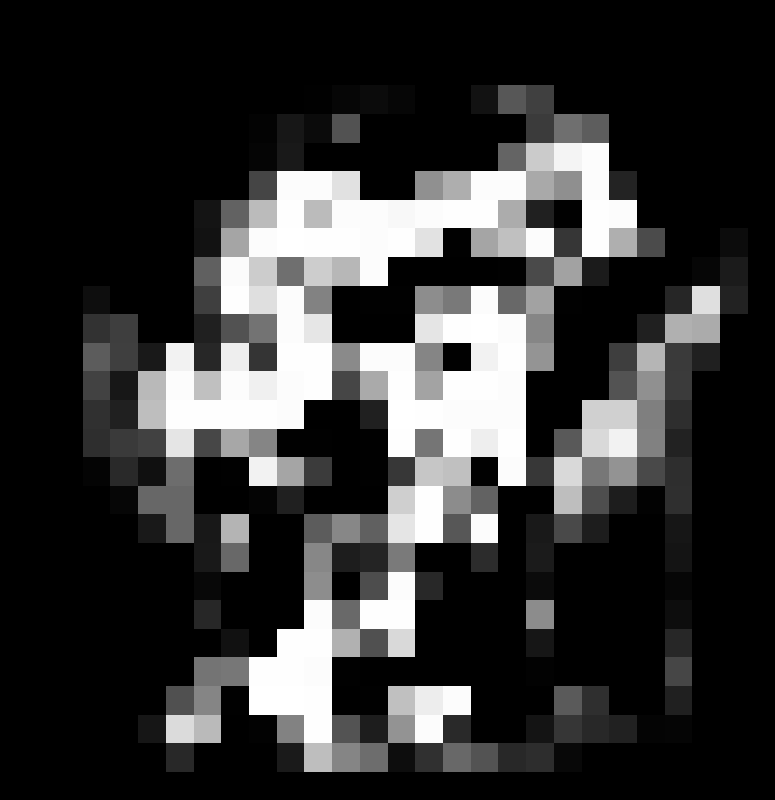}&
			\includegraphics[width=0.11\textwidth]{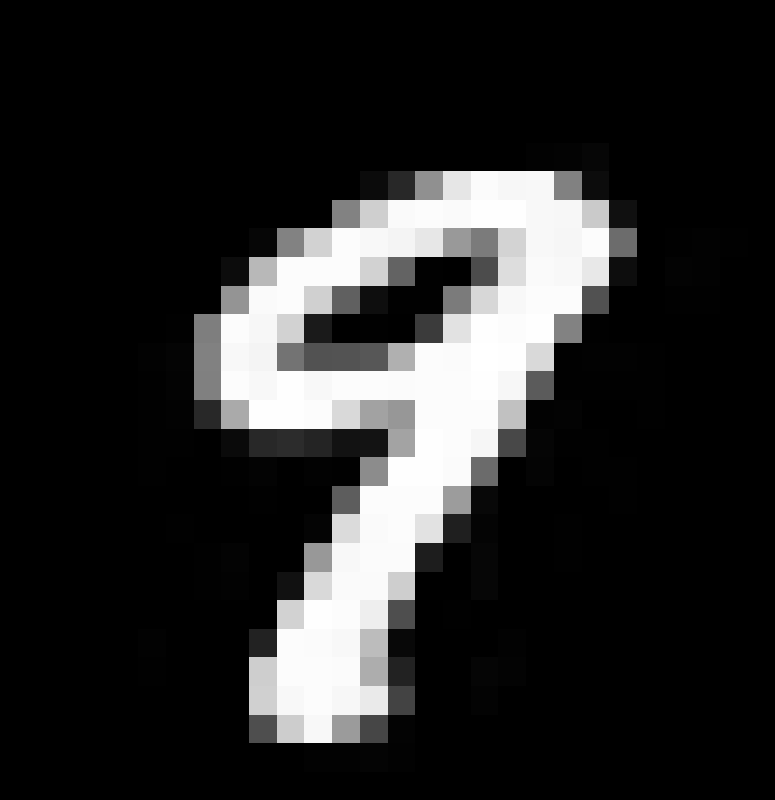}&
			\includegraphics[width=0.11\textwidth]{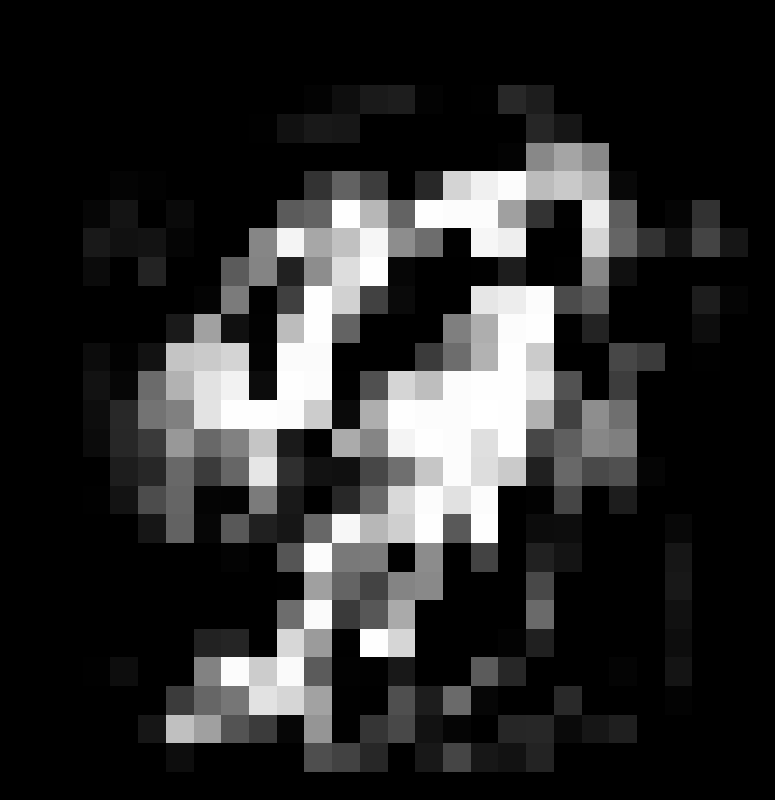}&
			\includegraphics[width=0.11\textwidth]{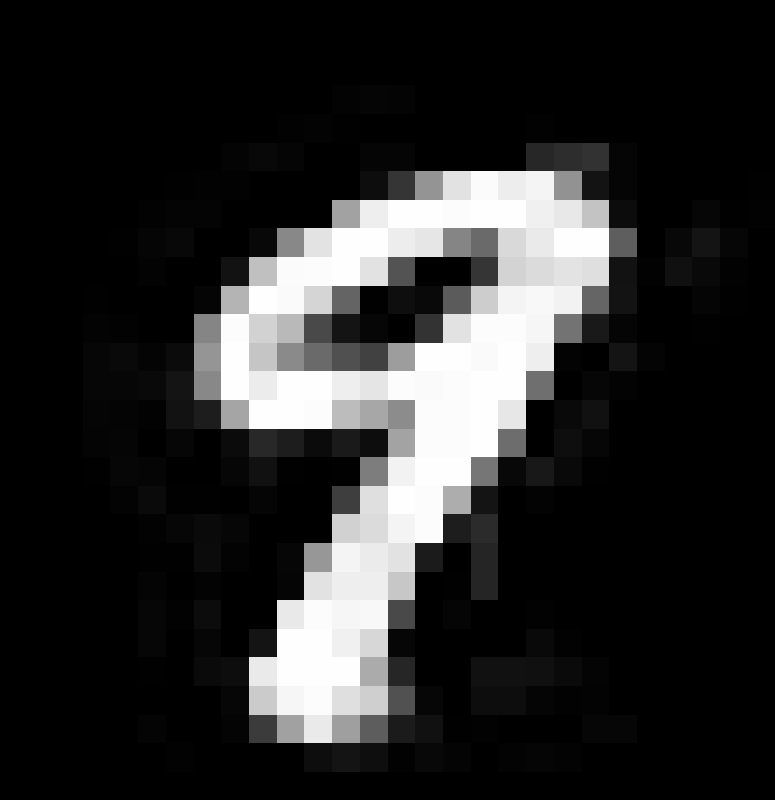}\\
		\end{tabular}}
		\caption{Poisoning digits crafted against linear and RBF SVMs. Larger perturbations are required to have significant impact on low-complexity classifiers ($L$), while minimal changes are very effective on high-complexity SVMs ($H$).}
		\vspace{-5pt}
		\label{fig:poisoning-digits}
	\end{figure}

	\noindent {\bf  How do poisoning attacks transfer between models in black-box settings?}
	The results for black-box poisoning are reported in Fig.~\ref{fig:pois-lk-mnist}.
	For poisoning attacks, the best surrogates are those matching the complexity of the target, as they tend to be better aligned and to share similar local optima, except for low-complexity logistic and ridge surrogates, which seem to transfer better to linear classifiers.
	This is also witnessed by gradient alignment in Fig.~\ref{fig:pois-corr-angles-mnist89}, which is again not only correlated to the similarity between black- and white-box perturbations (Fig.~\ref{fig:pois-scatter-mnist89-c}), but also to the ratio between the black- and white-box test errors (Fig.~\ref{fig:pois-scatter-mnist89-d}). 
	Interestingly, these error ratios are larger than one in some cases, meaning that attacking a surrogate model can be more effective than running a white-box attack against the target. 
	A similar phenomenon has been observed for evasion attacks~\cite{papernot16-transf}, and it is  due to the fact that optimizing attacks against a \emph{smoother} surrogate may find better local optima of the target function (\eg, by overcoming gradient obfuscation~\cite{athalye18}).
	According to our findings, for poisoning attacks, reducing the variability of the loss landscape (V) of the surrogate model is less important than finding a good alignment between the surrogate and the target. In fact, from Fig.~\ref{fig:pois-scatter-mnist89-b} it is evident that increasing V is even beneficial for SVM-based surrogates (and all these results are statistically significant according to the $p$-values in the sixth column of Table~\ref{tab:stats}).
	A visual inspection of the poisoning digits in Fig.~\ref{fig:poisoning-digits} reveals that the poisoning points crafted against high-complexity classifiers are only minimally perturbed, while the ones computed against low-complexity classifiers exhibit larger, visible perturbations. This is again due to the presence of closer local optima in the former case.
	Finally, a surprising result is that RFs are quite robust to poisoning, as well as NNs when attacked with low-complexity linear surrogates. The reason may be that these target classifiers have a large capacity, and can thus fit \emph{outlying} samples (like the digits crafted against low-complexity classifiers in Fig.~\ref{fig:poisoning-digits}) without affecting the classification of the other training samples.

	\subsubsection{Face Recognition}

	\begin{figure}[t]
		\centering
		\raisebox{0.02\height}{\includegraphics[width=.75\columnwidth,trim={0 0.2cm 0 0},clip]{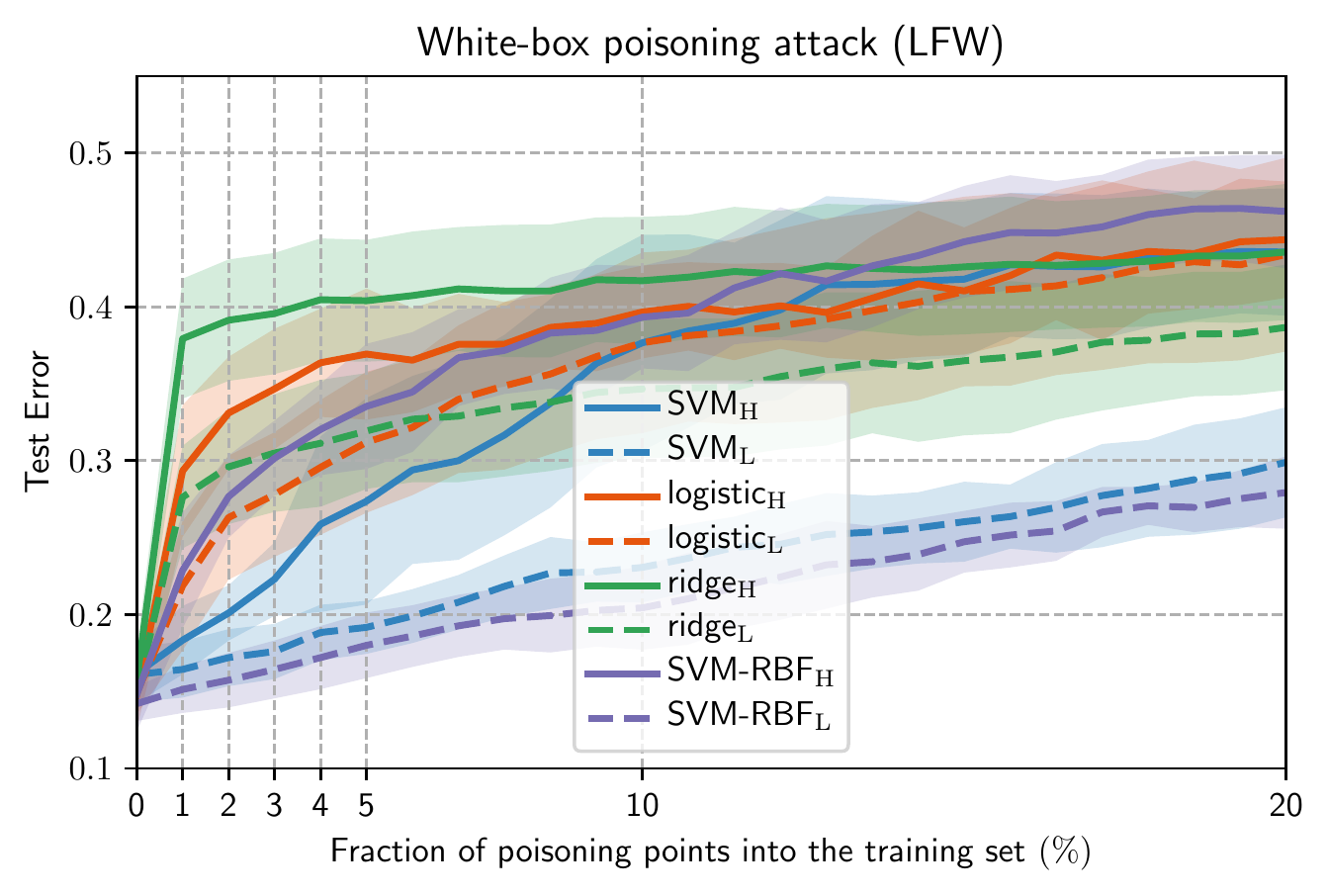}}
		\vspace{-0.5em}
		\caption{White-box poisoning attacks on LFW. Test error against an increasing fraction of poisoning points.}
		\vspace{-5pt}
		\label{fig:pois-pk-curves-lfw}
	\end{figure}
	
	\begin{figure*}[t]
		\centering
		\begin{subfigure}[t]{0.22\textwidth}
			\includegraphics[width=\textwidth]{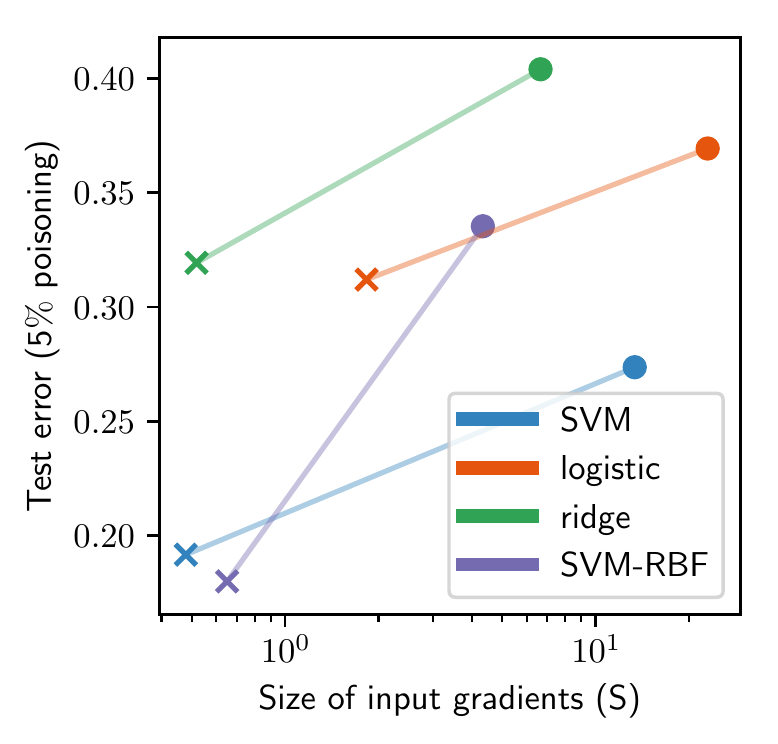}
			\subcaption{}
		\end{subfigure}
		\begin{subfigure}[t]{0.22\textwidth}
			\includegraphics[width=\textwidth]{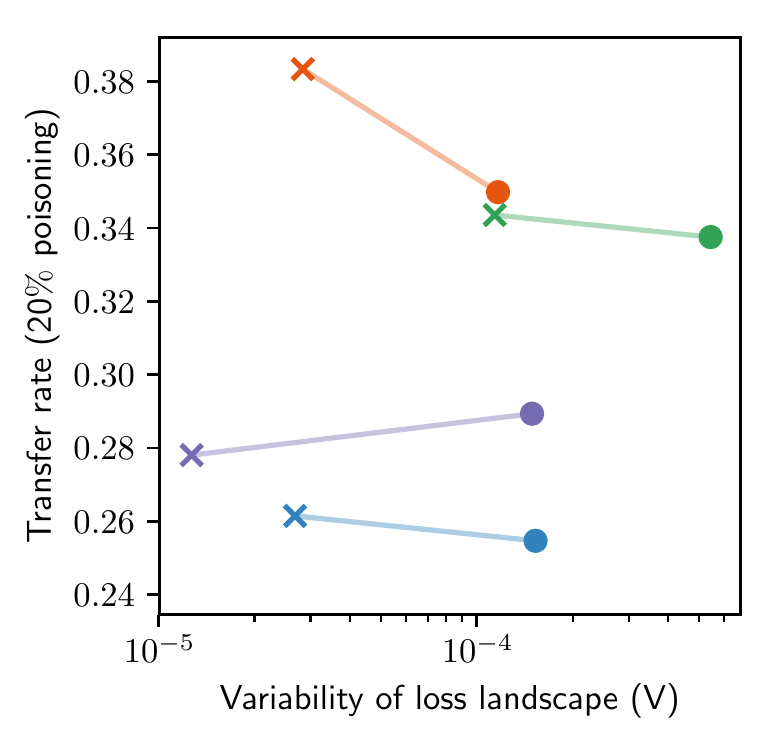}
			\subcaption{}
		\end{subfigure}
		\begin{subfigure}[t]{0.217\textwidth}
			\includegraphics[width=\textwidth]{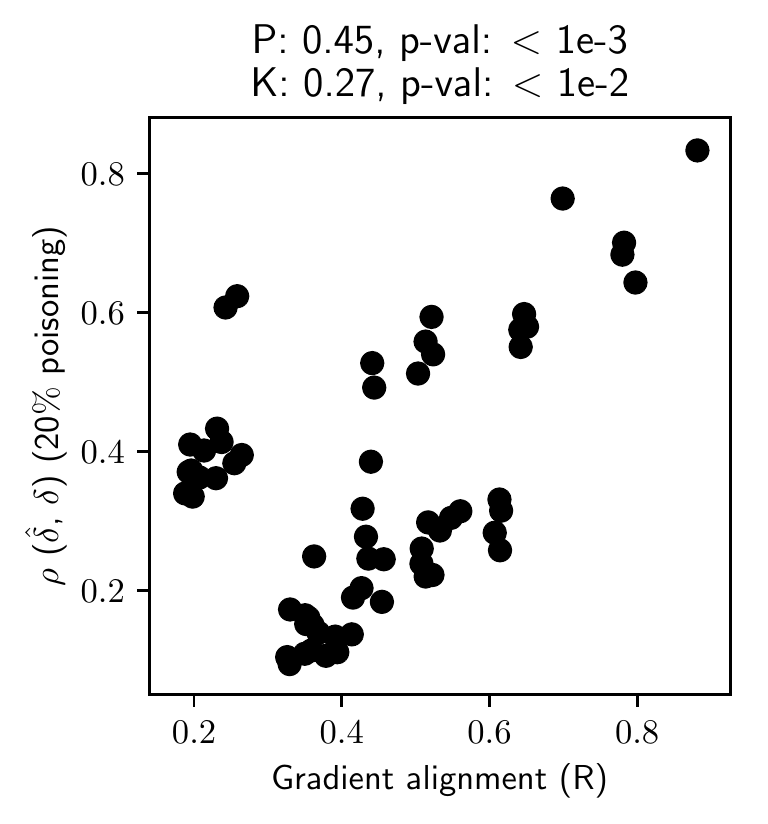}
			\subcaption{}
		\end{subfigure}
		\begin{subfigure}[t]{0.215\textwidth}
			\includegraphics[width=\textwidth]{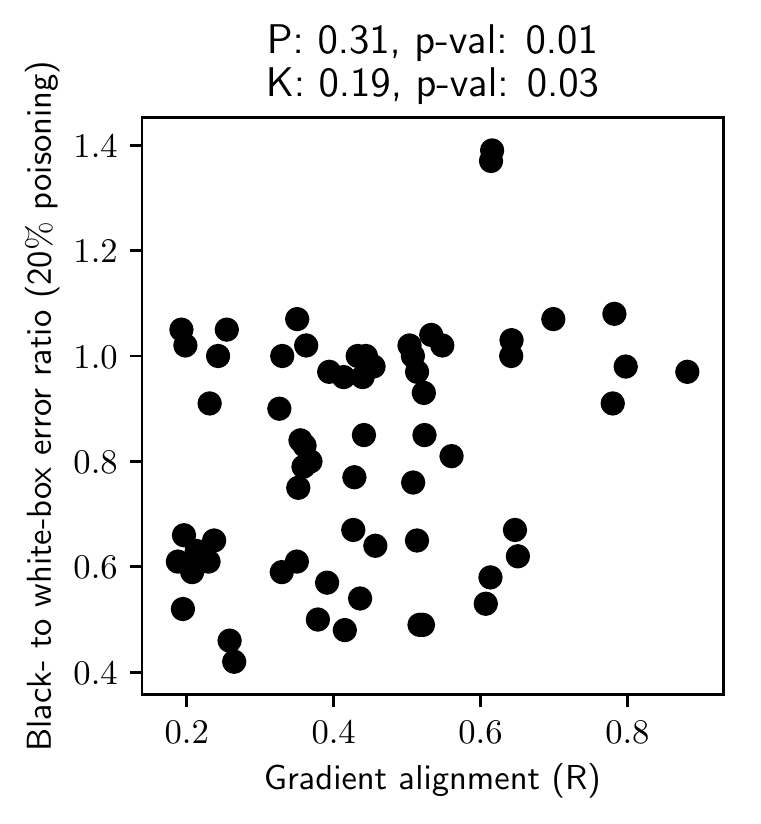}	
			\subcaption{}
		\end{subfigure}
		\caption{Evaluation of our metrics for poisoning attacks on LFW. See the caption of Fig.~\ref{fig:ev-scatter-mnist89} for further details.}
		\label{fig:pois-scatter-LFW}
	\end{figure*}
	
	\begin{figure*}[t]
		\centering
		\begin{subfigure}[t]{0.331\textwidth}
			\centering
			\includegraphics[width=\textwidth,trim={0 0 0 0},clip]{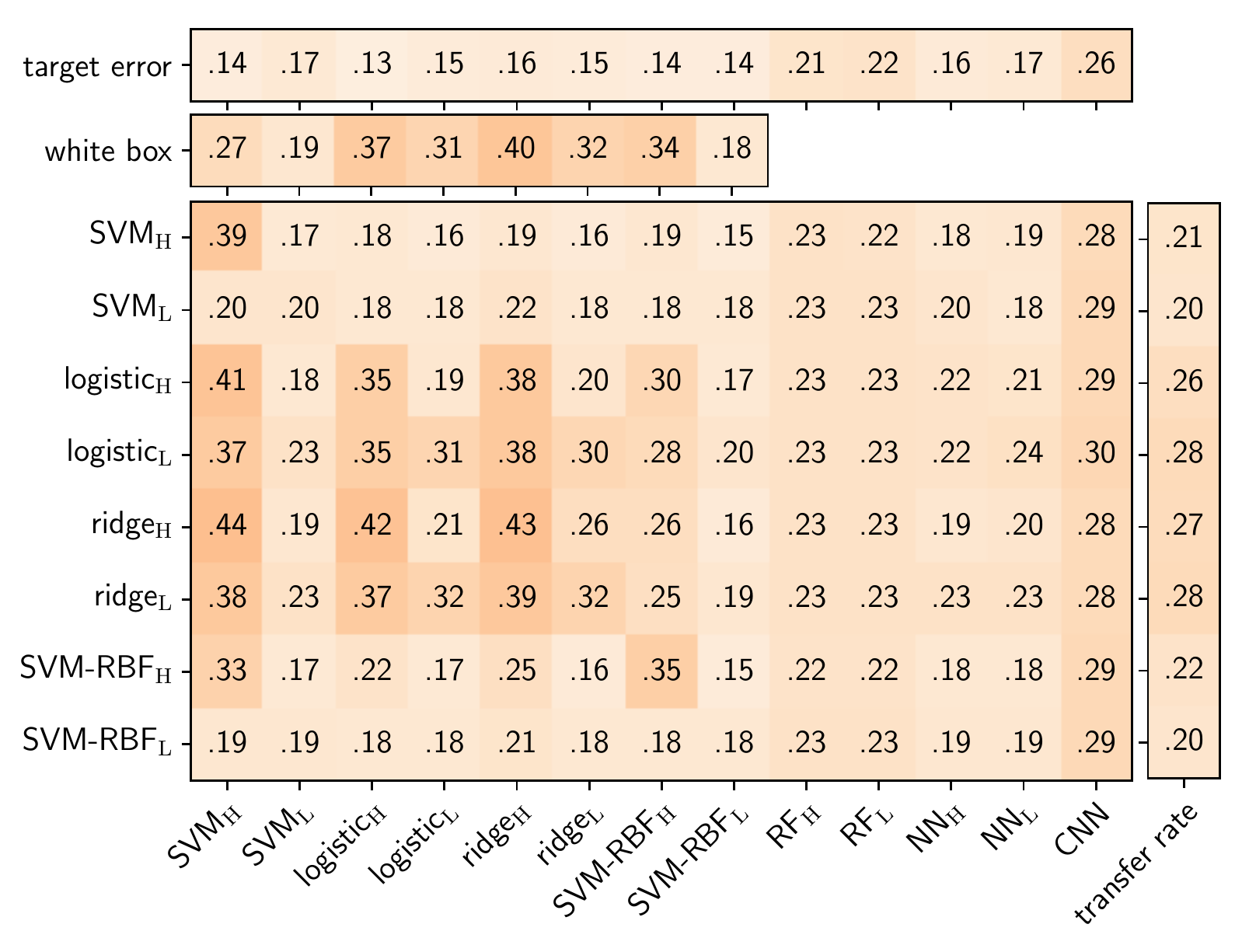}
			\subcaption{5\% poisoning}
		\end{subfigure}
		\begin{subfigure}[t]{0.285\textwidth}
			\centering
			\includegraphics[width=\textwidth,trim={2.20cm 0 0 0},clip]{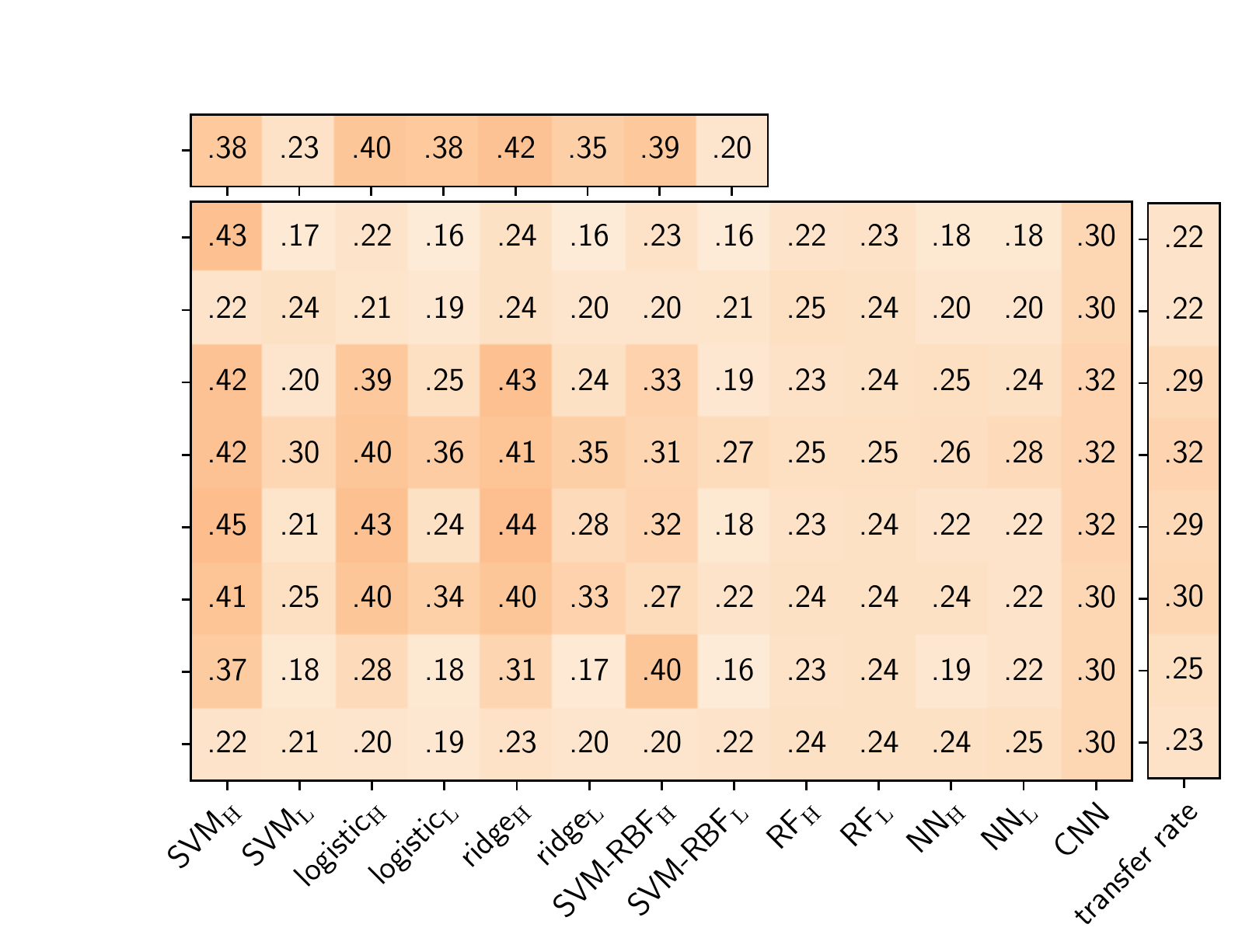}
			\subcaption{10\% poisoning}
		\end{subfigure}
		\begin{subfigure}[t]{0.285\textwidth}
			\centering
			\includegraphics[width=\textwidth,trim={2.20cm 0 0 0},clip]{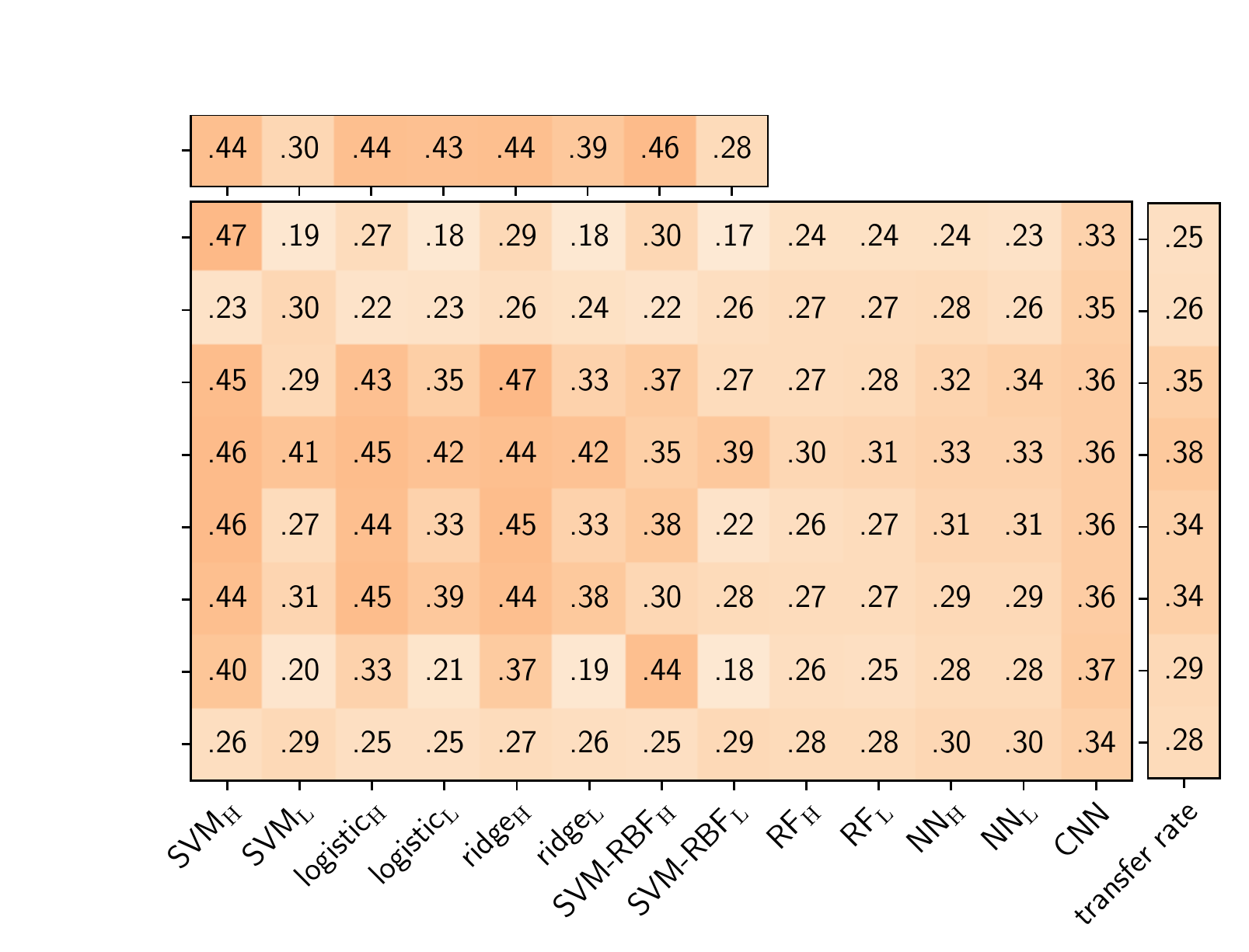}
			\subcaption{20\% poisoning}
		\end{subfigure}
		\caption{Black-box (transfer) poisoning attacks on LFW. See the caption of Fig.~\ref{fig:ev-lk-mnist89} for further details.}
		\label{fig:pois-lk-lfw}
	\end{figure*}
	
	\begin{figure}[t]
		\centering
		\includegraphics[height=.49\columnwidth]{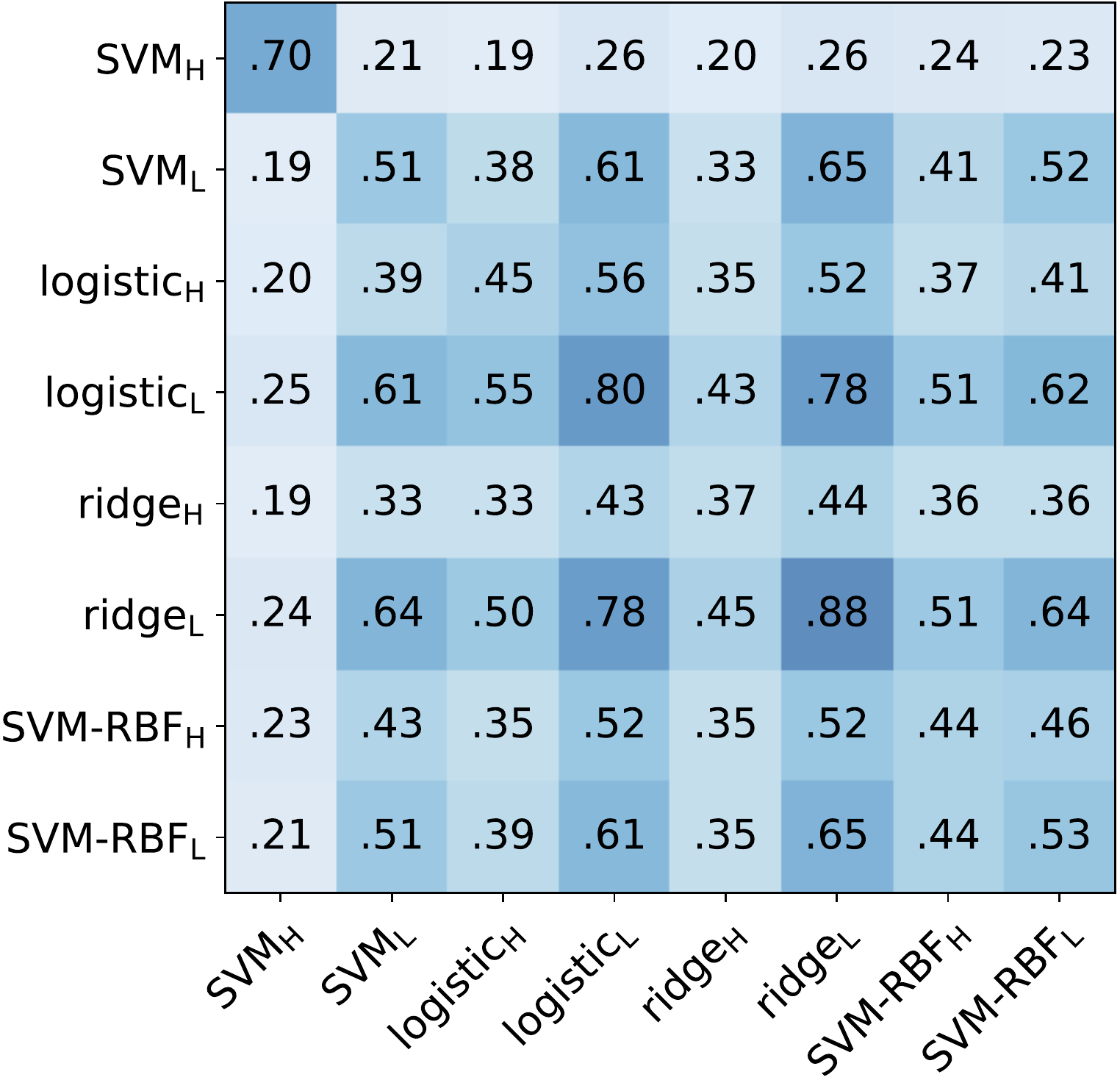}
		\includegraphics[height=.49\columnwidth]{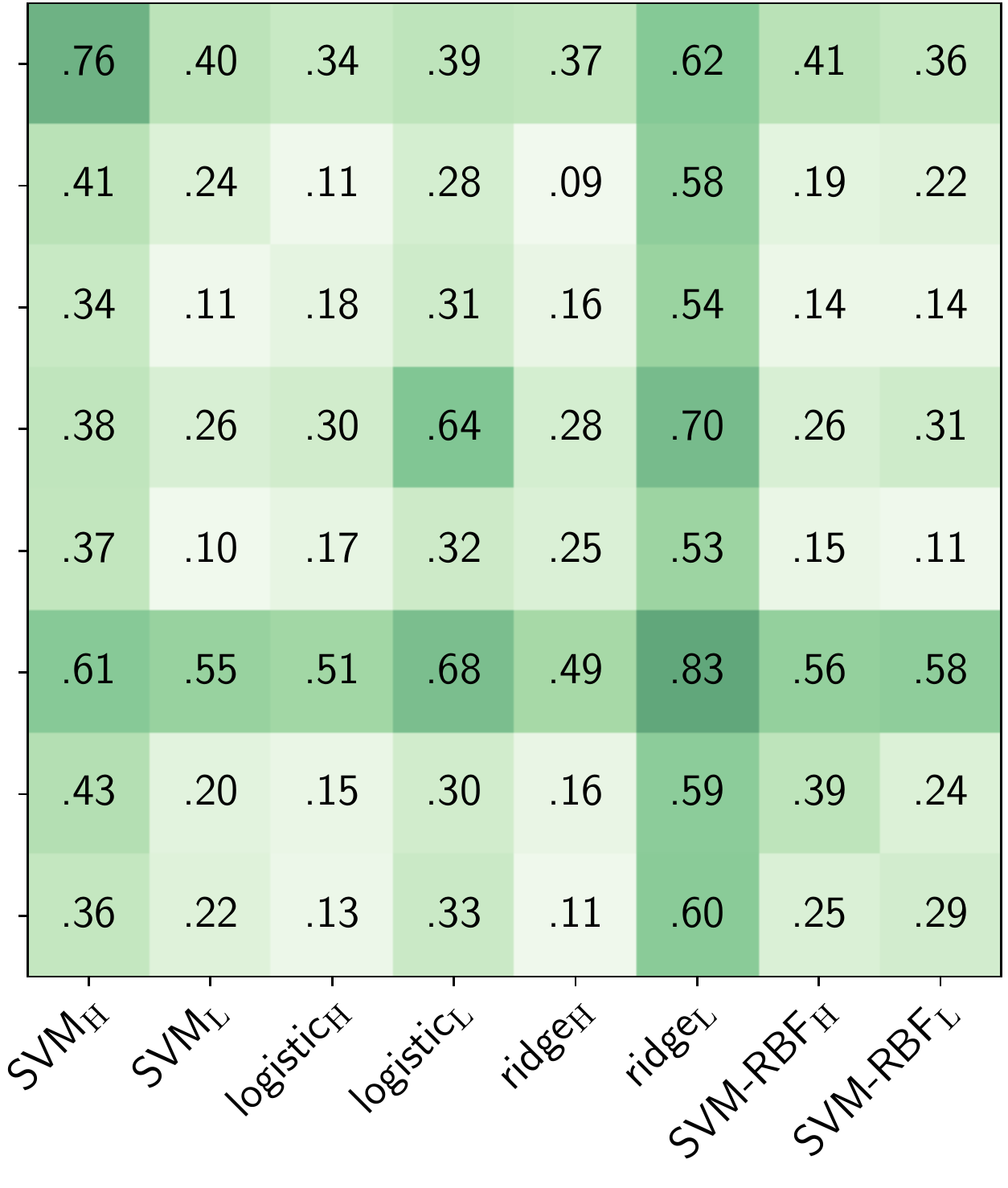}
		\vspace{-0.5em}
		\caption
		{Gradient alignment and perturbation correlation (at 20\% poisoning) for poisoning attacks on LFW. See the caption of Fig.~\ref{fig:ev-corr-angles-mnist89} for further details.}
		\vspace{-1pt}
		\label{fig:pois-corr-angles-lfw}
	\end{figure}

	\begin{figure}[ht]
		\centering
		\resizebox{0.95\columnwidth}{!}{
			\begin{tabular}{cccc}
				SVM$_{\mathrm L}$&
				SVM$_{\mathrm H}$&
				SVM-RBF$_{\mathrm L}$&
				SVM-RBF$_{\mathrm H}$\\
				\includegraphics[width=0.12\textwidth]{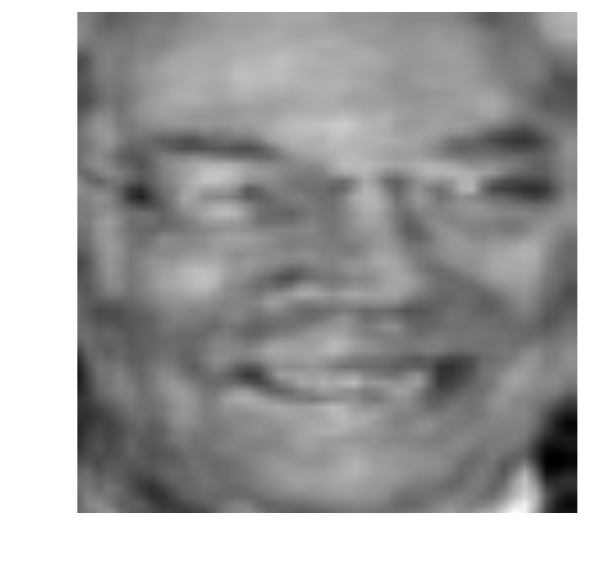}&
				\includegraphics[width=0.12\textwidth]{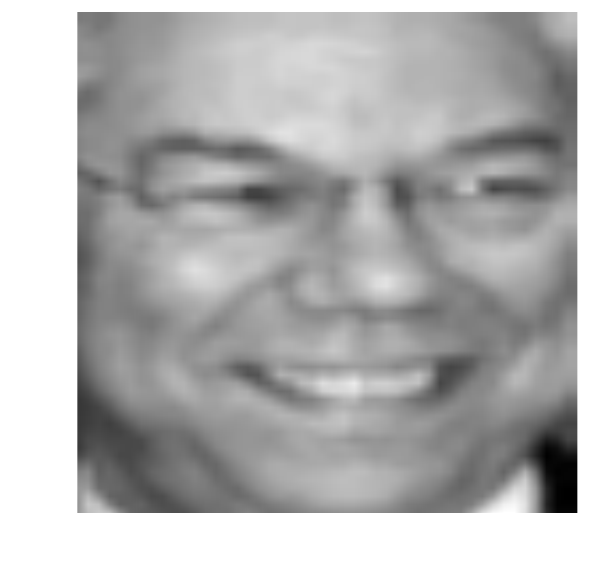}& 		
				\includegraphics[width=0.12\textwidth]{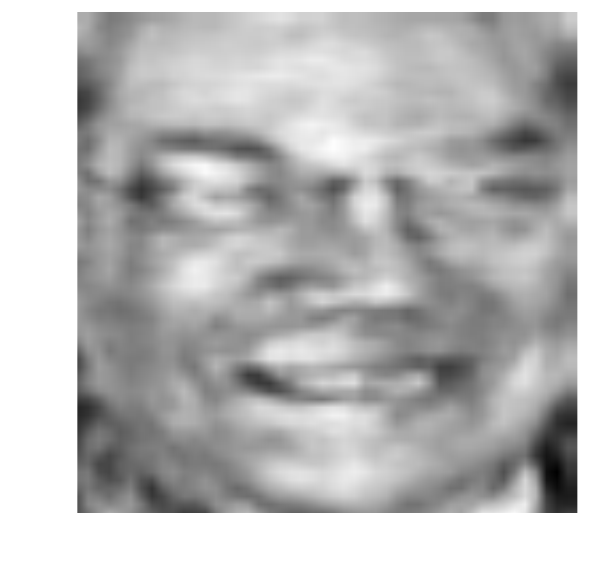}&
				\includegraphics[width=0.12\textwidth]{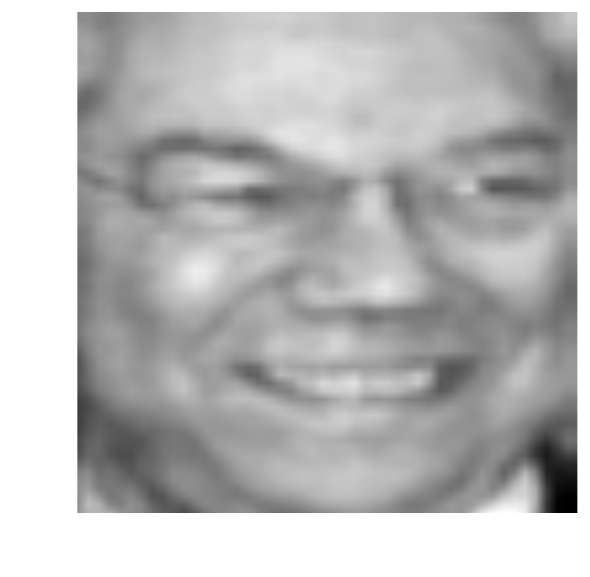} \\		
			\end{tabular}}
			\vspace{-5pt}
			\caption{Adversarial examples crafted against linear and RBF SVMs. Larger perturbations are required to have significant impact on low-complexity classifiers ($L$), while minimal changes are very effective on high-complexity SVMs ($H$).}
			\label{fig:svm_adv_examples}
		\end{figure}
		
		The Labeled Face on the Wild (LFW) dataset consists of faces of famous peoples collected on Internet.  We considered the six identities with the largest number of images in the dataset. We considered the person with most images as positive class, and all the others as negative. Our dataset consists of 530 positive and 758 negative images. 
		The classifiers and their hyperparameters are the same used for MNIST89, except that we set: ($i$) $C=0.1$ for logistic$_{\mathrm L}$, ($ii$) $\alpha=1$ for ridge$_{\mathrm H}$, ($iii$) $\gamma=0.001, C=10$ for SVM-RBF$_{\mathrm L}$, ($iv$) $\gamma=0.001, C=1000$ for SVM-RBF$_{\mathrm H}$, and ($v$) weight decay to $0.001$ for NN$_{\rm L}$.
		We run 10 repetitions with $300$ samples in each training, validation and test set.
		The results are shown in Figs~\ref{fig:pois-pk-curves-lfw},~\ref{fig:pois-scatter-LFW},~\ref{fig:pois-lk-lfw} and~\ref{fig:pois-corr-angles-lfw}, confirming the main findings discussed for poisoning attacks on MNIST89. Statistical tests for significance are reported in Table~\ref{tab:stats} (seventh and eighth columns). In this case, there is not a significant distinction between the mean transfer rates of high- and low-complexity surrogates, probably due to the reduced size of the training sets used. 
		Finally, in Fig.~\ref{fig:svm_adv_examples} we report examples of perturbed faces against surrogates with different complexities, confirming again that larger perturbations are required to attack lower-complexity models.

\subsection{Summary of Transferability Evaluation}

We summarize the results of transferability for evasion and poisoning attacks below.

\myparagraph{(1) Size of input gradients.} Low-complexity target classifiers are less vulnerable to evasion and poisoning attacks than high-complexity target classifiers trained with the same learning algorithm, due to the reduced size of their input gradients. In general, nonlinear models are more robust than linear models to both types of attacks.

\myparagraph{(2) Gradient alignment.} Gradient alignment is correlated with transferability. Even though it cannot be directly measured in black-box scenarios, some useful guidelines can be derived from our analysis. For evasion attacks, low-complexity surrogate classifiers provide stabler gradients which are better aligned, on average, with those of the target models; thus, it is generally preferable to use strongly-regularized surrogates. For poisoning attacks, instead, gradient alignment tends to improve when the surrogate matches the complexity (regularization) of the target (which may be estimated using techniques from~\cite{wang18-sp}).

\myparagraph{(3) Variability of the loss landscape.} Low-complexity surrogate classifiers provide loss landscapes with lower variability than high-complexity surrogate classifiers trained with the same learning algorithm, especially for evasion attacks. This results in better transferability.

To summarize, for evasion attacks, decreasing complexity of the surrogate model by properly adjusting the hyperparameters of its learning algorithm provides adversarial examples that transfer better to a range of models. For poisoning attacks, the best surrogates are generally models with similar levels of regularization as the target model. The reason is that the poisoning objective function is relatively stable (\ie, it has low variance) for most classifiers, and  gradient alignment between  surrogate and target becomes a more important factor.

Understanding attack transferability has two main implications. First, even when attackers do not know the target classifier, our findings suggest that low-complexity surrogates have a better chance of transferring to other models. Our recommendation to performing black-box evasion attacks  is  to choose surrogates with low complexity (\eg, by using strong regularization and reducing model variance). To perform poisoning attacks, our recommendation is to acquire additional information about the level of regularization of the target and train a surrogate model with a similar level of regularization. Second, our analysis also provides recommendations to defenders on how to design more robust models against evasion and poisoning attacks. In particular, lower-complexity models tend to have more resilience compared to more complex models. Of course, we need to take into account the bias-variance trade-off and choose models that still perform relatively well on the original prediction tasks.

\section{Related Work}
\label{sect:related}

\myparagraph{Transferability for evasion attacks.}
Transferability of evasion attacks has been studied in previous work, e.g.,~\cite{biggio13-ecml,szegedy14-iclr,goodfellow15-iclr,papernot16-transf,papernot17-asiaccs,moosavi17-cvpr,dong18-cvpr,liu17-iclr,tramer17-transf,wu18}.
Biggio et al.~\cite{biggio13-ecml} have been the first to consider evasion attacks against surrogate models in a limited-knowledge scenario. Goodfellow et al.~\cite{goodfellow15-iclr}, Tramer et al.~\cite{tramer17-transf}, and Moosavi et al.~\cite{moosavi17-cvpr} have made the observation that different models might learn intersecting decision boundaries in both benign and adversarial dimensions and in that case adversarial examples transfer better. Tramer et al. have also performed a detailed study of transferability of model-agnostic perturbations that depend only on the training data, noting that adversarial examples crafted against linear models can transfer to higher-order models. We answer some of the open questions they posed about factors contributing to attack transferability. Liu et al.~\cite{liu17-iclr} have empirically observed the gradient alignment between models with transferable adversarial examples. Papernot \etal~\cite{papernot16-transf,papernot17-asiaccs} have observed that adversarial examples transfer across a range of models, including logistic regression, SVMs and neural networks, without providing a clear explanation of the phenomenon.
Prior work has also investigated the role of input gradients and Jacobians. Some authors have proposed to decrease the magnitude of input gradients during training to defend against evasion attacks~\cite{lyu15-icdm,ross18} or improve classification accuracy~\cite{sokolic17,varga17-arxiv}. In~\cite{simon18,ross18}, the magnitude of input gradients has been identified as a cause for vulnerability to evasion attacks.
A number of papers have shown that transferability of adversarial examples is increased by averaging the gradients computed for ensembles of models~\cite{liu17-iclr,wu18,tramer17-transf,dong18-cvpr}. We highlight that we obtain similar effect by attacking a strongly-regularized surrogate model with a smoother and stabler decision boundary (resulting in a lower-variance model). The advantage of our approach is to reduce the computational complexity compared to attacking classifier ensembles.
Through our formalization, we shed light on the most important factors for transferability. In particular, we identify a set of conditions that explain transferability including the gradient alignment between the surrogate and targeted models, and the size of the input gradients of the target model, connected to model complexity. We demonstrate that adversarial examples crafted against lower-variance models (e.g., those that are strongly regularized) tend to transfer better across a range of models.

\myparagraph{Transferability for poisoning attacks.} There is very little work on the transferability of poisoning availability attacks, except for a preliminary investigation in~\cite{biggio17-aisec}. That work indicates that poisoning examples are transferable among very simple network architectures (logistic regression, MLP, and Adaline). Transferability of targeted poisoning attacks has been addressed recently in~\cite{suciu18-usenix}. We are the first to study in depth transferability of poisoning availability attacks.

\section{Conclusions}
\label{sec:conclusions}

We have conducted an analysis of the transferability of evasion and poisoning attacks under a unified optimization framework. Our theoretical transferability formalization sheds light on various factors impacting the transfer success rates. In particular, we have defined three metrics that impact the transferability of an attack, including the  complexity of the target model, the gradient alignment between the surrogate and target models, and the variance of the attacker optimization objective. The lesson to system designers is to evaluate their classifiers against these criteria and select lower-complexity, stronger regularized models that tend to provide higher robustness to both evasion and poisoning.
Interesting avenues for future work include extending our analysis to multi-class classification settings, and considering a range of gray-box models in which attackers might have additional knowledge of the machine learning system (as in \cite{suciu18-usenix}). Application-dependent scenarios such as cyber security might provide additional constraints on threat models and attack scenarios and could impact transferability in interesting ways.

\vspace{-2pt}
\section*{Acknowledgements}
The authors would like to thank Neil Gong for shepherding our paper and the anonymous reviewers for their constructive feedback.
This work was partly supported by the EU H2020 project ALOHA, under the European Union's Horizon 2020 research and innovation programme (grant no.780788). This research was also sponsored by the Combat Capabilities Development Command Army Research Laboratory and was accomplished under Cooperative Agreement Number W911NF-13-2-0045 (ARL Cyber Security CRA). The views and conclusions contained in this document are those of the authors and should not be interpreted as representing the official policies, either expressed or implied, of the Combat Capabilities Development Command Army Research Laboratory or the U.S. Government. The U.S. Government is authorized to reproduce and distribute reprints for Government purposes not withstanding any copyright notation here on. We would also like to thank Toyota ITC for funding this research.

\end{document}